\documentclass[final,5p,times,twocolumn,nopreprintline]{elsarticle}

\usepackage{framed,multirow}

\usepackage{amssymb}
\usepackage{latexsym}

\usepackage{placeins}
\usepackage{booktabs,arydshln}
\usepackage{subfig}
\usepackage{lipsum} 
\usepackage{array}
\newcolumntype{P}[1]{>{\centering\arraybackslash}p{#1}}
\usepackage{tabularx} 

\usepackage{pgfplots}
\usepackage{tikz}
\usetikzlibrary{positioning}
\usetikzlibrary{arrows.meta}
\usetikzlibrary{calc}
\usetikzlibrary{shapes.geometric}
\usepackage{xcolor}

\usepackage{pifont}
\usepackage{suffix}		
\newcommand\sref[1]{\protect\subref{#1}}
\WithSuffix\newcommand\sref*[1]{\protect\subref*{#1}}

\usepackage{hyperref}
\usepackage{url}

\makeatletter
\def\adl@drawiv#1#2#3{%
        \hskip.5\tabcolsep
        \xleaders#3{#2.5\@tempdimb #1{1}#2.5\@tempdimb}%
                #2\z@ plus1fil minus1fil\relax
        \hskip.5\tabcolsep}
\newcommand{\cdashlinelr}[1]{%
  \noalign{\vskip\aboverulesep
           \global\let\@dashdrawstore\adl@draw
           \global\let\adl@draw\adl@drawiv}
  \cdashline{#1}
  \noalign{\global\let\adl@draw\@dashdrawstore
           \vskip\belowrulesep}}
\makeatother

\usepackage{fancyhdr}
\begin{document}

\begin{frontmatter}

\title{Methods and datasets for segmentation of minimally invasive surgical instruments in endoscopic images and videos: A review of the state of the art}

\author[1]{Tobias Rueckert\corref{cor1}}
    \cortext[cor1]{
    Corresponding author. \\
    \indent \indent \hspace{-1.2mm} \textit{E-mail address:} tobias.rueckert@oth-regensburg.de \\
	This article is published under the CC BY-NC-ND license (\url{https://creativecommons.org/licenses/by-nc-nd/4.0/}).
        }
    \author[2,3]{Daniel Rueckert}
    \author[1,4]{Christoph Palm}
    
    \address[1]{Regensburg Medical Image Computing (ReMIC), Ostbayerische Technische Hochschule Regensburg (OTH Regensburg), Germany}
    \address[2]{Artificial Intelligence in Healthcare and Medicine, Klinikum rechts der Isar, Technical University of Munich, Germany}
    \address[3]{Department of Computing, Imperial College London, UK}
    \address[4]{Regensburg Center of Health Sciences and Technology (RCHST), OTH Regensburg, Germany}

\begin{abstract}
    In the field of computer- and robot-assisted minimally invasive surgery, enormous progress has been made in recent years based on the recognition of surgical instruments in endoscopic images and videos. 
    In particular, the determination of the position and type of instruments is of great interest.  
    Current work involves both spatial and temporal information, with the idea that predicting the movement of surgical tools over time may improve the quality of the final segmentations. 
    The provision of publicly available datasets has recently encouraged the development of new methods, mainly based on deep learning. 
    In this review, we identify and characterize datasets used for method development and evaluation and quantify their frequency of use in the literature.
    We further present an overview of the current state of research regarding the segmentation and tracking of minimally invasive surgical instruments in endoscopic images and videos.
    The paper focuses on methods that work purely visually, without markers of any kind attached to the instruments, considering both single-frame semantic and instance segmentation approaches, as well as those that incorporate temporal information.
    The publications analyzed were identified through the platforms Google Scholar, Web of Science, and PubMed. 
    The search terms used were ``instrument segmentation'', ``instrument tracking'', ``surgical tool segmentation'', and ``surgical tool tracking'', resulting in a total of 741 articles published between 01/2015 and 07/2023, of which 123 were included using systematic selection criteria.
    A discussion of the reviewed literature is provided, highlighting existing shortcomings and emphasizing the available potential for future developments.
\end{abstract}

 
\end{frontmatter}


\section{Introduction}
\label{introduction}

Minimally invasive surgery (MIS) offers several advantages over conventional types of surgery, making it the standard method for many surgical interventions today~\cite{darzi2002recent, hammad2019open, de2019minimally, sluis2019robot}.
By introducing surgical instruments through narrow incisions, surgical trauma can be significantly reduced, ensuring faster patient recovery and a shortened hospital stay~\cite{darzi2002recent}.
The movements of the surgical tools in the human body are commonly recorded using an endoscopic camera, and the images are made visible to the surgeon on a monitor.
However, MIS also faces several challenges arising from the more elaborate surgical procedure, the limited field of view, and the complex hand-eye coordination.
In addition, the guidance of the endoscope by a human assistant results in increased time and costs~\cite{fuchs2002minimally}.

To address these challenges, the research area of computer- and robot-assisted minimally invasive surgery (RAMIS) has been experiencing increasing interest in recent years~\cite{haidegger2022robot, maier2022surgical}.
The development of new techniques in this field aims to assist the surgeon in overcoming these drawbacks in the best possible way and opens up further possibilities such as surgical skill assessment, operating workflow optimization, or the training of junior surgeons.
One popular approach to this is the utilization of information regarding the positions and movements of the surgical instruments over time.
This can be achieved by tracking surgical tools using electromagnetic or infrared-based methods or by attaching external markers~\cite{bouarfa2012vivo, Mamone2017RobustLI, sorriento2019optical, wang2022visual}.
However, all of these techniques have in common that they require additional effort to prepare for surgery.
In addition, these approaches cannot be easily integrated into the existing workflow of an operation.

For these reasons, current work focuses exclusively on purely visual-based processing of the endoscopic video stream.
Due to the great success of deep-learning-based methods in the field of image processing in recent years, approaches developed in this way also represent the state of the art for this application~\cite{anteby2021deep, rivas2021review, yang2020image}.
Here, the detection of surgical instruments is often performed by object detection methods that localize the tools through comprehensive axis-aligned bounding boxes and follow them across multiple video frames using appropriate tracking methods~\cite{qiu2019real, wang2022visual, zhao2019real}.
Approaches developed using this technique often exhibit fast processing time but are usually imprecise since surgical instruments often protrude from the bottom two corners into the center of an image, resulting in maximally unfavorable conditions for the axis-aligned bounding box detection.

In contrast, segmentation methods allow a much more accurate determination of the position of surgical tools in images since the shape of the instruments can also be predicted on the basis of this pixel-by-pixel classification.
Moreover, the technical advances achieved in recent years make it possible to segment surgical tools in real-time, thus providing an accurate and fast method~\cite{huang2022simultaneous, islam2019learning, islam2019real, jha2021exploring,pakhomov2020searching}.
In addition, these characteristics can be further improved by incorporating the temporal information inherent in video recordings through novel techniques.
In addition to these technical innovations, the availability of publicly accessible datasets has also supported the increased interest in this research area.
A specific contribution to this development is presented by the organization of challenges conducted as part of an annual conference organized by the Medical Image Computing and Computer Assisted Intervention Society (MICCAI)\footnote{\url{http://www.miccai.org/}}~\cite{surgical_tool_datasets}.

In the present work, we analyze current approaches concerning semantic and instance-based segmentation of surgical instruments in endoscopic images and videos and provide possible research directions. 
To this end, the contributions of this work are threefold:

\begin{itemize} 
\item Identification and characterization of robotic and non-robotic datasets used for method development and evaluation and quantification of the frequency of use in the literature.
\item Systematic search and analysis of articles concerning semantic and instance-based segmentation of minimally invasive surgical instruments in endoscopic images and videos.
\item Discussion of the reviewed literature, identification of existing shortcomings, and highlighting the potential for future developments.
\end{itemize}

The structure of this document is as follows.
In Chapter~\ref{related_work}, we describe related reviews and their considered research areas and highlight differences between the present work and these publications.
Chapter~\ref{review_methodology} presents the conducted systematic search, describes the criteria used to select the relevant publications, and analyzes the results. 
The datasets employed in the relevant publications are described in more detail in Chapter~\ref{datasets}, both in terms of the frequency with which the datasets were used as well as concerning their characteristic properties.
A description of the selected literature regarding semantic segmentation of surgical instruments is provided in Chapter~\ref{semantic_segmentation_methods}.
Here, the articles are divided into single-image segmentation and segmentation involving temporal information.
Instance-based methods of current literature are analyzed in Chapter~\ref{instance_segmentation_methods}, following the same structure.
Subsequently, Chapter~\ref{further_processing} presents articles that use the segmentation of surgical tools as a basis for further processing.
The discussion of the results of the systematic search takes place in Chapter~\ref{discussion}.
Finally, Chapter~\ref{conclusion} summarizes the results of this work, highlights the main findings, and provides an outlook on possible further developments in this research area.

\section{Related Work}
\label{related_work}

In the following, we briefly describe the objectives of related reviews and contextualize the contributions of the present work. 
For this purpose, we only consider papers that exclusively address visually-based approaches without markers of any kind and that have been published since 2015.
The works are arranged chronologically in ascending order.

In 2017, Bouget et al.~\cite{bouget2017vision} published a comprehensive review of the current state of research, identifying and analyzing commonly used datasets for the development and testing of surgical tool recognition algorithms, providing an in-depth comparison of instrument detection methods as well as highlighting existing shortcomings and providing an analysis of validation techniques to measure the quality of the approaches considered. For this purpose, the authors reviewed datasets and publications between 2000 and 2015, using the keywords ``surgical tool detection'', ``surgical tool tracking'', ``surgical instrument detection'', and ``surgical instrument tracking'' on the search platforms Google Scholar and PubMed, resulting in a total of 28 articles, and considered both endoscopic and microscopic image data.

Three years later, Yang et al.~\cite{yang2020image} presented a survey with similar goals, focusing on methods based on Convolutional Neural Networks (CNNs) architectures.
The authors group the considered papers based on methodological CNN-oriented approaches but do not provide specific details about the applied review method, such as the used search platforms or the number of observed publications.
The search terms are composed of two necessary parts, one consisting of the terms ``neural network*'' or ``deep learning'', and the other of ``mini* invasive'', ``robot* surger*'', ``robot* surgical'', or ``laparoscop*'', where the * operator represents an arbitrary continuation of the respective expression.

In 2021, the work of Anteby et al.~\cite{anteby2021deep} was published, which, in addition to determining the presence and localization of surgical tools, considers many other tasks, such as surgical phase detection, classification and segmentation of anatomy, determination of surgical actions, or even prediction of the required surgical time.
A total of 32 identified papers were considered relevant after searching the platforms Medline, Embase, IEEE Xplore, and Web of Science for publications between January 2012 and May 2020.

In the same year, Rivas-Blanco et al.~\cite{rivas2021review} provided a review on deep-learning-based methods in minimally invasive surgery that covers the area of surgical image analysis, surgical task analysis, surgical skill assessment, and automation of surgical skills.
The keywords for the search on the IEEE Xplorer, Springer Link, Science Direct, and ACM Digital Library platforms consisted of two parts, the first of which had to contain the terms ``deep learning'' or ``deep neural network'' and the second of which had to contain ``laparoscopic surgery'', ``minimally invasive surgery'', ``robotic surgery'', or ``robot-assisted surgery''. A total of 85 papers between 2015 and 2020 were considered.

In 2022, Nema and Vachhani~\cite{nema2022surgical} investigated current trends in artificial intelligence (AI)-based visual detection and tracking of surgical instruments and its application to surgical skill assessment.
The authors do not specify details regarding the applied review methodology, such as databases considered, search terms, or time period.

A more comprehensive analysis and summary of current research regarding visual detection and tracking of minimally invasive surgical instruments was presented in 2022 by Wang et al.~\cite{wang2022visual}, who analyze the literature concerning various aspects.
The approaches were identified by searching the platforms Web of Science, Google Scholar, PubMed, and CNKI using the search terms ``object detection'', ``object tracking'', ``surgical tool detection'', ``surgical tool tracking'', ``surgical instrument detection'', and ``surgical instrument tracking'', including articles published between 1985 and 2021.

\begin{figure*}[tb]
	\includegraphics[width=\textwidth]{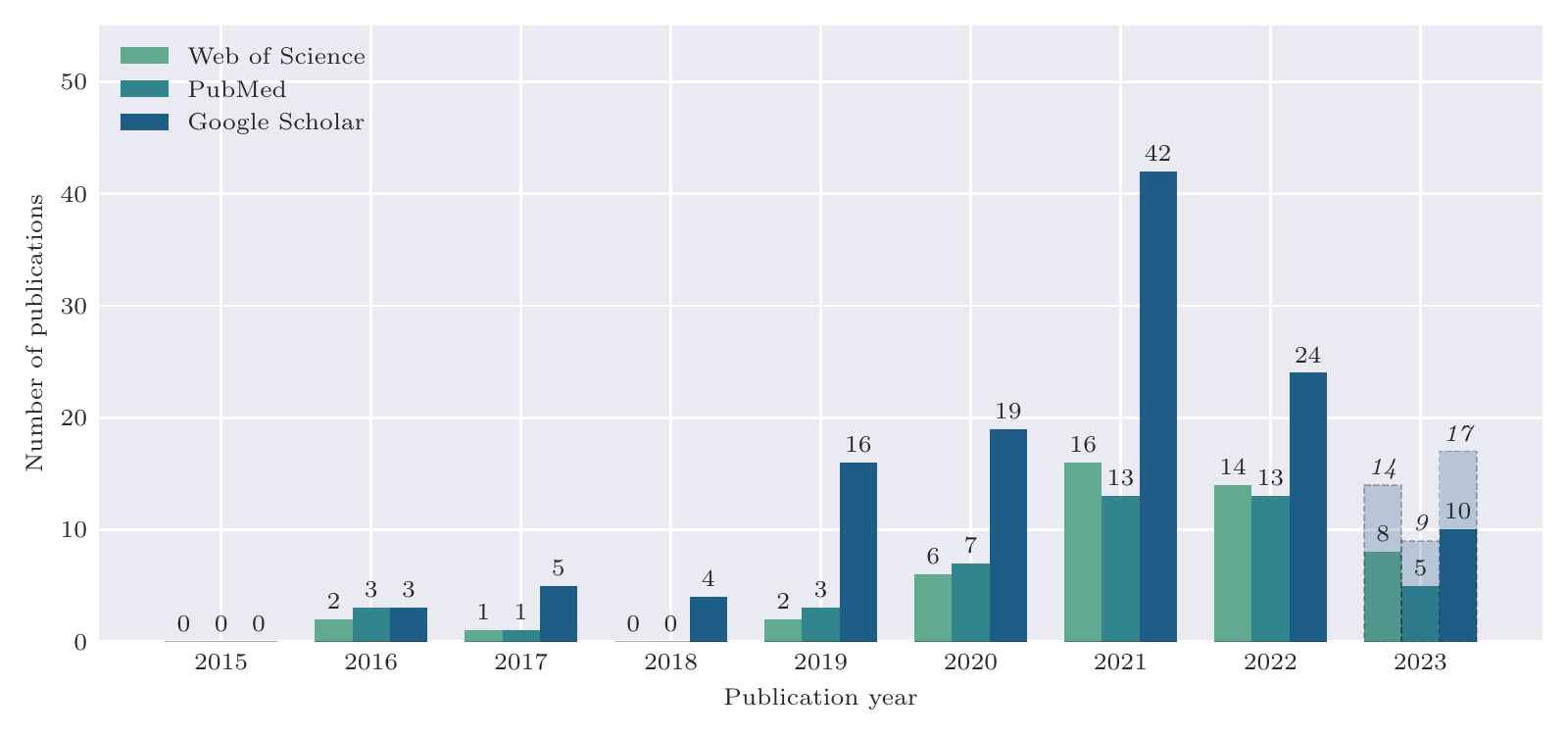}
	\caption{Number of relevant publications by year resulting from searches on the platforms Web of Science, PubMed, and Google Scholar, according to the selection criteria. For 2023, in addition to the publications up to and including July, the numbers of expected papers by the end of the year are shown in italics at the top of the transparent boxes.}
	\label{fig:review_method:publications}
\end{figure*}

Looking at these reviews, it is noticeable that the focus is often on tasks such as binary tool presence detection, surgical action determination, surgical phase recognition, or surgical skill assessment, with object detection methods using axis-oriented bounding boxes being the central approach while segmentation of minimally invasive instruments tends to play a minor role.
This observation is due to the fact that none of the search terms used in the aforementioned works contains the keyword ``segmentation'', whereas the term ``detection'' is frequently used.
Due to the rapid increase in the number of publications in this area in recent years, which is illustrated in Figure~\ref{fig:review_method:publications} and described in more detail in Chapter~\ref{review_methodology}, some of the results of the above reviews no longer represent the current state of the art, both in terms of freely available datasets as well as regarding published methods.
Another limitation of the presented works is that the search methods used often lack central details, such as the time period considered, the search platforms used, keywords, inclusion and exclusion criteria, or the number and type of articles identified, which reduces the traceability and informative value of the results.

The focus of this paper is explicitly to provide an overview of the current state of research concerning methods and datasets for the semantic and instance-based segmentation and tracking of minimally invasive surgical instruments in endoscopic images and videos.
To this end, we specify our approach in detail concerning the search methodology, including the search platforms, the search terms, the time period, the information extracted from the articles, and the selection criteria used, followed by an analysis of the results.
In order to demonstrate the transferability and applicability of segmentation methods to clinical applications, we also present publications that use surgical instrument segmentation as a basis for further processing in robotic-assisted settings.
We present a discussion and in-depth analysis of the literature reviewed, related to the datasets currently \pagebreak being used to develop and validate new approaches and related to the methodologies identified, emphasizing existing shortcomings and highlighting the available potential for future developments.

\section{Review Methodology}
\label{review_methodology}

The publications for this work were identified by systematic search on the platforms Google Scholar, Web of Science, and PubMed. 
The search terms used were ``instrument segmentation'', ``instrument tracking'', ``surgical tool segmentation'', and ``surgical tool tracking'', and all articles published between 01/2015 and 07/2023 were included.
For all platforms, the search was based on the titles of the publications. 
An article was retrieved if all parts appeared for one of these search terms.
For each publication, we identified the title, the datasets used, and the type of segmentation, i.e., whether it is semantic or instance segmentation, as well as whether it concerns binary, instrument type, or instrument part segmentation.
Furthermore, for each article, it was reviewed whether temporal information was used to track surgical instruments or to improve segmentation quality, and if so, what methodology was used to do so. 
Only publications written in the English language and which have undergone a peer-review process were considered, resulting in the exclusion of 5 and 26 articles, respectively.
In the following, the criteria that a publication must fulfill to be included as a relevant paper in this review are presented first.
Subsequently, the results of the systematic search are analyzed. 

\subsection{Selection Criteria}
\label{method:selection_criteria}

We considered only publications with methods developed and validated on the basis of endoscopic imaging data, excluding all articles employing data of a different nature. 
A publication meets the criterion if the developed method has been validated with at least one endoscopic data set.
Furthermore, we concentrate on new methodological developments and not on the pure application of methods in the field of semantic and instance-based segmentation of minimally invasive surgical instruments.
Articles concerning other research areas as object recognition or pose estimation are excluded from the results.
The processing of the endoscopic image data must be exclusively visual, which means that we excluded methods utilizing external markers of all possible types on the instruments from the results. 
Publications that do not present new methodological developments regarding the segmentation of surgical tools, but which utilize the results of the segmentation as a foundation for further actions, were included in the present work and described separately from the novel methodological developments.

\subsection{Analysis of the Results} 
\label{method:analysis_of_the_results}

The search on the Google Scholar platform retrieved 423 results, the search on Web of Science returned 208 matches, and the search on PubMed resulted in 110 articles. 
According to the selection criteria, 123 (29.1\,\%), 49 (23.6\,\%), and 45 (40.9\,\%) results were relevant for Google Scholar, Web of Science, and PubMed, respectively.
A representation of the relevant publications for the respective years is shown in Figure~\ref{fig:review_method:publications}. 
It can be seen that Google Scholar has the highest number of publications meeting the selection criteria in each year.
Furthermore, all included articles are already covered by Google Scholar, resulting in 123 publications taking all three search platforms into account.
A positive trend can be noticed across all search platforms until 2021, indicating a strong increase in research in this area in recent years.
Since this review only covers the first seven months of 2023, the expected number of publications by the end of the year is given on the top of the transparent boxes, assuming a linear increase in the number of papers.

\section{Public Datasets}
\label{datasets}

\begin{figure*}[tb]
\includegraphics[width=\textwidth]{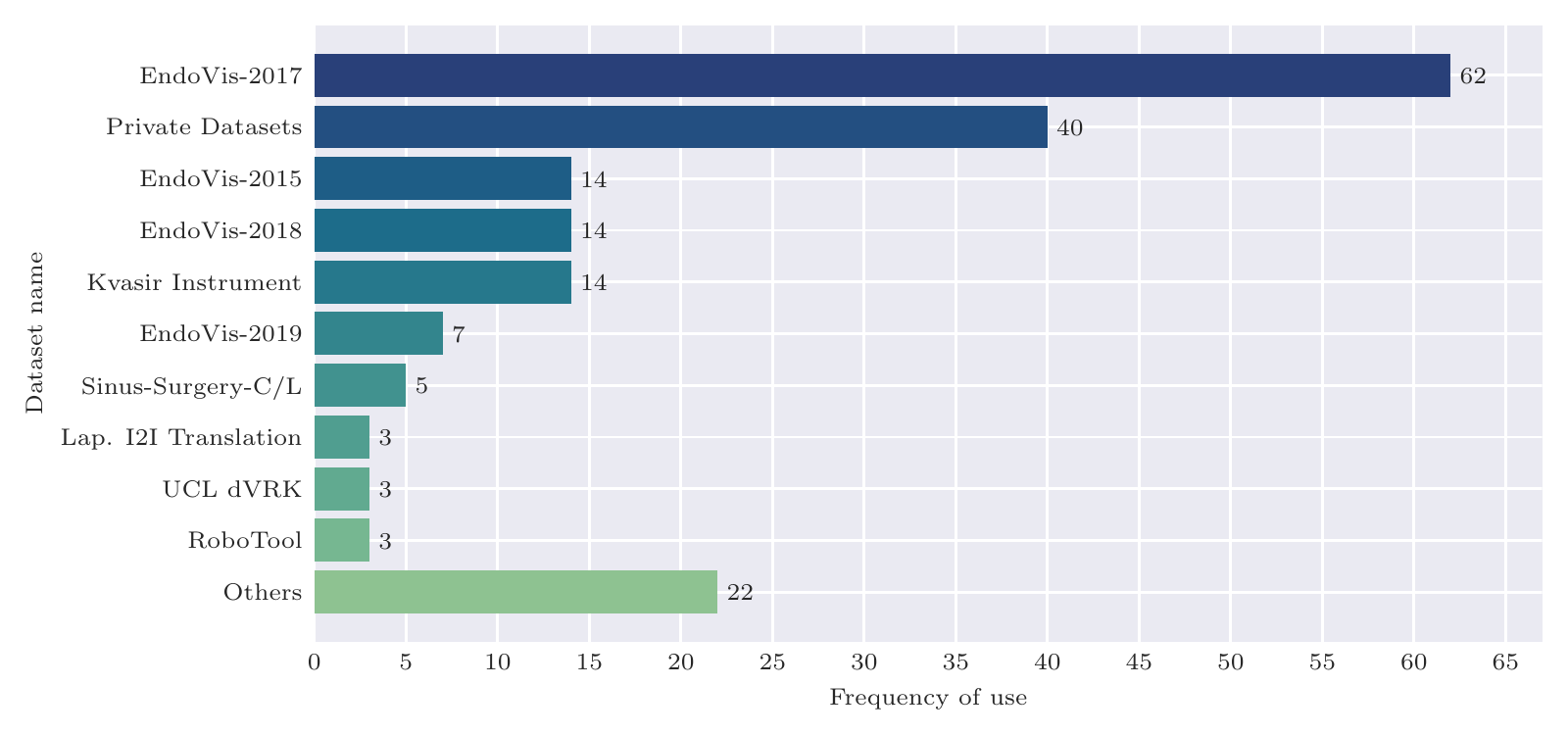}
\caption{Visualization of commonly used datasets along with their frequency of use in recent publications.}
\label{fig:datasets:usage}
\end{figure*}

\begin{table*}[tbh]
	\centering
	\caption{Publicly available endoscopic datasets together with their corresponding references (Ref.) and website links. The upper part contains datasets used in surveyed literature, while the lower part lists other freely available data that are available for the development of segmentation methods. The datasets are each sorted in ascending order by year of publication.}
	\begin{tabularx}{\textwidth}{l c p{12.5cm}}
		\toprule
		\multirow{1}{*}{Dataset} & \multicolumn{1}{c}{Ref.} & Website \\
		\midrule
		EndoVis 2015 & \cite{endovis_2015} & 
            \url{https://endovissub-instrument.grand-challenge.org} \\
		EndoVis 2017 & \cite{endovis_2017} & 
            \url{https://endovissub2017-roboticinstrumentsegmentation.grand-challenge.org} \\
		EndoVis 2018 & \cite{endovis_2018} & 
            \url{https://endovissub2018-roboticscenesegmentation.grand-challenge.org} \\
		EndoVis 2019 & \cite{ross2021comparative} & 
            \url{https://robustmis2019.grand-challenge.org/} \\
		Lap. I2I Translation & \cite{laparoscopic_I2I_translation} & 
            \url{http://opencas.dkfz.de/image2image} \\
		Sinus-Surgery-C/L & \cite{qin2020towards, lc_gan_image_to_image_translation} & 
            \url{http://hdl.handle.net/1773/45396} \\
		UCL dVRK & \cite{dvrk} &
            \url{https://www.ucl.ac.uk/interventional-surgical-sciences/weiss-open-research/weiss-open-data-server/ex-vivo-dvrk-segmentation-dataset-kinematic-data}  \\
		Kvasir Instrument & \cite{kvasir} &   
            \url{https://datasets.simula.no/kvasir-instrument} \\
		RoboTool & \cite{robo_tool} & 
            \url{https://www.synapse.org/\#!Synapse:syn22427422} \\
        \midrule
        InstrumentCrowd & \cite{maier2014can} & 
            \url{https://opencas.webarchiv.kit.edu/?q=InstrumentCrowd} \\
        NeuroSurgicalTools & \cite{bouget2015detecting} & 
            \url{https://medicis.univ-rennes1.fr/software} \\
        CholecSeg8k & \cite{Hong2020CholecSeg8kAS} & 
            \url{https://www.kaggle.com/datasets/newslab/cholecseg8k} \\
        HeiSurF & \cite{heisurf} & 
            \url{https://www.synapse.org/\#!Synapse:syn25101790/wiki/608802} \\
        ART-Net & \cite{hasan2021detection} & 
            \url{https://github.com/kamruleee51/ART-Net} \\
        CaDIS & \cite{cadis} & \url{https://cataracts-semantic-segmentation2020.grand-challenge.org} \\
        AutoLaparo & \cite{wang2022autolaparo} & 
            \url{https://autolaparo.github.io} \\
        SAR-RARP50 & \cite{Psychogyios2023SARRARP50} & \url{https://www.synapse.org/\#!Synapse:syn27618412/wiki/616881} \\
        \bottomrule
	\end{tabularx}
	\label{tab:datasets_references}
\end{table*}

\begin{table*}[tbh]
	\centering
	\caption{Properties of publicly available datasets. Indicated are the type of instruments, the resolution of the video images, the kind of procedure shown, whether the frames are in-vivo or ex-vivo recordings along with the type of tissue, and the size of the dataset consisting of the number of annotated images and the number of sequences. The upper part contains datasets used in surveyed literature, while the lower part lists other freely available data that are available for the development of segmentation methods. The datasets are each sorted in ascending order by year of publication.}
	\begin{tabular}{lccclllrr}
		\toprule
		\multirow{2}{*}{Dataset} & \multicolumn{2}{c}{Instruments} & & & \multicolumn{2}{c}{Data type} & \multicolumn{2}{c}{Size} \\
		\cmidrule(llrr){2-3} \cmidrule(llrr){6-7} \cmidrule(lr){8-9}
		& Rigid & Robotic & Resolution & Procedure & Ex-vivo & In-vivo & \#\,Images & \#\,Seq. \\
		\midrule
		EndoVis 2015 - Seg. & \ding{51} &  & 640 $\times$ 480 & laparoscopy  &  & human & 300 & 6 \\
            EndoVis 2015 - Track. & \ding{51} &  & 640 $\times$ 480 & laparoscopy &  & human & 360 & 6 \\
            
		EndoVis 2015 - Seg. &  & \ding{51} & 720 $\times$ 576 &  & porcine & & 9,000 & 6 \\
            EndoVis 2015 - Track. &  & \ding{51} & 720 $\times$ 576 &  & porcine & & 9,000 & 6 \\
		EndoVis 2017 &  & \ding{51} & 1920 $\times$ 1080 & abdominal &  & porcine & 3,000 & 10 \\
		EndoVis 2018 &  & \ding{51} & 1280 $\times$ 1024 & nephrectomy  &  & porcine & 2,831 & 19 \\
		EndoVis 2019 & \ding{51} &  & 960 $\times$ 540 & laparoscopy  &  & human & 10,040 & 30 \\
		Lap. I2I Translation & \ding{51} &  & 452 $\times$ 256 & laparoscopy &  & human & 20,000 & - \\
		Sinus-Surgery-C/L & \ding{51} &  & 240 $\times$ 240 & sinus surgery &  & human & 9,003 & 13 \\
		UCL dVRK &  & \ding{51} & 538 $\times$ 701 & animal tissue & various animals & & 4,200 & 20 \\
		\multirow{2}{*}{Kvasir Instrument} & \multirow{2}{*}{\ding{51}} &  & \vtop{\hbox{\strut $[720 \times 576$,}\hbox{\strut $1280 \times 1024]$}} & \vtop{\hbox{\strut gastroscopy \&}\hbox{\strut colonoscopy}}  &  & \multirow{2}{*}{human} & \multirow{2}{*}{590} & \multirow{2}{*}{-}  \\
		RoboTool &  & \ding{51} & varying & varying &  & human & 514 & 20 \\
        \midrule
        InstrumentCrowd & \ding{51} &  & 640 $\times$ 480 & laparoscopy &  & human & 120 & 6 \\
        NeuroSurgicalTools & \ding{51} &  & $612 \times 460$ & neurosurgery  &  & human & 2,476 & 14  \\
        CholecSeg8k & \ding{51} &  & 854 $\times$ 480 & laparoscopy &  & human & 8,080 & 17 \\
        \multirow{2}{*}{HeiSurF} & \multirow{2}{*}{\ding{51}} &  & \vtop{\hbox{\strut $[720 \times 576$,}\hbox{\strut $1920 \times 1080]$}} & \multirow{2}{*}{laparoscopy}  &  & \multirow{2}{*}{human} & \multirow{2}{*}{829} & \multirow{2}{*}{33}  \\
        ART-Net & \ding{51} &  & 1920 $\times$ 1080 & laparoscopy &  & human & 635 & 29 \\
        CaDIS & \ding{51} &  & 960 $\times$ 540 & cataract surgery &  & human & 4,670 & 25 \\
        AutoLaparo & \ding{51} &  & 1920 $\times$ 1080 & laparoscopy &  & human & 1,800 & 21 \\
        \multirow{2}{*}{SAR-RARP50} & & \multirow{2}{*}{\ding{51}} & \multirow{2}{*}{1920 $\times$ 1080} & radical & & \multirow{2}{*}{human} & \multirow{2}{*}{16,250} & \multirow{2}{*}{50} \\
         & & & & prostatectomy & & & &  \\
        \bottomrule
	\end{tabular}
	\label{tab:datasets_usages_tab}
\end{table*}

\begin{figure*}[tbh]
	\subfloat[\label{datasets:example_images_endovis15_rigid}]{ 
		\includegraphics[width=0.24\textwidth]{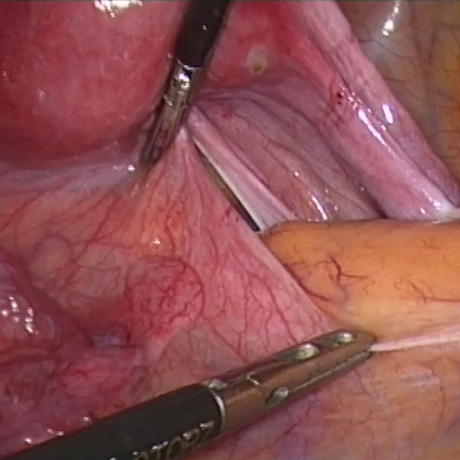}
    }
    \hfill
    \subfloat[\label{datasets:example_images_endovis17}]{
      	\includegraphics[width=0.24\textwidth]{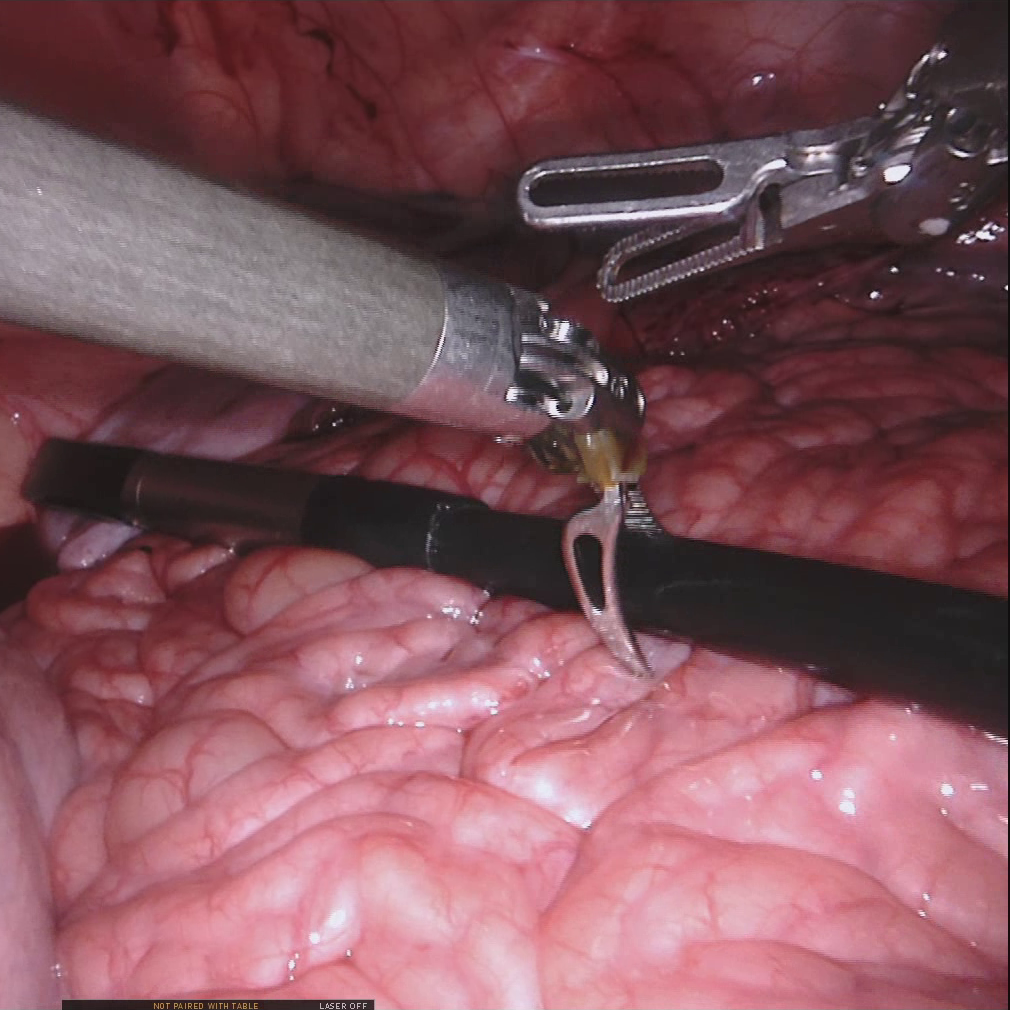}
    }
    \hfill
    \subfloat[\label{datasets:example_images_endovis18}]{
      	\includegraphics[width=0.24\textwidth]{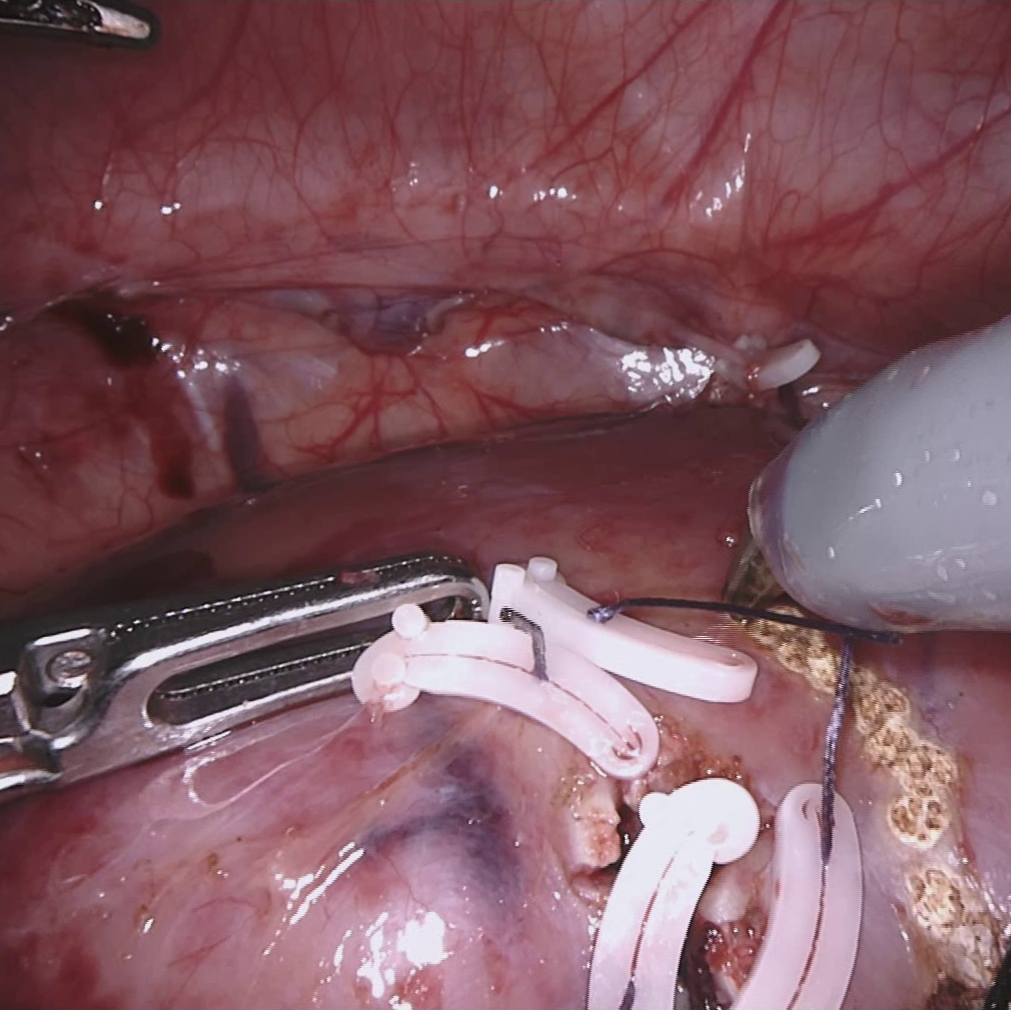}
    }
    \hfill
    \subfloat[\label{datasets:example_images_endovis19}]{
      	\includegraphics[width=0.24\textwidth]{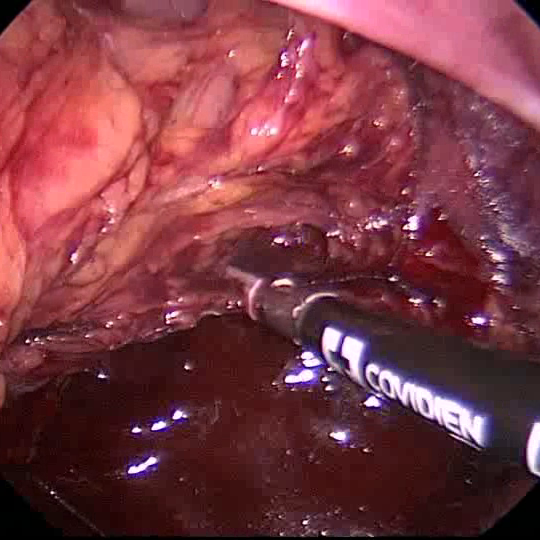}
    }
    \hfill
    \subfloat[\label{datasets:example_images_lap_i2i}]{
      	\includegraphics[width=0.19\textwidth]{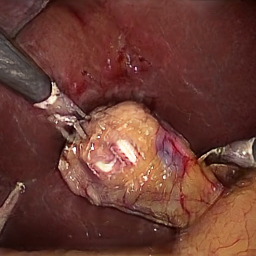}
    }
    \hfill
    \subfloat[\label{datasets:example_images_sinus_surgery}]{
      	\includegraphics[width=0.19\textwidth]{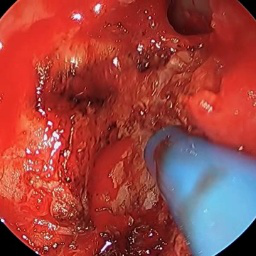}
    }
    \hfill
    \subfloat[\label{datasets:example_images_ucl_dvrk}]{
      	\includegraphics[width=0.19\textwidth]{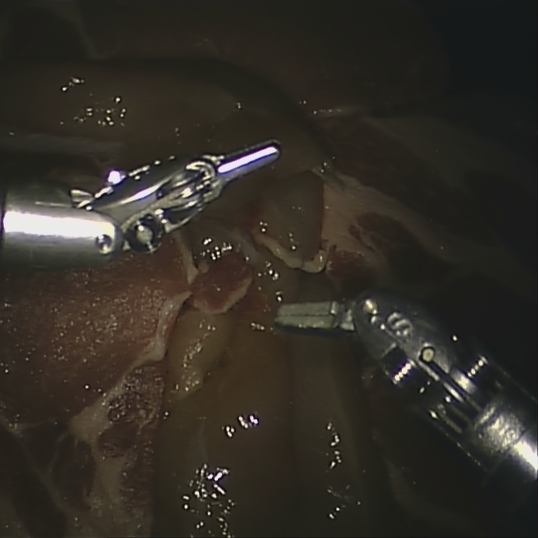}
    }
    \hfill
    \subfloat[\label{datasets:example_images_kvasir}]{
      	\includegraphics[width=0.19\textwidth]{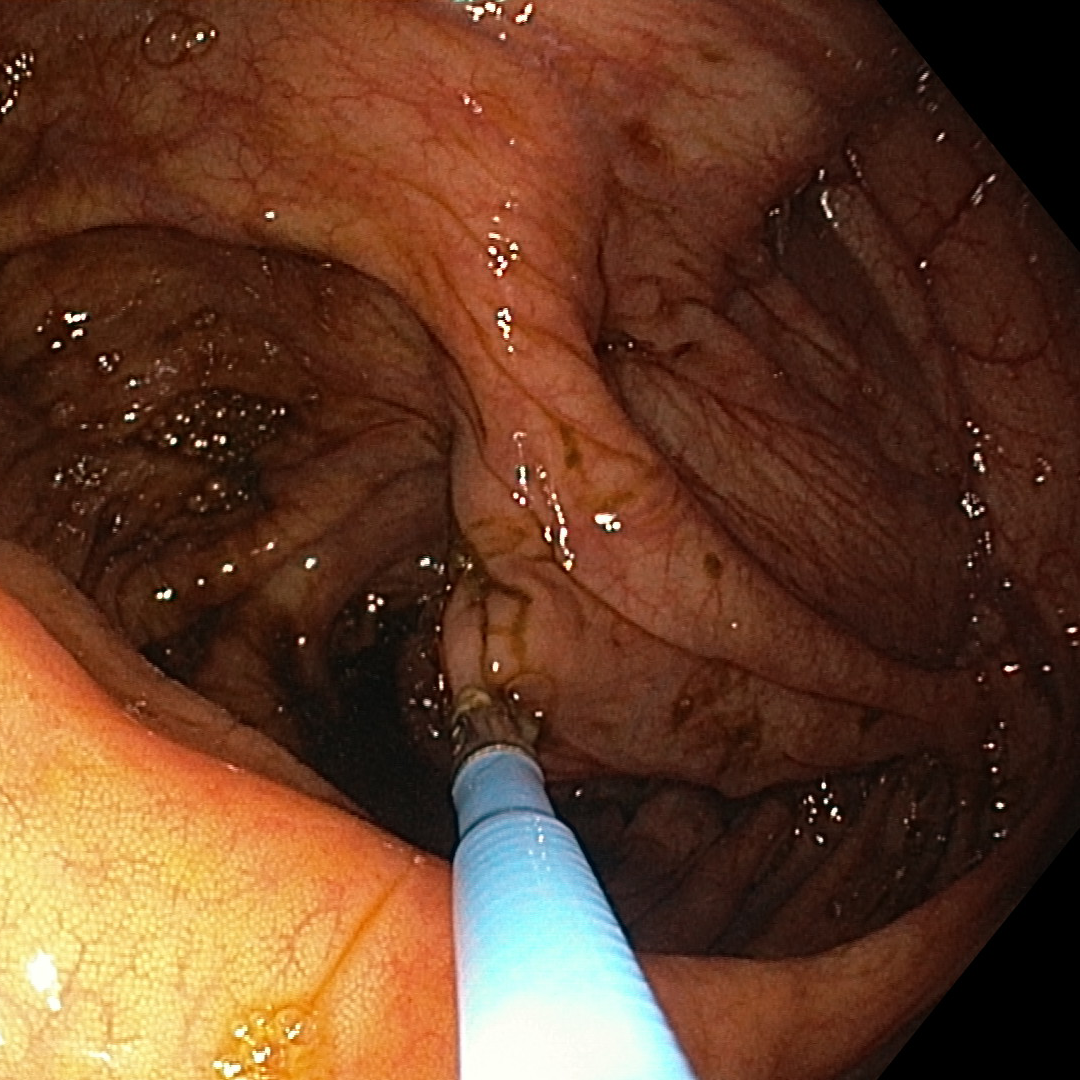}
    }
    \hfill
    \subfloat[\label{datasets:example_images_robotool_robotic}]{
      	\includegraphics[width=0.19\textwidth]{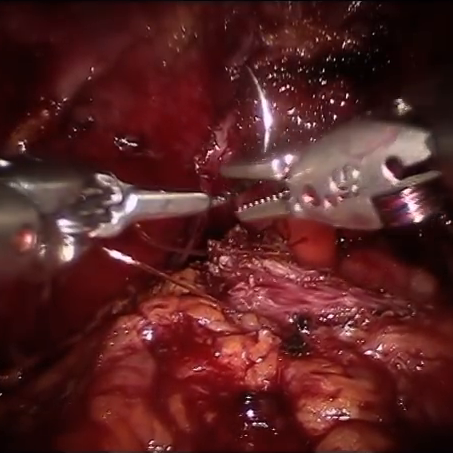}
    }
    \caption{Example images from frequently used endoscopic datasets, according to the upper part of Table~\ref{tab:datasets_usages_tab}. Images are taken from the EndoVis-2015~\sref{datasets:example_images_endovis15_rigid}, EndoVis-2017~\sref{datasets:example_images_endovis17}, EndoVis-2018~\sref{datasets:example_images_endovis18}, EndoVis-2019~\sref{datasets:example_images_endovis19}, Lap. I2I Translation~\sref{datasets:example_images_lap_i2i}, Sinus-Surgery-C/L~\sref{datasets:example_images_sinus_surgery}, UCL dVRK~\sref{datasets:example_images_ucl_dvrk}, Kvasir Instruments~\sref{datasets:example_images_kvasir}, and RoboTool~\sref{datasets:example_images_robotool_robotic} datasets.}
    \label{datasets:example_images}
\end{figure*}

The provision of freely available datasets had a central impact on the development of new methods concerning segmentation and tracking of surgical instruments in endoscopic images and videos in recent years~\cite{surgical_tool_datasets}. 
In the following, we present publicly available datasets for surgical instrument segmentation.
For this purpose, we first consider the datasets that are commonly used in recent work and in the publications included in this review, according to the selection criteria.
To provide a comprehensive overview of all available datasets, we further characterize datasets that are not used in the publications of this review but provide segmentations of surgical instruments and are available for the development of new methods.

Figure~\ref{fig:datasets:usage} shows a visualization regarding the frequency of use of the datasets employed in the considered papers.
A total of 187 datasets were used in the 123 publications considered, divided into nine relevant datasets, private non-public datasets, and datasets that did not meet the criteria for relevant datasets.
The criteria for a dataset to be considered relevant are that surgical instruments are present, that segmentation ground truths are available for these instruments, and that they consist of endoscopic images or videos.
All datasets that are not publicly available are grouped under the term ``Private Datasets''.
The term ``Others'' includes all datasets that are publicly available but do not meet the relevant criteria described above.
It can be seen that the EndoVis-2017 dataset~\cite{endovis_2017} was used in a total of 62 publications, making it by far the most frequently used dataset. 
This is followed by the group of private datasets used in a total of 40 publications. 
Ranked third, fourth, fifth, and sixth most frequently used datasets are the EndoVis-2015~\cite{endovis_2015}, EndoVis-2018~\cite{endovis_2018}, Kvasir Instrument~\cite{kvasir}, and EndoVis-2019~\cite{ross2021comparative} datasets, with 14, 14, 14, and seven utilizations, respectively. 
The Sinus Surgery-C/L dataset~\cite{qin2020towards, lc_gan_image_to_image_translation} was referenced seven times.
Three times each, the results of publications were based on the datasets labeled Laparoscopic Image-to-Image Translation~\cite{laparoscopic_I2I_translation}, UCL dVRK~\cite{dvrk}, and RoboTool~\cite{robo_tool}.
Furthermore, 22 publications utilized datasets that did not correspond to any of the above groups.

An overview of the datasets together with their corresponding references and website links can be seen in Table~\ref{tab:datasets_references}.
The upper part of the table contains datasets that have been used in the literature of this review and meet the above mentioned criteria.
The lower part describes datasets that also contain segmentations of surgical instruments, but either do not meet all of these criteria or were not used by any relevant publication in the time period considered.
Datasets are sorted in ascending order by the year of their publication.

Various characteristics of the datasets are given in Table~\ref{tab:datasets_usages_tab}. 
For each dataset, the type of instruments, the resolution of the video images, the type of procedure shown, whether the images are in-vivo or ex-vivo along with the type of tissue, and the size of the dataset consisting of the number of frames annotated and the number of sequences are listed. 
Unfortunately, no information is available if the datasets are retrospective or prospective in nature.
For the relevant datasets, which belong to the upper part of the table and are listed in Figure~\ref{fig:datasets:usage}, an example image for each dataset is shown in Figure~\ref{datasets:example_images}.


In the following, these nine relevant datasets with their characteristics are described in more detail, followed by the properties of the seven additional freely available datasets.
As in Tables~\ref{tab:datasets_references}, \ref{tab:datasets_usages_tab}, and Figure~\ref{fig:datasets:usage}, the datasets are each sorted in ascending order by the time of their publication.

\subsection{EndoVis 2015 - Instrument Segmentation and Tracking}
\label{datasets:endovis2015}

As part of the MICCAI 2015 conference, the Instrument Segmentation and Tracking sub-challenge was organized and conducted which can be divided into the two tasks of segmenting surgical instruments in single endoscopic images and tracking them over time in endoscopic image sequences.
The dataset includes in-vivo images of laparoscopic colorectal procedures using rigid instruments and frames from ex-vivo environments using robotic tools.

For rigid instrument segmentation, 160 annotated images are available for training, extracted from four sequences in equal numbers.
Ten additional images from each sequence and 50 frames from each of the two additional sequences are provided for validation purposes.
The tracking training dataset with rigid instruments consists of four sequences of 45 seconds each, of which one image per second is annotated.
For validation, one frame per second is provided for a further 15 seconds per sequence and two additional recordings of one minute each.
All in-vivo images have a resolution of 640 $\times$ 480. 
For segmentation, each pixel is assigned to one of the classes background, shaft, and manipulator, while for tracking the surgical tool, the coordinates of the center of the shaft end and the orientation of the instrument are also available.

For the segmentation of robotic instruments in ex-vivo images, four sequences of 45 seconds each are available for training, with each video frame annotated.
A further 15 seconds of recording from each of the four sequences and two one-minute videos are available for validation, with each frame annotated.
The video data of the tracking dataset regarding robotic instruments is identical to that of the segmentation, except that the annotations differ.
All ex-vivo images have a resolution of 720 $\times$ 576.
For segmentation, each pixel is assigned to one of the classes background, shaft, head, and clasper, while for surgical tool tracking, posture information in the form of rotation, translation, and articulation of the instrument head and claspers is provided. 

\subsection{EndoVis 2017 - Robotic Instrument Segmentation}
\label{datasets:endovis2017}

Two years after the first MICCAI sub-challenge focused on the segmentation of surgical tools in endoscopic images, the Robotic Instrument Segmentation sub-challenge was conducted in 2017 with a similar objective. 
The data provided for this sub-challenge deals with abdominal porcine procedures performed exclusively with robotic instruments and recorded by a da Vinci Xi system. 
Eight video sequences are available for training, each with 225 annotated frames, with one image annotated per second. 
An additional 75 video frames for each sequence and two additional sequences of 300 images each are provided for validation. 
All video frames have a resolution of 1920 $\times$ 1080 and represent in-vivo recordings. 
The segmentation distinguishes between the instruments large needle driver, prograsp forceps, monopolar curved scissors, cadiere forceps, bipolar forceps, vessel sealer, and a drop-in ultrasound probe. 
Furthermore, each instrument is divided into three parts. 

\subsection{EndoVis 2018 - Robotic Scene Segmentation}
\label{datasets:endovis2018}

As part of the MICCAI 2018 conference, the Robotic Scene Segmentation dataset was released, with video sequences showing porcine surgical procedures.
The dataset consists of 19 video sequences, 15 for training and four for validation.
As with the EndoVis 2017 dataset, only robotic instruments are visible over porcine tissue, and the recordings were captured using the da Vinci X and da Vinci Xi System.
The videos have a resolution of 1280 $\times$ 1024 and were taken with a stereo endoscope.
In addition to the instrument classes and their subdivisions of the different instrument parts used in the Robotic Instrument Segmentation sub-challenge of 2017, suturing needles, suturing thread, suction-irrigation devices, and surgical clips were segmented. 
In addition to these non-biological categories, the kidney parenchyma, the kidney fascia, and perinephric fat, which the authors called ``covered kidney'' and small intestine were also annotated.

\subsection{EndoVis 2019 - Robust Medical Instrument Segmentation}
\label{datasets:endovis2019}

The Robust Medical Instrument Segmentation (ROBUST-MIS) dataset was published in 2019. 
The aim was to investigate the robustness and generalization ability of different algorithms developed within the sub-challenge and to identify particular challenges in the images. 
For this purpose, the authors provided a total of 30 surgical procedures, representing in equal parts rectal resection procedures, proctocolectomy procedures, and sigmoid resection procedures, containing 10,040 annotated images. 
The recordings feature real-life in-vivo human surgeries performed with rigid instruments. 
The videos were recorded in high definition (HD) quality, downscaled and provided to challenge participants at a resolution of 960 $\times$ 540. 
The performance of the developed procedures in terms of generalization ability and quality is investigated in three stages, with increasing complexity.
In stage one, test data was taken from the procedures (patients) from which the training data were extracted. 
In stage two, test data was obtained from the same type of surgery as the training data but from procedures (patients) not included in the training. 
Finally, in stage three, test data originates from a different but similar type of surgery (and different patients) compared to the training data. 
The instruments are assigned an instance class with incremental numbering, which means that as soon as a new instrument appears in an image, it is labeled a number one higher than the previous highest category. 
As examples of instrument categories, the authors name grasper, scalpel, (transparent) trocar, clip applicator, hooks, stapling device, and suction.

\subsection{Laparoscopic Image-to-Image (I2I) Translation}
\label{datasets:i2i_translation}

The Laparoscopic Image-to-Image (I2I) Translation dataset contains 20,000 automatically annotated synthetic images generated by 3D laparoscopic simulations created from the CT scans of ten patients.
The images include a rendered view of laparoscopic scenes captured under random camera positioning.
The authors trained a translation network and subsequently translated all rendered images with five randomly drawn style vectors for each image, resulting in a total of 100,000 images. 
Furthermore, all images were translated with five styles from the Cholec80 \cite{twinanda2016endonet} dataset, which also shows images from in-vivo laparoscopic videos, resulting in another dataset of 100,000 images as well, representing the visual features of the Cholec80 dataset. 
All images are provided at a resolution of 452 $\times$ 256.
Included in the laparoscopic images are the two conventional surgical instruments grasper and hook, with the shafts and tips of the two instruments annotated as one instrument subclass each.

\subsection{Sinus-Surgery-C/L}
\label{datasets:uw_sinus}

The Sinus-Surgery-C/L dataset consists of two subsets, the Sinus-Surgery-C dataset, which shows surgical procedures on cadaver specimens, and the Sinus-Surgery-L dataset, which contains real procedures performed on living humans. 
For the Sinus-Surgery-C dataset, ten surgical procedures were recorded from five different cadaver specimens, ranging in duration from five to 23 minutes. 
The videos were recorded at a framerate of 30 FPS and a resolution of 320 $\times$ 240, and one frame was annotated every two seconds.
The authors provide 4,345 annotated images cropped to a size of 240 $\times$ 240, of which 3,606 contain instruments.
The Sinus Surgery-L dataset includes only three surgical procedures on three different patients, with a total duration of 2.5 hours. 
The recording and creation of the annotations followed the same procedure as for the Sinus-Surgery-C dataset. 
A total of 4,658 annotated video frames are provided, of which 4,344 contain instruments.
The dataset creators note that the Sinus-Surgery-L dataset is more complicated than the Sinus-Surgery-C dataset due to challenges such as multiple instrument types, sometimes insufficient visualization, and body excretions that can alter the scene. 
In both datasets, the same micro-debrider instrument was used, and the segmentation masks contain information for each pixel whether it belongs to the instrument used or to the background.

\subsection{UCL dVRK}
\label{datasets:ucl_dvrk}

The dataset entitled ``Ex-vivo dVRK (da Vinci Research Kit) segmentation dataset with kinematic data'' provides both segmentation masks and kinematic data for 14 surgical procedures with 300 video frames each.
The authors first perform movements with surgical instruments recorded by the dVRK system. 
These movements are then placed over tissues from different types of animals and over a green screen. 
The segmentation masks of the instruments are then determined by subtracting the greenscreen background. 
The authors freely provide all the data used to generate the dataset, including the kinematic data of the motions of the instruments, the motion images over the greenscreen background, and the final movements of the tools over tissues along with the associated segmentation masks.
For the development of own methodologies with the help of the dataset, the first eight sequences should be used for training, the two following recordings for validation, and videos 11 - 14 for testing. 
The images were captured at an original resolution of 720 $\times$ 576 and cropped to 538 $\times$ 701.
The only instrument in the recordings is the EndoWrist Large Needle Driver, which may appear more than once in an image. 
The segmentation masks contain the information for each pixel, whether it belongs to an instrument or the background, which means that a binary segmentation is performed.

\subsection{Kvasir Instrument}
\label{datasets:kvasir}

The dataset ``Kvasir-Instrument: Diagnostic and Therapeutic Tool Segmentation Dataset in Gastrointestinal Endoscopy'' was released in 2021 and provides annotations for 590 endoscopic frames acquired in gastroscopies and colonoscopies. 
The images include diagnostic and therapeutic tools and have different resolutions between 720 $\times$ 576 and 1280 $\times$ 1024. 
The authors provide both binary segmentation masks and bounding boxes. 
Examples of GI instruments in the dataset include snares, balloons, and biopsy forceps. Examples of GI instruments in the dataset include snares, balloons, and biopsy forceps.

\subsection{RoboTool}
\label{datasets:robotool}

For the creation of the RoboTool dataset, 514 video frames were manually annotated from 20 publicly available surgical procedures.
The recordings show exclusively robotic instruments, and the resolution of the video frames is based on the respective source video recordings.
There exist only binary labels indicating whether a given pixel belongs to an instrument or the background.
In addition, the authors provide images that can be used to create a synthetic dataset.
This consists of 14,720 foreground images showing between one and three surgical robotic instruments (of a total of 17 different types) in front of a green screen, of which 13,613 images have a resolution of 4032 $\times $ 3024 and the remaining 3360 $\times $ 2240.
For each foreground image, the binary segmentation mask of the instruments is provided. 
Furthermore, 6,130 background images with different resolutions showing human tissue from 50 publicly available surgical frames in which no instruments are present are provided.
The authors also supply the source code for blending the foreground images onto the background ones.

\subsection{Further Datasets}
\label{datasets:other}

In addition to these regularly used datasets identified in the recent literature, several others are available for the development of novel segmentation methods, which are listed along with their respective characteristics in the lower part of Table~\ref{tab:datasets_usages_tab}.

In their work from 2014, Maier-Hein et al.~\cite{maier2014can} investigate whether the annotation quality of a large number of anonymized non-experts is comparable to that of medical experts.
For this purpose, the authors created and published a dataset consisting of six surgical procedures with a resolution of $640 \times 480$, three each of laparoscopic adrenalectomies and laparoscopic pancreatic resections.
Twenty frames were extracted from each procedure, ten of which were annotated by medical experts.
Ten non-expert annotations are available for all 120 available images.
A separate segmentation mask exists for each instrument within an image, resulting in a total of 2,350 segmented instruments.

The NeuroSurgicalTools dataset~\cite{bouget2015detecting} contains 2,476 annotated images with binary segmentations of surgical instruments during tumor removal. 
Fourteen sequences were acquired with a neurosurgical microscope and the resolution of all videos was scaled to $612 \times 460$.
Instruments in the dataset included suction tubes, bipolar forceps, retractors, hooks, scalpels, pliers, and scissors, resulting in 3,819 segmented individual tools.

To use the Cholec80 dataset~\cite{twinanda2016endonet} also for segmentation tasks, Hong et al. published the CholecSeg8k dataset~\cite{Hong2020CholecSeg8kAS} in 2020. 
The authors extracted and labeled 8,080 laparoscopic cholecystectomy images from 17 video recordings with a resolution of $854 \times 480$.
Thirteen classes were divided, two of which were laparoscopic instruments, namely the L-hook electrocautery and grasping forceps.

As part of the MICCAI 2021 conference, the ``HeiChole Surgical Workflow Analysis and Full Scene Segmentation (HeiSurF)'' challenge~\cite{heisurf} was conducted and the dataset used for it was published.
The represented surgeries are laparoscopic gallbladder resections.
The focus is on full scene segmentation and surgical workflow analysis, for which the actions performed and the current phase of an operation is annotated in addition to the ground truth of the segmentation.
The training data consists of two parts.
In the first part, images are annotated at two-minute intervals from 24 procedures, while the second part consists of short sequences of each video annotated at 1 FPS. 
The test data consists of nine unpublished surgeries.
The resolution of the recordings varies from $720 \times 576$ to $1920 \times 1080$. 
In addition to the categories for tissues and organs, there is a global instrument class, a class for drains, clips, one for trocars, and a class for specimen bags.

\definecolor{custom_blue}{HTML}{1D5D86}
\definecolor{custom_green}{HTML}{60AB90}

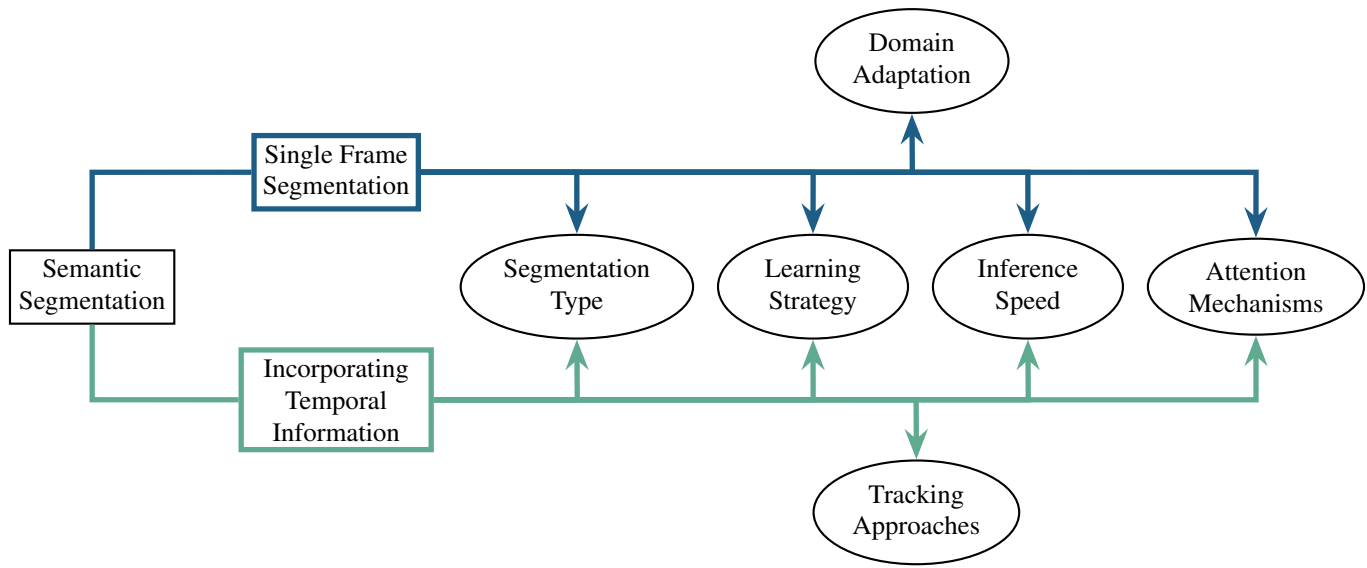
\begin{figure*}[tb]
\centering

\begin{tikzpicture}[node distance={0.1cm}, thick, main/.style = {draw, rectangle}] 
    \node[main, align=center] (1) {Semantic\\Segmentation}; 
    \node[main, line width=0.07cm, align=center, minimum width=2.5, draw=custom_blue] (2) [above right=0.5cm and 1cm of 1] {Single Frame\\Segmentation}; 
    \node[main, line width=0.07cm, align=center, minimum width=2.5cm, draw=custom_green] (3) [below=1.8cm of 2] {Incorporating\\Temporal\\Information}; 
    \node[ellipse, draw, align=center, minimum width=2.5cm] (5) [below right=0.5cm and 0.96cm of 2] {Segmentation\\Type}; 
    \node[ellipse, draw, align=center, minimum width=2.5cm] (6) [right=0.3cm of 5] {Learning\\Strategy}; 
    \node[ellipse, draw, align=center, minimum width=2.5cm] (7) [above right=2.0cm and -0.5cm of 6] {Domain\\Adaptation}; 
    \node[ellipse, draw, align=center, minimum width=2.5cm] (8) [below right=2.0cm and -0.5cm of 6] {Tracking\\Approaches};
    \node[ellipse, draw, align=center, minimum width=2.5cm] (9) [right=0.3cm of 6] {Inference\\Speed}; 
    \node[ellipse, draw, align=center, minimum width=2.5cm] (10) [right=0.3cm of 9] {Attention\\Mechanisms}; 
    \draw[line width=0.07cm, custom_blue] (1.north) -- (0,1.5209) -- (2.west);
    \draw[line width=0.07cm, custom_green] (1.south) -- (0,-1.495) -- (3.west);

    \draw [line width=0.07cm, custom_blue, -{Stealth[length=4mm]}] (2.east) -- (6.377,1.5209) -- (5.north); 
    \draw [line width=0.07cm, custom_blue, -{Stealth[length=4mm]}] (2.east) -- (9.487,1.5209) -- (6.north);
    \draw [line width=0.07cm, custom_blue, -{Stealth[length=4mm]}] (2.east) -- (10.792,1.5209) -- (7.south);
    \draw [line width=0.07cm, custom_blue, -{Stealth[length=4mm]}] (2.east) -- (12.312,1.5209) -- (9.north);
    \draw [line width=0.07cm, custom_blue, -{Stealth[length=4mm]}] (2.east) -- (15.314,1.5209) -- (10.north);
    
    \draw [line width=0.07cm, custom_green, -{Stealth[length=4mm]}] (3.east) -- (6.383, -1.4959) -- (5.south);
    \draw [line width=0.07cm, custom_green, -{Stealth[length=4mm]}] (3.east) -- (9.487, -1.4959) -- (6.south);
    \draw [line width=0.07cm, custom_green, -{Stealth[length=4mm]}] (3.east) -- (10.842, -1.4959) -- (8.north);
    \draw [line width=0.07cm, custom_green, -{Stealth[length=4mm]}] (3.east) -- (12.317, -1.4959) -- (9.south);
    \draw [line width=0.07cm, custom_green, -{Stealth[length=4mm]}] (3.east) -- (15.314, -1.4959) -- (10.south); 

    \end{tikzpicture}

    \caption{Structure of Section~\ref{semantic_segmentation_methods} regarding semantic segmentation methods, to be read from left to right. Indicated are the topics by which the identified publications are grouped. The methods of both single-frame segmentation (Section \ref{semantic_segmentation:single_frame_seg}) and those involving temporal information (Section \ref{semantic_segmentation:incorporating_temporal_information}) are divided by segmentation type~(Sections \ref{semantic_segmentation:single_frame_seg:seg_type} and \ref{semantic_segmentation:temporal_info:segmentation_type}), learning strategy~(Sections \ref{semantic_segmentation:single_frame_seg:learning_strategy} and \ref{semantic_segmentation:temporal_info:learning_strategy}), inference speed~(Sections \ref{semantic_segmentation:single_frame_seg:inference_speed} and \ref{semantic_segmentation:temporal_info:inference_speed}), and attention mechanisms~(Sections \ref{semantic_segmentation:single_frame_seg:attention_mechanisms} and \ref{semantic_segmentation:temporal_info:attention_mechanisms}). For single-frame segmentation, methods of domain adaptation~(\ref{semantic_segmentation:single_frame_seg:domain_adaptation}) are further presented, and for publications with temporal information tracking approaches~(Section \ref{semantic_segmentation:temporal_info:domain_adaptation}) are explained.}
    \label{semantic_seg:structure} 
\end{figure*}

The ``Augmented Reality Tool Network (ART-Net)'' dataset~\cite{hasan2021detection} consists of 29 procedures representing laparoscopic hysterectomies performed with non-robotic instruments.
In addition to binary segmentation of surgical instruments, tool presence, and instrument-specific geometric primitives are also annotated.
All recordings correspond to a resolution of $1920 \times 1080$.

The ``Cataract Dataset for Image Segmentation (CaDIS)''~\cite{cadis} is a collection of 25 video recordings and 4,670 annotated frames showing cataract surgery that was used in a MICCAI challenge in 2020.
Three tasks are described for which annotations are provided with different granularity, i.e., different instruments and tissue types are distinguished in more detail.
There are 29 different instrument classes, 4 anatomy classes, and 3 categories for different objects appearing in the scene.
All images have been scaled to a resolution of $960 \times 540$.

The AutoLaparo dataset~\cite{wang2022autolaparo} provides sub-datasets for the tasks workflow recognition, laparoscope motion prediction, and instrument and key anatomy segmentation.
A total of 21 videos of laparoscopic hysterectomies are provided, recorded at a resolution of $1920 \times 1080$.
In addition to the anatomy-based \pagebreak segmentation of the uterus, four types of instruments are distinguished in the recordings, namely grasping forceps, ligasure, dissecting and grasping forceps, and electric hook.

The dataset for performing the ``SAR-RARP50: Segmentation of surgical instrumentation and Action Recognition on Robot-Assisted Radical Prostatectomy Challenge''~\cite{Psychogyios2023SARRARP50} in 2022 consists of 50 procedures with a total of 16,250 annotated images showing radical prostatectomy performed with the DaVinci Si system. All images were captured at a resolution of $1920 \times 1080$ pixels, and annotations are available for the recognition of surgical actions as well as for the semantic segmentation of surgical instruments. The frames were labeled at a frequency of 1 FPS, and nine categories were used for semantic segmentation, divided into six non-tool objects and three part-level categories for shaft, wrist, and instrument claspers.

The freely available datasets described are limited by the fact that only those were selected that contain annotations for surgical instrument segmentation.
For a detailed overview of datasets available for other tasks in the surgical context, such as action classification, phase recognition, surgical skill assessment, or segmentation of anatomical structures, we refer to the works of Maier-Hein et al.~\cite{maier2022surgical} and Rodrigues et al.~\cite{surgical_tool_datasets}.

\section{Semantic Segmentation Methods}
\label{semantic_segmentation_methods}

The following presents current segmentation approaches, divided into two chapters and several subchapters, as shown in Figure~\ref{semantic_seg:structure}.
First, contributions to the segmentation of single images are described, followed by approaches that address the segmentation of images incorporating temporal information from successive video frames.

\subsection{Single Frame Segmentation}
\label{semantic_segmentation:single_frame_seg}

This section deals exclusively with publications focusing on the semantic segmentation of single images, for which 72 contributions were identified.
In the following these are characterized and logically grouped according to different characteristics. 
First, the type of segmentation is examined, i.e., work on binary, instrument type-based, and instrument part-based segmentation is reviewed.
Each of these three types has different requirements and challenges, and the segmentation of instrument parts is usually more complex than the segmentation of pure instruments, which in turn is more difficult to evaluate than the binary distinction between instruments and background.
The developments are then organized according to the following learning strategies: supervised learning, semi-supervised learning, weakly-supervised learning, and work that does not fit into any of these categories.
We further provide methods that focus on the adaptation to new domains.
Following this, we present contributions that focus on developing real-time or near-real-time approaches and explicitly state the processing speed of their methods for this purpose.
Finally, we give an overview of recent developments incorporating attention-based techniques into network architectures and highlight how this methodology improves segmentation quality.

\subsubsection{Segmentation Type}
\label{semantic_segmentation:single_frame_seg:seg_type}

\begin{figure}[!b]
\includegraphics[width=0.49\textwidth]{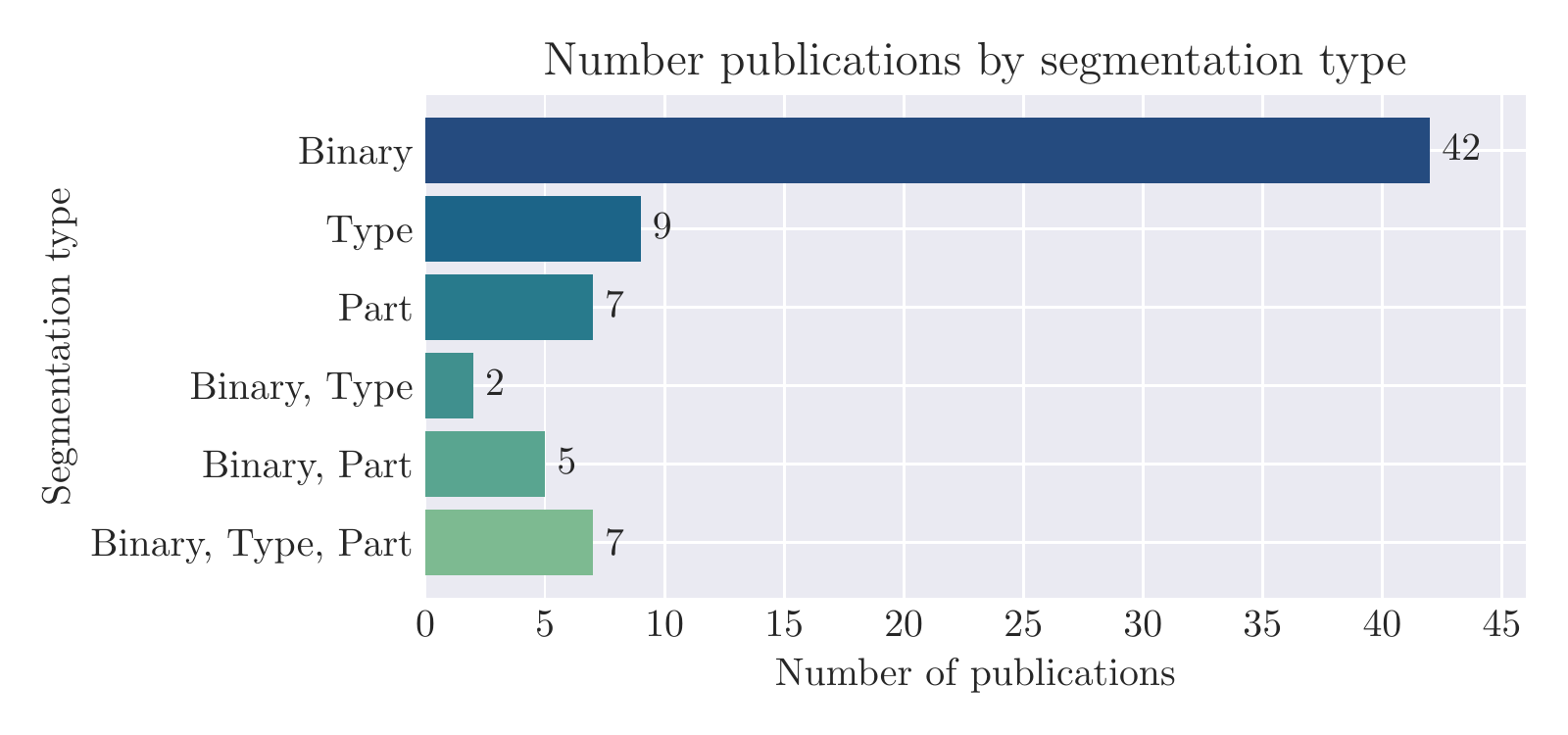}
\caption{Number of relevant publications per segmentation type for semantic single image segmentation.}
\label{fig:seg_types_sem_seg_single_frame}
\end{figure}

The number of relevant publications per segmentation type is visualized in Figure~\ref{fig:seg_types_sem_seg_single_frame}, where only those segmentation types and combinations are listed that are used in at least one paper.
Table~\ref{tab:seg:semantic_seg:single_frame_seg_binary} shows all works dealing with binary segmentation of instruments, which corresponds to the largest group.
Approaches focusing on other segmentation types are presented in Table~\ref{tab:seg:semantic_seg:single_frame_other_types}, subdivided according to the categories shown in Figure~\ref{fig:seg_types_sem_seg_single_frame}.
For each article the supervision strategy, a binary indicator of whether attention mechanisms were employed in the methodology, and the datasets used are provided.
Methods developed using a supervised approach are labeled SV, unsupervised developed methods are marked USV, weakly supervised approaches are identified as WE, semi-supervised methods using SE, methods dealing with the generation and/or use of synthetic data are indicated as SN and methods dealing with the task of domain adaptation are labeled DA.
When specifying the datasets used, the meaning of the columns ``Private'' and ``Other'' is identical to the one described in Chapter~\ref{datasets}.
If a publication uses several datasets that are not publicly available and thus belong to the ``Private'' category or uses several datasets that were not recorded endoscopically or not contain surgical instruments and thus belong to the ``Other'' category, the corresponding column is marked only once.

\subsubsection{Learning Strategies}
\label{semantic_segmentation:single_frame_seg:learning_strategy}

\begin{figure}[!b]
\centering
\includegraphics[width=0.40\textwidth]{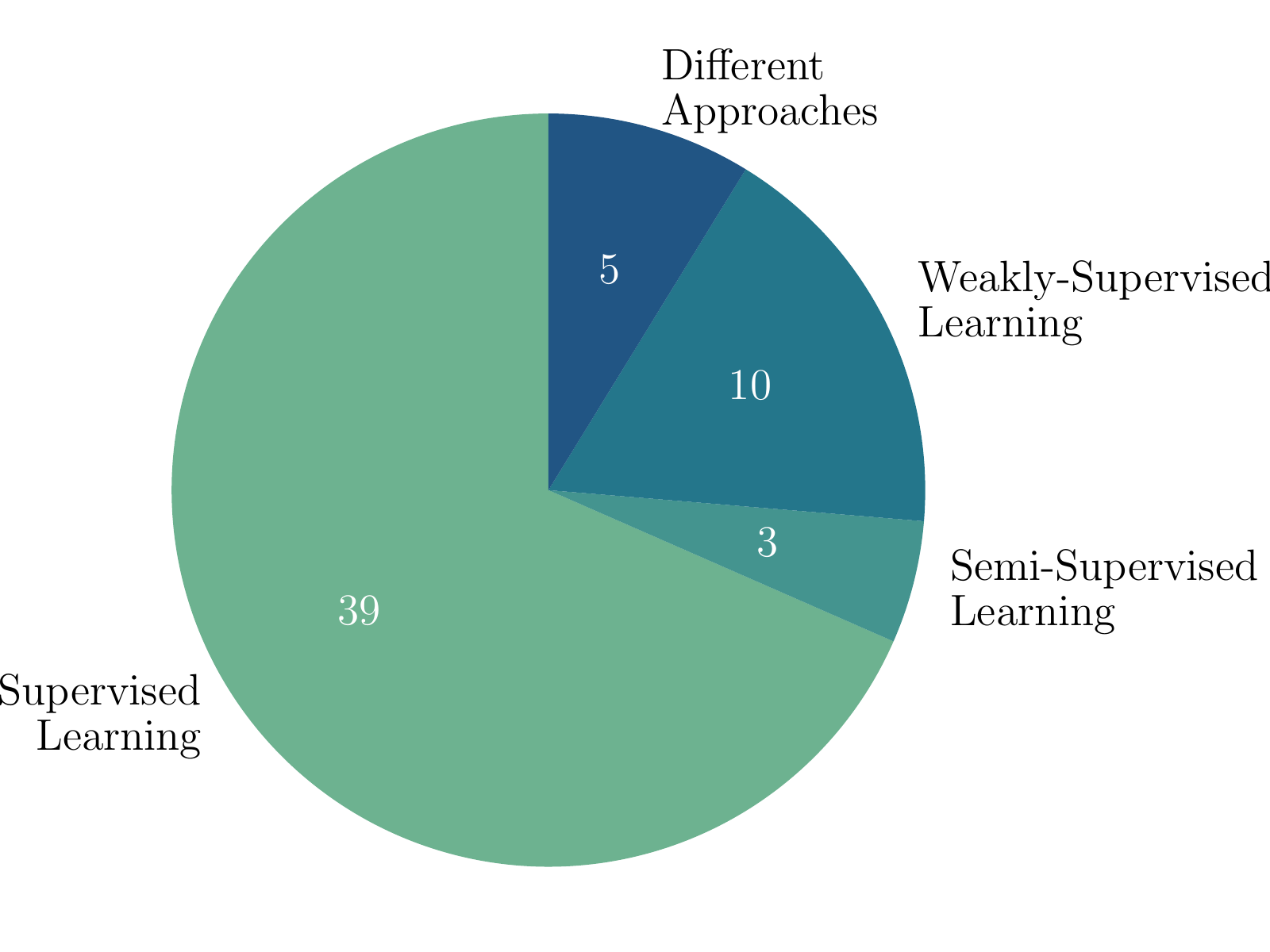}
\caption{Amount of relevant papers for single image semantic segmentation divided according to the respective learning strategies used.}
\label{fig:learning_types_sem_seg_single_frame}
\end{figure}

The following is a review of the works identified concerning the learning strategies used.
For this purpose, they are divided into supervised, semi-supervised, weakly supervised, and other types of learning methods. 
An overview of the number of papers according to each learning method is presented in Figure~\ref{fig:learning_types_sem_seg_single_frame}. 

\subsubsection*{Supervised Learning}

\begin{table*}[tbhp]
	\centering
	\caption{Publications that contributed in the area of binary semantic single-image segmentation of surgical instruments. For each publication, the type of methodology (SV = supervised, USV = unsupervised, WE = weakly-supervised, SE = semi-supervised, SN = synthetic data, DA = domain adaptation), whether or not attention mechanisms (Att.) were used in the methodology, and the datasets used are indicated.}
	\begin{tabularx}{\textwidth}{p{5.0cm}p{1.8cm}P{0.4cm}P{0.3cm}P{0.3cm}P{0.3cm}P{0.3cm}P{0.4cm}P{0.7cm}P{0.7cm}P{0.7cm}P{0.7cm}P{0.5cm}P{0.5cm}}
		\toprule
		\multirow{3}{*}{Publication}
		& \multirow{3}{*}{\begin{tabular}{l}\\Super- \\ vision\end{tabular}}
		& \multirow{3}{*}{Att.}
		& \multicolumn{10}{c}{Used Datasets} \\
		\cmidrule(llrr){4-14}
		&
		& 
		& \multicolumn{4}{c}{EndoVis}
		& \multirow{2}[3]{*}{\shortstack{Lap. \\ I2I}}
		& \multirow{2}[3]{*}{\shortstack{Sinus \\ Surg.}}
		& \multirow{2}[3]{*}{\shortstack{UCL \\ dVRK}}
		& \multirow{2}[3]{*}{\shortstack{Kvasir \\ Inst.}}
		& \multirow{2}[3]{*}{\shortstack{Robo- \\ Tool}}
		& \multirow{2}[3]{*}{\shortstack{Pri- \\ vate}}
		& \multirow{2}[3]{*}{\shortstack{Ot- \\ her}} \\
		\cmidrule(llrr){4-7}
		&
		& 
		& 15
		& 17
		& 18
		& 19
		& & & & & & & \\
		\midrule
            Banik et al. (2021) \cite{banik2021net} & SV & & & & & & & & & \ding{51} & & &  \\
            Chou (2021)\cite{chou2021automatic} & SV & \ding{51} & & & & & & & & \ding{51} & & &  \\	
		da Costa Rocha et al. (2019) \cite{da2019self} & WE & & & & & & & & & & & \ding{51} &  \\
            Devi et al. (2022) \cite{devi2022msdfn} & SV & & & & & & & & & \ding{51} & & &  \\	
            Galdran (2021) \cite{galdran2021polyp} & SV & & & & & & & & & \ding{51} & & &  \\
	    Garcia-P.-H. et al. (2017) \cite{garcia2017toolnet} & SV & & \ding{51} & & & & & & & & & &  \\
		Garcia-P.-H. et al. (2021) \cite{garcia2021image} & SN & & & \ding{51} & & & & & & & \ding{51} & &  \\
		Hasan et al. (2021) \cite{hasan2021detection} & SV & & & \ding{51}  & & & & & & & & \ding{51} &  \\
		Hasan et al. (2021) \cite{hasan2021segmentation} & SV & & \ding{51} & & & & & & & & & &  \\
		Heredia Perez et al. (2020) \cite{heredia2020effects} & SN & & & & & & & & & & & \ding{51} &  \\
		Huang et al. (2022) \cite{huang2022simultaneous} & WE & & & & & & & & & & & \ding{51} &  \\
            Huang et al. (2022) \cite{huang2022surgical} & SV & & & & & & & \ding{51} & & & & &  \\
		Jha et al. (2021) \cite{jha2021exploring} & SV & \ding{51} & & & & \ding{51} & & & & & & &  \\
		Kalavakonda et al. (2019) \cite{kalavakonda2019autonomous} & SV & & & \ding{51} & & & & & & & & & \ding{51} \\
		Kalia et al. (2021) \cite{kalia2021co} & DA & & & \ding{51} & & & & & \ding{51} & & & &  \\
		Keprate and Pandey (2021) \cite{keprate2021kvasir} & SV & & & & & & & & & \ding{51} & & &  \\
		Lee et al. (2019) \cite{lee2019weakly} & WE & & & & & & & & & & & \ding{51} &  \\
		Lee et al. (2019) \cite{lee2019segmentation} & WE & & & \ding{51} & & & & & & & & &  \\
		Leifman et al. (2022) \cite{leifman2022pixel} & WE,~SN,~DA & & & \ding{51} & & \ding{51} & & & & & \ding{51} & &  \\
            Lou et al. (2023) \cite{lou2023min} & SE & & & \ding{51} & & & & & & \ding{51} & & \ding{51} & \\
		Ni et al. (2019) \cite{ni2019rasnet} & SV & \ding{51} & & \ding{51} & & & & & & & & &  \\
		Pakhomov et al. (2020) \cite{pakhomov2020towards} & WE & & & \ding{51} & & & & & & & & &  \\
		Papp et al. (2022) \cite{papp2022surgical} & WE & & & \ding{51} & & & \ding{51} & & & & & &  \\
		Psychogyios et al. (2022) \cite{psychogyios2022msdesis} & SV, DA & & & \ding{51} & & & & & & & & & \ding{51} \\
		Qin et al. (2019) \cite{qin2019surgical} & USV, SV & & & & & & & & & & & \ding{51} &  \\
		Qin et al. (2020) \cite{qin2020towards} & SV & & & \ding{51} & & & & \ding{51} & & & & &  \\
		Rajak and Mirza (2021) \cite{rajak2021segmentation} & SV & & & & & & & & & \ding{51} & & &  \\
		Sahu et al. (2020) \cite{sahu2020endo} & SN,~DA & & \ding{51} & & & & \ding{51} & & & & & \ding{51} &  \\
		Sahu et al. (2021) \cite{sahu2021simulation} & SE,~SN,~DA & & \ding{51} & & & \ding{51} & \ding{51} & & & & & \ding{51} &  \\
		Su et al. (2018) \cite{su2018comparison} & USV & & & & & & & & & & & \ding{51} &  \\
		Su et al. (2018) \cite{su2018real} & USV & & & & & & & & & & & \ding{51} &  \\
		Suzuki et al. (2019) \cite{suzuki2019depth} & SV & & & \ding{51} & & & & & & & & &  \\
		Wang et al. (2021) \cite{wang2021surgical} & SV & & \ding{51} & & & & & & & & & & \ding{51} \\
		Wang et al. (2022) \cite{wang2022rethinking} & SN,~DA & & & \ding{51} & \ding{51} & & & & & & & &  \\
		Yang et al. (2022) \cite{yang2022attention} & SV & \ding{51} & & \ding{51} & & & & & & \ding{51} & & &  \\
            Yang et al. (2022) \cite{yang2022tmf} & SV & \ding{51} & & \ding{51} & & & & & & \ding{51} & & &  \\
		Yang et al. (2022) \cite{yang2022weakly} & WE & & & \ding{51} & & & & & & & & &  \\
            Yang et al. (2023) \cite{yang2023maf} & SV & \ding{51} & & \ding{51} & & & & & & \ding{51} & & &  \\
            Yang et al. (2023) \cite{yang2023tma} & SV & \ding{51} & & \ding{51} & & & & \ding{51} & & & & &  \\
		Yeung (2021) \cite{yeung2021attention} & SV & \ding{51} & & & & & & & & \ding{51} & & &  \\
		Yu et al. (2020) \cite{yu2020holistically} & SV & & & \ding{51} & & & & & & & & &  \\
		Zhang et al. (2021) \cite{zhang2021surgical} & USV & & & \ding{51} & & & & & & & & \ding{51} &  \\
		\bottomrule
	\end{tabularx}
	\label{tab:seg:semantic_seg:single_frame_seg_binary}
\end{table*}

Supervised learning methods represent a group of approaches in which neural networks are trained using pixel-precise segmentation masks, which serve as the ground truth.
Most of the reviewed work regarding single-image segmentation of surgical instruments in endoscopic images and videos has been developed using this approach.
Here, most of the work focuses on improving the segmentation quality \linebreak by developing new architectures or adapting existing architectures by enhancing certain components~\cite{banik2021net, chou2021automatic, devi2022msdfn, dong2021semantic, galdran2021polyp, guo2022conditional, hasan2021detection, hasan2021segmentation, he2020multiscale, kamrul2019unetplus, keprate2021kvasir, laina2017concurrent, ni2019rasnet, qin2020towards, rajak2021segmentation, shen2023branch, suzuki2019depth, vishal2018robotic, wang2021pai, wang2021surgical, wang2023cgba, yang2022attention, yang2022tmf, yang2023maf, yang2023tma, yeung2021attention, yu2020holistically, zhou2021hierarchical}.

Other developments aim at improving segmentation quality by adding parallel processing of an auxiliary task. 
In addition to predicting the segmentation, this can include information about whether an instrument is present in a video frame or not, as well as the determination of geometric primitives~\cite{hasan2021detection}.
Other work estimates a saliency map in addition to segmentation and computes a scan path from it~\cite{islam2019learning}.
Improvements in segmentation quality can be achieved by predicting bounding box coordinates in parallel~\cite{islam2020ap} or by estimating a localization heatmap~\cite{laina2017concurrent}.
Recent research also shows that additional depth estimation in stereo images leads to better segmentation results~\cite{psychogyios2022msdesis, suzuki2019depth}.
Regarding instrument shape, current developments show that additional auxiliary contour supervision leads to more precise contour shape predictions for final segmentation~\cite{qin2020towards}.
Another supervised method relies on tool-pose-informed variable center morphological polar transform, by which segmentation in endoscopic images is improved~\cite{huang2022surgical}.

In contrast to the above works, current research also deals with the comparison regarding the strengths and weaknesses of existing supervised methods on specific validation datasets~\cite{han2021ceid, jha2021exploring, kalavakonda2019autonomous}.

However, following a supervised learning strategy and, thus, needing annotated data to develop novel methods faces several challenges, as labeling the data is a time-consuming and costly process that is typically performed manually.

\subsubsection*{Semi-Supervised Learning}

Semi-supervised learning is one approach to address the challenges posed by the need for annotated data.
This requires a sufficiently large amount of data, of which only a small fraction needs to be annotated, thus significantly reducing the annotation effort.
Developments in this area are based on teacher-student learning approaches, which are well-established methods for this purpose and have been used successfully in other domains of semi-supervised learning.
Recent work proposes an approach that learns on annotated synthetic and unlabeled real-world data, combining loss functions from both branches~\cite{sahu2021simulation}.
A different approach by Sanchez-Matilla et al. \cite{sanchez2021scalable} describes a method based on semi-supervised learning and weak supervision.
The authors propose an adaptive loss that allows training the network when only a small amount of annotated data is available.
A semi-supervised contrastive learning approach published by Lou et al. \cite{lou2023min} adapts the idea of solving a min-max similarity problem by classifying and projecting \pagebreak all-negative and positive-negative feature pairs, which are used to optimize the minimum-maximum similarity problem.

\subsubsection*{Weakly-Supervised Learning}

\begin{table*}[!t]
	\centering
	\caption{Publications that contributed in the area of non-binary semantic single-image segmentation of surgical instruments. For each publication, the type of methodology (SV = supervised, USV = unsupervised, WE = weakly-supervised, SE = semi-supervised, SN = synthetic data, DA = domain adaptation), whether or not attention mechanisms (Att.) were used in the methodology, and the datasets used are indicated. The works are grouped according to the segmentation types visualized in Figure~\ref{fig:seg_types_sem_seg_single_frame}, excluding the binary category.}
	\begin{tabularx}{\textwidth}{p{5.0cm}p{1.8cm}P{0.4cm}P{0.3cm}P{0.3cm}P{0.3cm}P{0.3cm}P{0.4cm}P{0.7cm}P{0.7cm}P{0.7cm}P{0.7cm}P{0.5cm}P{0.5cm}}
		\toprule
		\multirow{3}{*}{Publication}
		& \multirow{3}{*}{\begin{tabular}{l}\\Super- \\ vision\end{tabular}}
		& \multirow{3}{*}{Att.}
		& \multicolumn{10}{c}{Used Datasets} \\
		\cmidrule(llrr){4-14}
		&
		& 
		& \multicolumn{4}{c}{EndoVis}
		& \multirow{2}[3]{*}{\shortstack{Lap. \\ I2I}}
		& \multirow{2}[3]{*}{\shortstack{Sinus \\ Surg.}}
		& \multirow{2}[3]{*}{\shortstack{UCL \\ dVRK}}
		& \multirow{2}[3]{*}{\shortstack{Kvasir \\ Inst.}}
		& \multirow{2}[3]{*}{\shortstack{Robo- \\ Tool}}
		& \multirow{2}[3]{*}{\shortstack{Pri- \\ vate}}
		& \multirow{2}[3]{*}{\shortstack{Ot- \\ her}} \\
		\cmidrule(llrr){4-7}
		&
		& 
		& 15
		& 17
		& 18
		& 19
		& & & & & & & \\
		\midrule
		Liu et al. (2021) \cite{liu2021prototypical} & DA & & & \ding{51} & \ding{51} & & & & & & & &  \\
		Ni et al. (2020) \cite{ni2020attention} & SV & \ding{51} & & \ding{51} & & & & & & & & & \ding{51} \\
		Ni et al. (2020) \cite{ni2020pyramid} & SV & \ding{51} & & \ding{51} & & & & & & & & & \ding{51} \\
		Ni et al. (2021) \cite{ni2021barnet} & SV & \ding{51} & & \ding{51} & & & & & & & & & \ding{51} \\
		Ni et al. (2022) \cite{ni2022surginet} & SV & \ding{51} & & \ding{51} & & & & & & & & & \ding{51} \\
		Sanchez-Matilla et al. (2021) \cite{sanchez2021scalable} & SE & & & & \ding{51} & & & & & & & &  \\
            Shen et al. (2023) \cite{shen2023branch} & SV & \ding{51} & & \ding{51} & \ding{51} & & & & & & & \ding{51} &  \\
            Wang et al. (2023) \cite{wang2023cgba} & SV & \ding{51} & & & \ding{51} & & & & & & & & \ding{51} \\
		  Xue and Gu (2021) \cite{xue2021surgical} & SV & \ding{51} & & & & & & & & & & \ding{51} &  \\
            \cdashlinelr{1-14}
            Andersen et al. (2021) \cite{andersen2021real} & SV & & & & & & & & & & & \ding{51} &  \\
            Dong et al. (2021) \cite{dong2021semantic} & SV & & & \ding{51} & & & & & & \ding{51} &  \\
            Guo et al. (2022) \cite{guo2022conditional} & SV & & \ding{51} & \ding{51} & \ding{51} & & & & & &  \\
            He et al. (2020) \cite{he2020multiscale} & SV & \ding{51} & \ding{51} & \ding{51} & \ding{51} & & & & & & & &  \\
            Nema and Vachhani (2023) \cite{nema2023unpaired} & SV, USV & & \ding{51} & & & & & & & & & & \\
            Ozawa et al. (2021) \cite{ozawa2021synthetic} & SN, DA & & & & & & & & & & & \ding{51} &  \\
            Sun et al. (2021) \cite{sun2021lightweight} & SV & & & \ding{51} & & & & & & & & &  \\
            \cdashlinelr{1-14}
            Islam et al. (2019) \cite{islam2019learning} & SV & \ding{51} & & \ding{51} & & & & & & & & &  \\
            Islam et al. (2020) \cite{islam2020ap} & SV & \ding{51} & & \ding{51} & & & & & & & & &  \\
            \cdashlinelr{1-14}
            Colleoni et al. (2022) \cite{colleoni2022ssis} & SN, DA & \ding{51} & & \ding{51} & & & & & \ding{51} & & & & \ding{51} \\
            Laina et al. (2017) \cite{laina2017concurrent} & SV & & \ding{51} & & & & & & & & & &  \ding{51} \\
            Mahmood et al. (2022) \cite{mahmood2022dsrd} & SV & & & \ding{51} & & & & & & \ding{51} & & & \\
            Pakhomov et al. (2019) \cite{pakhomov2019deep} & SV & & & \ding{51} & & & & & & & & &  \\
            Pakhomov and Navab (2020) \cite{pakhomov2020searching} & SV & & & \ding{51} & & & & & & & & &  \\
            \cdashlinelr{1-14}
            Islam et al. (2019) \cite{islam2019real} & SV & & & \ding{51} & & & & & & & & &  \\
            Kamrul H. and Linte (2019) \cite{kamrul2019unetplus} & SV & & & \ding{51} & & & & & & & & &  \\
            Liu et al. (2022) \cite{liu2021graph} & DA & & & \ding{51} & \ding{51} & & & & & & & &  \\
            Shvets et al. (2018) \cite{shvets2018automatic} & SV & & & \ding{51} & & & & & & & & &  \\
            Vishal and Kumar (2018) \cite{vishal2018robotic} & SV & \ding{51} & & \ding{51} & & & & & & & & &  \\
            Wang et al. (2021) \cite{wang2021pai} & SV & \ding{51} & & \ding{51} & & & & & & & & & \ding{51} \\
            Zhou et al. (2021) \cite{zhou2021hierarchical} & SV & \ding{51} & & \ding{51} & & & & & & & & &  \\
		\bottomrule
	\end{tabularx}
	\label{tab:seg:semantic_seg:single_frame_other_types}
\end{table*} 

To overcome the lack of annotated data, the field of self-supervised learning, often referred to as weakly supervised learning, is an active area of research. These approaches either require no annotation at all or only very simple and rapidly generated manual annotations.
Two types of weakly supervised learning can be distinguished.

One method is to automatically generate labels based on image data and then use this generated target data to train a neural network. 
An approach to creating the training labels is to use the estimate of a robot's kinematic model in terms of the shape of the surgical tool to learn a projection of this data onto the input image using a cost function based on the GrabCut algorithm.
This target data is then employed to train a segmentation network~\cite{da2019self}. 
In the work of Pakhomov et al. \cite{pakhomov2020towards}, the authors also propose a process where imprecise kinematic data is used to generate training labels, and mappings between surgical images and automatically generated masks are performed using an unpaired GAN-based I2I translation approach. 
The training of a segmentation network is then performed using the weak target data generated in this way.
Another method is to obtain labels using both electromagnetic tracking and laparoscopic image processing and then use these labels to train a lightweight network~\cite{lee2019weakly}.
Another work uses existing models trained on publicly available datasets to semi-automatically generate the training dataset and processes the results using a watershed-based segmentation method.
A two-step training process is then used where the input images are first processed using the publicly available segmentation model, and the results are refined using a GrabCut-based algorithm~\cite{lee2019segmentation}.
A similar approach by Huang et al. \cite{huang2022simultaneous} generates ground truth segmentation masks through a model trained on a publicly available dataset and uses these masks to train their network for joint segmentation prediction and disparity estimation.

A second way to apply weakly supervised learning is to work only with manually generated coarse annotations that can be acquired quickly and cost-effectively.
By providing bounding boxes as weak annotations, training datasets for segmentation tasks can be generated by applying the DeepMAC~\cite{birodkar2021surprising} method.
Subsequently, segmentation models are trained using these target data and domain-adapted synthetically generated data~\cite{leifman2022pixel}.
An efficient method for creating a training dataset is described by Papp et al. \cite{papp2022surgical}, where only one video frame per second is manually labeled, and all intervening video frames are semi-automatically annotated by applying optical flow (OF)-based tracking.
Four established segmentation networks were trained with the annotations generated in this way, to assess the surgical skills of the operators.
The creation of weak labels can also be aided by the use of a virtual reality (VR) simulator, where the ground truth is determined from this data with minimal manual effort, rendering the instruments with a solid white color and the background with uniform black color at three different levels of realism.
The data generated in this way can then be used to train segmentation networks \cite{heredia2020effects}.
An alternative approach to semi-automatically generate the training dataset is to create coarse, scribble-like weak annotations of the tools and transform them into training data using a graph-based method.
Subsequently, a segmentation model is trained using these noisy annotations in a two-step process divided into a warm-up phase and a stabilization phase, characterized by the use of highly noisy weak annotations in the warm-up phase and less noisy labels in the stabilization phase~\cite{yang2022weakly}.

\subsubsection*{Different Learning Approaches}

This section deals with approaches that do not fit strictly into the previously mentioned learning variants.
In their work, Qin et al. \cite{qin2019surgical} describe a segmentation network that is trained in a two-step procedure.
In the first step, the feature extractor is trained unsupervised on a large set of unlabeled data.
In the second step, they train the segmentation part of the network in a supervised fashion on a small set of annotated images.
The CNN results are fused with the kinematic pose of the instruments to obtain the final predictions.
The application of a GAN-based unpaired I2I approach to surgical instrument segmentation is presented by Zhang et al. \cite{zhang2021surgical}.
The core concepts are an embedded constraint, which specifies that each pixel belongs to either an instrument or background, and the use of textured instrument images as annotations for the network generator rather than segmentation masks.
The approach of Nema and Vachhani \cite{nema2023unpaired} also relies on a CycleGAN architecture that learns a mapping between raw surgical images and segmentation maps for instruments in an unsupervised manner, with the objective of using sparsely available training data to achieve optimal results for a large amount of unseen test data.
In contrast to previously presented methods, the approach described by Su et al. \cite{su2018real} is based on classical image processing methods and does not include machine learning elements.
The authors present a multi-stage approach that is essentially based on the fusion of shape matching using a discrete Fourier transform and a color mask determined by log-likelihood.
The same authors describe a similar method, which relies on the same elements and deals with the segmentation of surgical tools and their 3D reconstruction~\cite{su2018comparison}.

\subsubsection{Domain Adaptation}
\label{semantic_segmentation:single_frame_seg:domain_adaptation}

Another frequent requirement is the adaptation to changes in the domain, e.g., concerning new instrument sets or different types of operations.
In these cases, retraining for the adapted circumstances would be needed.
One approach that meets this challenge is called domain adaptation.
Often, such approaches employ a two-step process in which synthetic data is generated first and then used in the training process, which is also referred to as simulation-supervised learning~\cite{colleoni2022ssis}.

A commonly used way to do this is to use unpaired I2I translation to translate data from a synthetic to a given real-world domain, i.e., to generate artificial images that look as realistic as possible on the desired target domain.
For this purpose, architectures based on generative adversarial networks (GANs), specifically CycleGAN architectures~\cite{zhu2017unpaired}, are frequently used.
The resulting artificial images are then employed to train a neural network \cite{colleoni2021robotic, colleoni2022ssis, leifman2022pixel, ozawa2021synthetic}.
However, domain adaptation based on the generation of synthetic data can also be performed in other ways. 
One possibility is to use a VR simulator to generate endoscopic data with different levels of realism, train a segmentation network with this data, and evaluate the influence of the degree of realism of the simulated data on the predictive quality of the network~\cite{heredia2020effects}.
A further possibility for the generation of artificial training data is the blending of images. 
In a promising approach, Wang et al. \cite{wang2022rethinking} select an image from the EndoVis-2018 dataset that shows only background and does not contain any tools, and multiply it by applying a variety of augmentations.
Then, images containing instruments are selected for the foreground, these instruments are extracted, and the background is made transparent.
The instruments are then blended onto the background images, and the resulting dataset is used to train a segmentation network whose performance is also evaluated on a different real-world dataset.

In addition, there is work in the field of domain adaptation that does not use a two-step process not uses I2I-based techniques.
Instead, a model based on consistency is trained simultaneously for real-world and simulated data in an end-to-end manner.
In this approach, the simulated data are processed according to a supervised technique, whereas the images of the real-world domain are processed according to a consistency learning strategy, combining the losses of both branches~\cite{sahu2020endo}.
An improvement of this method is described by Sahu et al. \cite{sahu2021simulation} by improving the generation of the pseudo-labels for data belonging to the real-world domain.

Another type of domain adaptation deals with the transfer from an annotated dataset to an unknown one for which no labels are available and which represents a divergent surgical scene.
Here, both datasets show real-world interventions, and unlike the methods described above, this formulation of the problem does not involve artificially generated data been  in advance.
Regardless of the domain shift considered, unpaired I2I translation based on CycleGAN architectures is also used for this purpose. 
Kalia et al. \cite{kalia2021co} present an approach in which the generator performs the transfer from the labeled to the unlabeled dataset, learns a segmentation model based on these results, and influences the generator in the process.
Other approaches use graph-based neural networks to best model the relationships between the two domains.
This involves extracting relevant features of the input images of both domains through a CNN and processing them through different regions with associated responsibilities in the graph-based network~\cite{liu2021prototypical, liu2021graph}.

A further type of domain adaptation is described in the work of Psychogyios et al. \cite{psychogyios2022msdesis}. 
The authors present an architecture that jointly estimates the disparity and segmentation of surgical tools.
The segmentation task is performed in a supervised manner using annotations available only for the left image of the stereo camera, and the disparity estimation is modeled as an auxiliary task, improving the performance in terms of disparity estimation only by training the segmentation task.

In addition to work aimed at improving quality in various aspects through domain adaptation, some publications focus on the creation and provision of synthetic datasets.
One way is by moving and recording instruments against a uniform background, automatically segmenting the tools, and blending them into images showing only the background~\cite{garcia2021image}.

\subsubsection{Inference Speed}
\label{semantic_segmentation:single_frame_seg:inference_speed}

\begin{table*}[tbh]
	\centering
	\caption{Semantic segmentation methods for single images that have specified processing speeds at inference time. Indicated are the publication, the speed in FPS, the image resolution, the GPU used as well as the number of graphics cards, and the CPU used and the number of CPUs. The publications are sorted in descending order according to the reported inference speed.}
	\begin{tabularx}{\textwidth}{lrclclc}
	\toprule
        \multirow{2}{*}{Publication} & \multicolumn{1}{c}{Inf. Speed} & \multirow{2}{*}{Resolution} & \multirow{2}{*}{GPU} & \multirow{2}{*}{$\#$\,GPU} & \multirow{2}{*}{CPU} & \multirow{2}{*}{$\#$\,CPU} \\    
        & \multicolumn{1}{c}{(FPS)} & & & & & \\
        \midrule
        Islam et al. (2019) \cite{islam2019real} & 174 & $1024 \times 1280$ & GTX 1080 Ti & 1 & - & - \\
        Huang et al. (2022) \cite{huang2022simultaneous} & 172 & $192 \times 384$ & RTX 2080 Ti & 1 & - & - \\
        \multirow{2}{*}{Guo et al. (2022) \cite{guo2022conditional}} & 171, 175 & $1024 \times 1280$ & \multirow{2}{*}{GTX 1070} & \multirow{2}{*}{1} & \multirow{2}{*}{2.80 GHz} & \multirow{2}{*}{1} \\
        & 167 & $480 \times 640$ & & & &\\
        Islam et al. (2019) \cite{islam2019learning} & 127 & - & GTX 1080 Ti & 3 & - & - \\
        Pakhomov and Navab (2020) \cite{pakhomov2020searching} & 125 & $1024 \times 1280$ & Tesla P100 & 1 & - & - \\
        Jha et al. (2021) \cite{jha2021exploring} & 102 & $540 \times 960$ & DGX-2 & 1 & - & - \\
        Shen et al. (2023) \cite{shen2023branch} & 102 & $512 \times 640$ & RTX A6000 & 1 & Xeon Gold 6226R & 1 \\
        \multirow{2}{*}{Andersen et al. (2021) \cite{andersen2021real}} & 90 & $448 \times 448$ & \multirow{2}{*}{RTX 2080 Ti} & \multirow{2}{*}{1} & \multirow{2}{*}{-} & \multirow{2}{*}{-} \\
        & 36 & $672 \times 1120$ & & & & \\
        Garcia-P.-H. et al. (2017) \cite{garcia2017toolnet} & 43 / 29 / 8 & $576 \times 720$ & GTX Titan X & 1 & Xeon E5-1650 3.50 GHz & 1 \\
        Lou et al. (2023) \cite{lou2023min} & 40 & $288 \times 512$ & RTX A5000 & 1 & - & - \\
        Ni et al. (2020) \cite{ni2020attention} & 39 & $544 \times 960$ & Titan X & 1 & - & - \\
        Sun et al. (2021) \cite{sun2021lightweight} & 37 & $1024 \times 1280$ & P6000 & 1 & - & - \\
        Xue and Gu (2021) \cite{xue2021surgical} & 25 & $224 \times 224$ & - & - & - & - \\
        Psychogyios et al. (2022) \cite{psychogyios2022msdesis} & 22 & $1024 \times 1280$ & DGX V100 & 1 & - & - \\
        Sanchez-Matilla et al. (2021) \cite{sanchez2021scalable} & 22 & $512 \times 512$ & RTX 6000 & 1 & - & - \\
        \multirow{2}{*}{Mahmood et al. (2022) \cite{mahmood2022dsrd}} & \multirow{2}{*}{21} & $512 \times 512$ & \multirow{2}{*}{GTX 1070} & \multirow{2}{*}{1} & \multirow{2}{*}{Core-i7-7700 3.60 GHz} & \multirow{2}{*}{1} \\
        & & $512 \times 640$ & & & & \\
        Islam et al. (2020) \cite{islam2020ap} & 18 & $1024 \times 1280$ & RTX 2080 Ti & 1 & - & - \\
        Shvets et al. (2018) \cite{shvets2018automatic} & 11 / 6 / 5 & $1024 \times 1280$ & GTX 1080 Ti & 1 & - & - \\
        Su et al. (2018) \cite{su2018real} & 6 & $480 \times 640$ & No GPU & 0 & - & - \\
        \bottomrule
	\end{tabularx}
	\label{tab:sem_seg:single_frame:inference_speed}
\end{table*}

Besides improving the quality of segmentation, recent work focuses on fast processing and preferably real-time processing capability of the proposed applications, which represents a prerequisite to be applicable in real-world scenarios.
In this context, the speed at inference time represents an important criterion.
In the literature, this speed is often specified in frames per second (FPS), i.e., the number of video frames a method processes within one second.
In our review, we obtained a total of 19 contributions presented in Table~\ref{tab:sem_seg:single_frame:inference_speed} that determined and reported the inference speed of their methods concerning the semantic segmentation of individual frames. 
To allow for a fair comparison, these works are listed together with their respective processing speed at inference time in FPS, image resolution, used GPU and the number of graphics cards used, and CPU and the amount of CPUs used.

\subsubsection{Attention Mechanisms}
\label{semantic_segmentation:single_frame_seg:attention_mechanisms}

In addition to the established network architectures and components used for segmentation tasks, current research explores the potential of attention techniques.
These address the limitations associated with traditional CNN-based approaches.
The core idea is to use attention-based mechanisms to extract more useful features and to better suppress irrelevant ones resulting in improved segmentation quality.

Current work in the field of single-frame segmentation of surgical instruments shows promising results that achieved by using attention-oriented methods, in particular by modifying and extending parts of encoder-decoder-based architectures.
Other work uses attention-based mechanisms for a dual-model filtering strategy to reduce the number of false-positive predicted pixels~\cite{chou2021automatic}.
In unpaired I2I translation with CycleGAN-based architectures, generated synthetic images can be improved by integrating an attention module trained in parallel with the generator, supporting the model to focus on domain-specific features~\cite{colleoni2022ssis}.
Approaches that address the problem of overparameterization of neural networks demonstrate the benefit of dynamic attention-based pruning strategies that identify and eliminate non-meaningful features after each encoder block~\cite{islam2020ap}.
Other recent research examines the impact of differently designed attention-based modules deployed in the skip connections, i.e., the links between the encoder and decoder blocks~\cite{vishal2018robotic, wang2021pai, yeung2021attention}.
In addition to this work the use of attention-based techniques with a particular focus on the decoder part of architectures is being investigated, for example, by adding such an attention module after each decoder block~\cite{ni2019rasnet}.
Related recent research describes how the use of attention-based techniques in a second, parallel decoder branch with a divergent task has a positive effect on segmentation quality.
Of all the models compared by Jha et al. \cite{jha2021exploring}, the model with the best results uses an autoencoder combined with attention modules as an additional branch in the decoder to refine the features of the encoder part, and Islam et al. \cite{islam2019learning} include attention-based modules into a parallel decoder branch to predict a saliency map.
Also focusing on the decoder, Shen et al. \cite{shen2023branch} integrate a block attention fusion module into the decoder for efficient feature localization and global context understanding.
Between the lightweight encoder and the decoder, a branch balance aggregation module is placed, which is responsible for the fusion of feature maps of different levels as well as for the suppression of noise in the low-level feature maps.
Yang et al. \cite{yang2022attention} use attention-based techniques both in the bottleneck layer in the encoder part of the network to better capture global concepts of the input, as well as in the form of a dual-attention module that combines the high-level features and the low-level features of the decoder and encoder, respectively. Therein, valuable features are recognized and irrelevant ones are suppressed, thus improving the overall segmentation quality.
To overcome the drawbacks of convolutional neural networks concerning the capture of features of smaller objects as well as weaknesses in the processing of local semantic features, Yang et al. \cite{yang2023maf} introduce a multi-scale attention fusion network, which relies on an attention fusion module that integrates multi-scale information by cross-scale feature fusion.
Capturing multi-scale information and using attention-based fusion modules is also the central objective of the method of Yang et al. \cite{yang2022tmf, yang2023tma}, who aim to suppress irrelevant information in the low-level features by using attention feature fusion modules and additive attention and concatenation modules in their proposed network architecture.
By using guidance connection attention modules, Wang et al. \cite{wang2023cgba} also attempt to eliminate irrelevant low-level features.
In addition, the inclusion of local information and local-global dependencies through bidirectional attention modules to identify accurate instrument features is also a central aspect of their method.

\subsection{Incorporating Temporal Information}
\label{semantic_segmentation:incorporating_temporal_information}

This section presents recent research on semantic surgical instrument segmentation incorporating temporal information.
In the following, these works are analyzed according to the type of segmentation, the learning strategies, the methodological approaches used to segment and track the tools over time, the inference speed, and the incorporation of attention-based techniques to improve the segmentation results.

\subsubsection{Segmentation Type}
\label{semantic_segmentation:temporal_info:segmentation_type}

\begin{figure}[!b]
\includegraphics[width=0.49\textwidth]{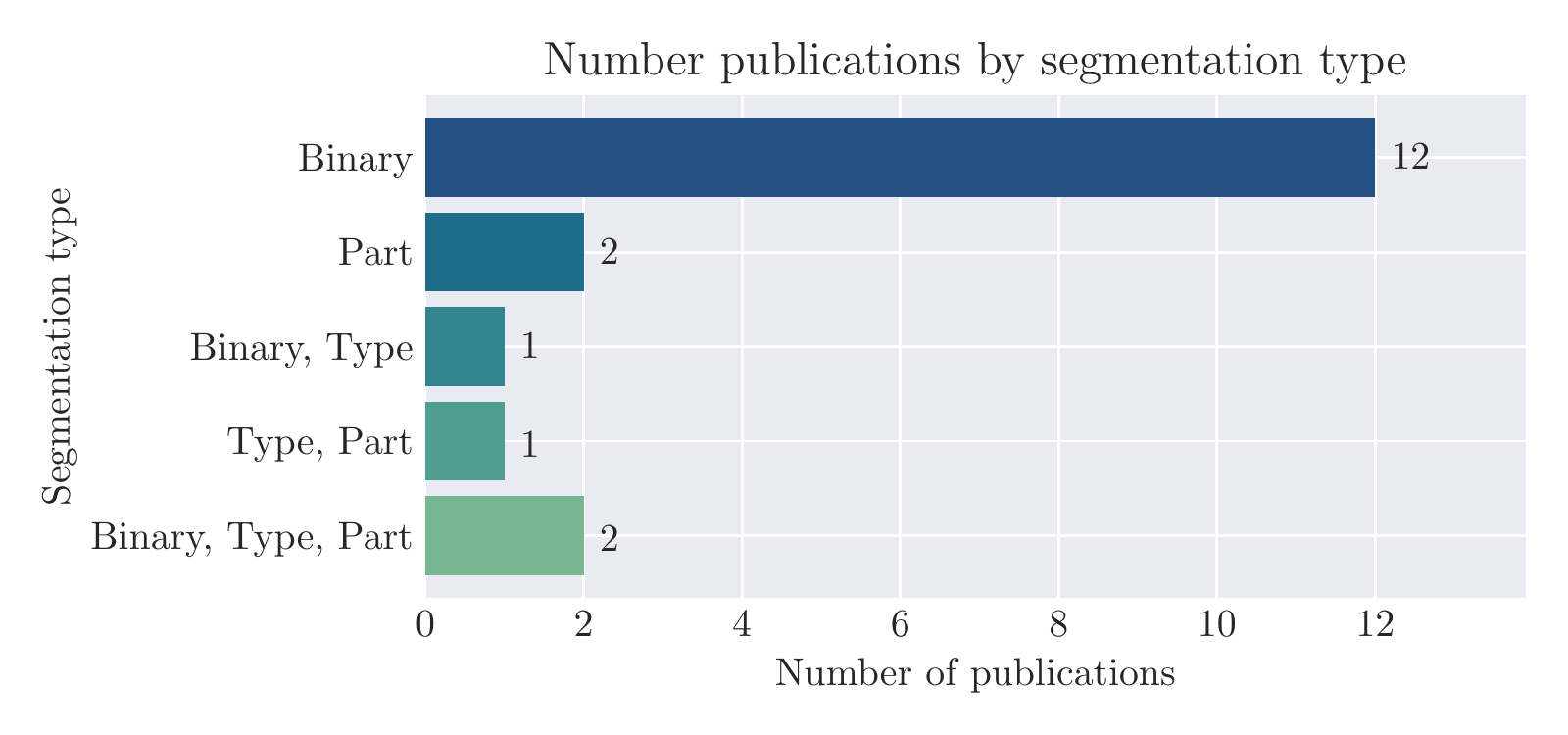}
\caption{Number of relevant publications per segmentation type for semantic image segmentation incorporating temporal information.}
\label{fig:seg_types_sem_seg_temporal_info}
\end{figure}

The division of segmentation types for the papers considered in this chapter is visualized in Figure~\ref{fig:seg_types_sem_seg_temporal_info}, and a listing showing all publications grouped by these types is presented in Table~\ref{tab:seg:semantic_seg:temporal_info}, whose structure is identical to the one described in Chapter~\ref{semantic_segmentation:single_frame_seg}.

\subsubsection{Learning Strategies}
\label{semantic_segmentation:temporal_info:learning_strategy}

\begin{table*}[tbph]
	\centering
	\caption{Publications that have contributed in the area of semantic segmentation of surgical instruments by incorporating temporal information. For each publication, the type of methodology (SV = supervised, USV = unsupervised, WE = weakly-supervised, SE = semi-supervised, SN = synthetic data, DA = domain adaptation), whether or not attention mechanisms (Att.) were used in the methodology, and the datasets used are indicated. The works are grouped according to the used segmentation types visualized in Figure~\ref{fig:seg_types_sem_seg_temporal_info}.}
	\begin{tabularx}{\textwidth}{p{5.0cm}p{1.8cm}P{0.4cm}P{0.3cm}P{0.3cm}P{0.3cm}P{0.3cm}P{0.4cm}P{0.7cm}P{0.7cm}P{0.7cm}P{0.7cm}P{0.5cm}P{0.5cm}}
		\toprule
		\multirow{3}{*}{Publication}
		& \multirow{3}{*}{\begin{tabular}{l}\\Super- \\ vision\end{tabular}}
		& \multirow{3}{*}{Att.}
		& \multicolumn{10}{c}{Used Datasets} \\
		\cmidrule(llrr){4-14}
		&
		& 
		& \multicolumn{4}{c}{EndoVis}
		& \multirow{2}[3]{*}{\shortstack{Lap. \\ I2I}}
		& \multirow{2}[3]{*}{\shortstack{Sinus \\ Surg.}}
		& \multirow{2}[3]{*}{\shortstack{UCL \\ dVRK}}
		& \multirow{2}[3]{*}{\shortstack{Kvasir \\ Inst.}}
		& \multirow{2}[3]{*}{\shortstack{Robo- \\ Tool}}
		& \multirow{2}[3]{*}{\shortstack{Pri- \\ vate}}
		& \multirow{2}[3]{*}{\shortstack{Ot- \\ her}} \\
		\cmidrule(llrr){4-7}
		&
		& 
		& 15
		& 17
		& 18
		& 19
		& & & & & & & \\ 
		\midrule
		Agustinos and Voros (2015) \cite{agustinos20152d} & USV & & & & & & & & & & & \ding{51} &  \\
		Amini Khoiy et al. (2016) \cite{amini2016automatic} & USV & & & & & & & & & & & \ding{51} &  \\
		Attia et al. (2017) \cite{attia2017surgical} & SV & & \ding{51} & & & & & & & & & &  \\
		Du et al. (2016) \cite{du2016combined} & SV, USV & & \ding{51} & & & & & & & & & & \ding{51} \\
		  Garcia-P.-H. et al. (2016) \cite{garcia2016real} & SV, USV & & \ding{51} & & & & & & & & & \ding{51} & \ding{51} \\
            Lin et al. (2019) \cite{lin2019automatic} & SV & & & & & & & & & & & \ding{51} &  \\
            Lin et al. (2021) \cite{lin2021multi} & SV & \ding{51} & & & & \ding{51} & & \ding{51} & & & & &  \\
            Liu et al. (2020) \cite{liu2020unsupervised} & WE & & & \ding{51} & & & & & & & & &  \\
            Sestini et al. (2023) \cite{sestini2023fun} & USV &  & & \ding{51} & & & & & & & \ding{51} & & \ding{51} \\
            Yang et al. (2022) \cite{yang2022drr} & SV & \ding{51} & & & & & & \ding{51} & & \ding{51} & & &  \\
            Zhao et al. (2021) \cite{zhao2021anchor} & DA & \ding{51} & & \ding{51} & \ding{51} & & & & & & & \ding{51} & \ding{51} \\
		  Zhao et al. (2021) \cite{zhao2021one} & DA & & & \ding{51} & \ding{51} & & & & & & & \ding{51} &  \\
            \cdashlinelr{1-14}
            Li et al. (2021) \cite{li2021preserving} & SE, USV & & & \ding{51} & & & & & & & & &  \\
            Zhang and Gao (2020) \cite{zhang2020object} & SV & & & & & & & & & & & \ding{51} &  \\
            \cdashlinelr{1-14}
		Islam et al. (2021) \cite{islam2021st} & SV & \ding{51} & & \ding{51} & & & & & & & & &  \\
            \cdashlinelr{1-14}
            Wang et al. (2021) \cite{wang2021efficient} & SV & \ding{51} & & \ding{51} & \ding{51} & & & & & & & &  \\
            \cdashlinelr{1-14}
		Jin et al. (2019) \cite{jin2019incorporating} & SE, USV & \ding{51} & & \ding{51} & & & & & & & & &  \\
		  Zhao et al. (2020) \cite{zhao2020learning} & SE, USV & & & \ding{51} & & & & & & & & &  \\
		\bottomrule
	\end{tabularx}
	\label{tab:seg:semantic_seg:temporal_info}
\end{table*}

\begin{figure}[!b]
\centering
\includegraphics[width=0.40\textwidth]{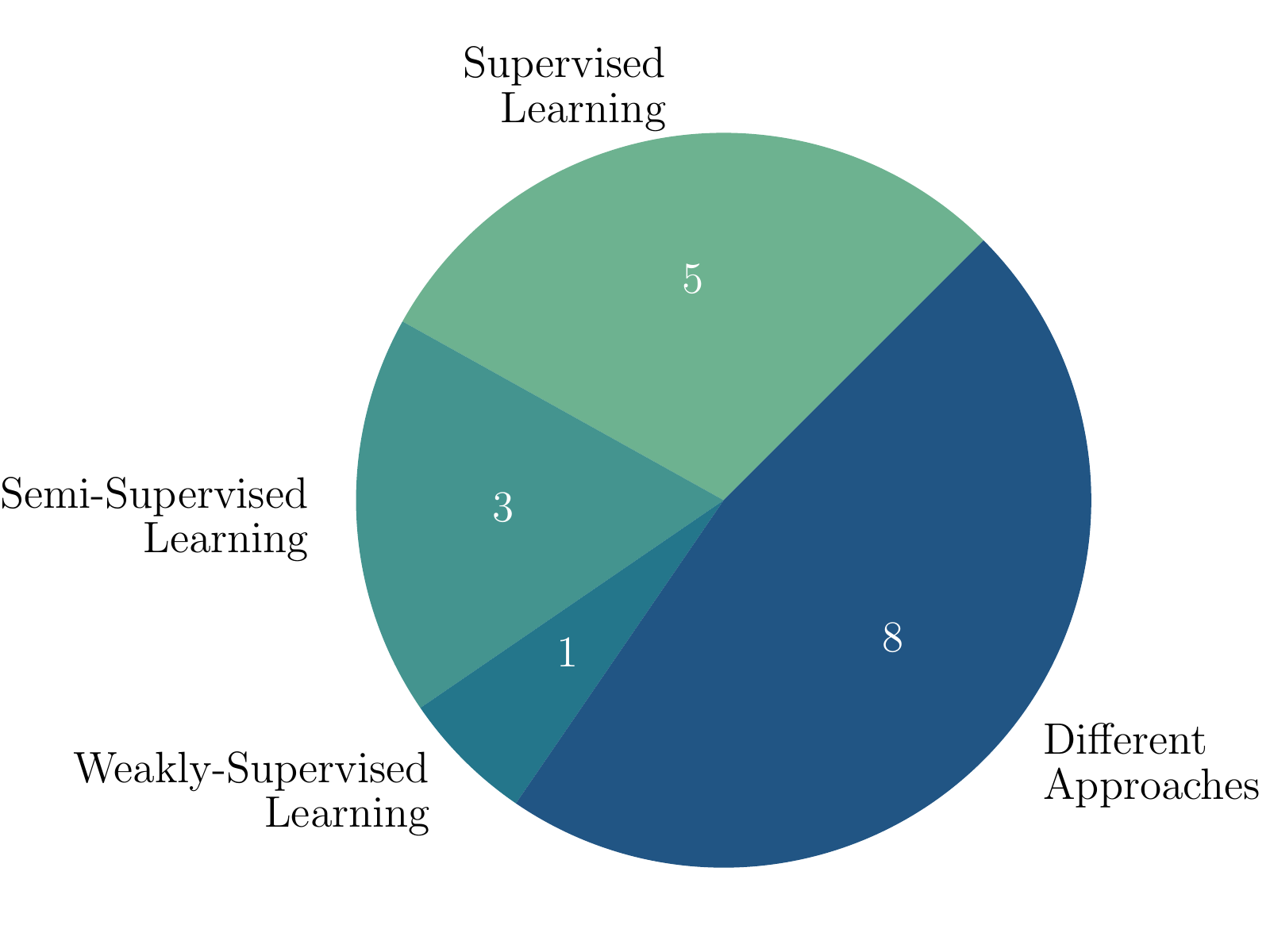}
\caption{Amount of relevant papers for semantic segmentation incorporating temporal information divided according to the respective learning strategies used.}
\label{fig:learning_types_sem_seg_temporal_info}
\end{figure}

Regarding the learning strategies, the included literature can be divided into various groups presented together with their appearance in Figure~\ref{fig:learning_types_sem_seg_temporal_info}.
The majority of the reviewed methods are based on supervised approaches~\cite{attia2017surgical, garcia2016real, islam2021st, lin2019automatic, wang2021efficient}.
To overcome the lack of annotated data, some works present semi-supervised methods using sparsely labeled video sequences \cite{jin2019incorporating, li2021preserving, zhao2020learning}.
Jin et al. \cite{jin2019incorporating} present a module through which the network's prediction for an unlabeled frame is applied to the previous annotated frame, and a loss is computed using its ground truth.
In the contribution of Zhao et al. \cite{zhao2020learning}, motion flows are learned in two parallel branches to propagating the annotations available for a few locations to the unlabeled video frames.
Then, the manually created annotations, the propagated annotations, and the labels for additional interpolated frames are used to train a segmentation network.
Based on this approach, the work of Li et al. \cite{li2021preserving} presents a method for improving the interpolation of the images and the associated labels between the annotated frames.
Various approaches use elements of unsupervised learning in addition to the above learning strategies to track the instruments over time, which are described in a later topic within this section \cite{garcia2016real, jin2019incorporating, li2021preserving, sestini2023fun, zhao2020learning}.

A different approach is taken by Liu et al. \cite{liu2020unsupervised}, which operates in a weakly supervised manner. Instead of annotated data, automatically generated pseudo-labels based on the input images and manually specified cues are used.
These pseudo-labels then serve as ground truth and are used to compute an anchor loss.
As with single-image segmentation, there exists research in the area of domain adaptation based on the incorporation of temporal information.
Zhao et al. \cite{zhao2021anchor, zhao2021one} present related approaches in which initial models are learned through meta-learning and then applied to new domains through an online adaptation scheme, requiring only an annotation for the first frame of a video sequence at inference time.
An unsupervised three-stage approach that consists solely of implicit motion information and instrument-oriented shape-priors is provided by Sestini et al. \cite{sestini2023fun}.
Working entirely without manual annotations or prior knowledge is the approach of Agustinos and Voros \cite{agustinos20152d}, where the 3D pose of the instruments is determined in a multi-step procedure using classical image processing techniques, essentially based on the use of the Frangi filter.
The method from Amini Khoiy et al. \cite{amini2016automatic} also works only with color features, first converting the input image to the more informative hue, saturation, value (HSV) color space, then performing V+S histogram thresholding followed by the application of a filter on the H channel.
Almost unsupervised is the method proposed by Du et al. \cite{du2016combined}, where only the 3D pose of the surgical tool in the first video frame needs to be known, and, based on this the position of the instrument in all subsequent frames is estimated.

\subsubsection{Tracking Approaches}
\label{semantic_segmentation:temporal_info:domain_adaptation}

\begin{table*}[tb]
	\centering
	\caption{Semantic segmentation methods incorporating temporal information that have specified processing speeds at inference time. Indicated are the publication, the speed in FPS, the image resolution, the GPU used as well as the number of graphics cards, and the CPU used and the number of CPUs. The publications are sorted in descending order according to the reported inference speed.}
	\begin{tabularx}{\textwidth}{lrclclc}
	\toprule
        \multirow{2}{*}{Publication} & \multicolumn{1}{c}{Inf. Speed} & \multirow{2}{*}{Resolution} & \multirow{2}{*}{GPU} & \multirow{2}{*}{$\#$\,GPU} & \multirow{2}{*}{CPU} & \multirow{2}{*}{$\#$\,CPU} \\    
        & \multicolumn{1}{c}{(FPS)} & & & & & \\
        \midrule
        \multirow{2}{*}{Lin et al. (2021) \cite{lin2021multi}} & 182 & $240 \times 240$ & \multirow{2}{*}{Titan Xp} & \multirow{2}{*}{1} & \multirow{2}{*}{Core-i7-7700K 4.20 GHz} & \multirow{2}{*}{1} \\
        & 117 & $360 \times 640$ & & & & \\
        Islam et al. (2021) \cite{islam2021st} & 42 & $1024 \times 1280$ & RTX 2080 Ti & 1 & - & - \\
        Wang et al. (2021) \cite{wang2021efficient} & 38 & $512 \times 640$ & Titan RTX & 4 & - & - \\
        Agustinos and Voros (2015) \cite{agustinos20152d} & 30 & $556 \times 720$ & No GPU & 0 & Xeon 2.67 GHz & 1 \\
        Amini Khoiy et al. (2016) \cite{amini2016automatic} & 30 & $480 \times 640$ & No GPU & 0 & Core Duo 2.5 GHz & 1 \\
        \multirow{4}{*}{Garcia-P.-H. et al. (2016) \cite{garcia2016real}} & \multirow{4}{*}{30} & $480 \times 640$, & \multirow{4}{*}{GTX Titan X} & \multirow{4}{*}{1} & & \multirow{4}{*}{1} \\
        & & $576 \times 720$, & & & Xeon E5-1650 v3, & \\
        & & $[460 \times 612,$ & & & 3.50 GHz & \\
        & & $1080 \times 1920]$ & & & & \\
        Zhang and Gao (2020) \cite{zhang2020object} & 15 & $480 \times 640$ & GTX 1080 Ti & 1 & Core i-7-7700 3.60 GHz & 1 \\
        Zhao et al. (2021) \cite{zhao2021one} & 4 & $1024 \times 1280$ & Titan Xp & 1 & - & - \\
        Zhao et al. (2021) \cite{zhao2021anchor} & 3 & $1024 \times 1280$ & Titan Xp & 1 & - & - \\
        \bottomrule
	\end{tabularx}
	\label{tab:sem_seg:temporal_info:inference_speed}
\end{table*}

This section summarizes the key concepts in the current literature for tracking surgical tools over time.
Often, this information is processed within the network architecture, for example, by using recurrent layers.
Attia et al. \cite{attia2017surgical} present an approach in which the low-level features resulting from the network's encoder are processed in four alternating convolutional and recurrent layers and then converted into the final output by the decoder.
Also based on the idea of recurrent layers is the method of Yang et al. \cite{yang2022drr}, which uses blocks of recurrent convolutional layers in both the encoder and decoder of its architecture.
Adapting the idea of processing recurrent sequences, Lin et al. \cite{lin2021multi} present a module located between the encoder and decoder of a segmentation network and can be used to temporally and spatially aggregate features of video frames.

In addition to recurrent components, long short-term memory (LSTM) blocks are often incorporated into the network structure.
Current work exploits temporal information by processing features at the bottleneck of an encoder-decoder architecture in LSTM blocks \cite{li2021preserving, zhao2020learning}.
Other articles integrate such layers into the decoder of an auxiliary task to improve overall segmentation results~\cite{islam2021st}.
Related publications describe an approach in which the features of successive frames are aggregated by convolutional LSTM layers~\cite{wang2021efficient}.

In addition to recurrent layers and LSTM mechanisms, determining the motion flow between two successive frames by computing the OF field is a commonly used approach.
To this end, Lin et al. \cite{lin2019automatic} present a method that computes the OF as an intermediate step in a multi-stage procedure to determine the potential position of the surgical instrument in the successive video frame.
The approach of Sestini et al. \cite{sestini2023fun} also relies on a multistep procedure in which unsupervised optical-flow segmentation is the first step.
For the generation of realistic binary segmentation maps, a generative-adversarial approach is used to learn the mapping between the domain of optical-flow images and the domain of shape-priors.
Garcia-Peraza-Herrera et al. \cite{garcia2016real} present a method that computes an affine transformation between feature points in the last segmented image and feature points in the current image and then applies this affine transformation to the predicted segmentation masks of a CNN.
Also based on OF computation is the approach of Jin et al. \cite{jin2019incorporating}, which uses the network's prediction for the previous image as a reliable prediction for the position and shape of an instrument in the subsequent frame.
In their works, Li et al. and Zhao et al. \cite{li2021preserving, zhao2020learning} use a network that determines the OF based on a pre-trained network.
The results are then processed in two parallel branches to jointly apply the network's predictions and associated annotations to subsequent video images.

Online adaptation describes a research approach that involves dynamically adapting existing models trained using an easily accessible source domain and then applying them to different target domains at test time, with only an annotated segmentation mask available for the first frame of the target domain.
A contribution by Zhao et al. \cite{zhao2021one} describes a meta-learning-based dynamic online adaptive learning approach using a two-step procedure. 
In step one, they train a segmentation network using a meta-learning approach that provides fast adaptable parameters for step two.
Then, this network is dynamically adapted to a new video using the first annotated frame in several iterations, employing the output of the previous frame and the updated parameters to generate pseudo-masks.
An improvement of this method is presented by Zhao et al. \cite{zhao2021anchor} by applying an anchor-oriented online meta-adaptation approach dealing with one-shot instrument segmentation in robotic surgical videos.
For this purpose, the meta-training of the segmentation network is modified by an anchor-matching mechanism. 
At test time, the pseudo-masks used for online supervision are created using the first annotated frame, called the anchor, and inter-frame visual information instead of the outputs of previous images.

Instead of specifying the segmentation mask as the ground truth of a surgical instrument in the first frame of a sequence, approaches exist that use the 3D pose of the tool as the ground truth.
Instruments are then tracked by a 2D tracker based on a generalized hough transformation and a probabilistic segmentation model that makes the probability of a pixel belonging to a tool dependent on the segmentation of the previous video frame.
The results of this 2D tracker are combined with a 3D tracking algorithm to estimate the 3D pose of the tool~\cite{du2016combined}.
Further methods limit the relevant region considered for the segmentation of a frame to the results of previous images, tracking the joint of the surgical instrument over time~\cite{zhang2020object}.
Similar approaches incorporate the results of previous frames in a related manner, identifying candidate bounding boxes in each image, but searching for an instrument exclusively in the boxes related to the position of the tool in the last frame \cite{agustinos20152d}. 
The proposed segmentation algorithm from Amini Khoiy et al. \cite{amini2016automatic} also works only within the region of interest (RoI) identified as relevant in the previous frame and propagated to the current one.

\subsubsection{Inference Speed}
\label{semantic_segmentation:temporal_info:inference_speed}

As with single-frame segmentation, processing speed at test time plays a key role in the applicability of the proposed methods in real-world scenarios. 
In total, we identified nine papers that specifically stated the speed of their processes in FPS, presented together with various characteristics in Table~\ref{tab:sem_seg:temporal_info:inference_speed}.

\subsubsection{Attention Mechanisms}
\label{semantic_segmentation:temporal_info:attention_mechanisms}

In total, six papers used attention-based techniques in their work.
A popular approach is to leverage attention mechanisms within the network architecture.
This can be realized by using squeeze \& excitation (SE) techniques built into attention-dense recurrent convolutional blocks (ADRCB) in the encoder and in the decoder and into a context fusion block (CFB) between the two parts~\cite{yang2022drr}.
The work of Islam et al. \cite{islam2021st} relies on improved SE blocks, executing skip competitive spatial and channel squeeze \& excitation (SC-scSE) blocks several times in succession in the decoder path.
Another approach is the incorporation of attention-based techniques in multi-frame feature aggregation (MFFA) blocks placed between the encoder and decoder \cite{lin2021multi}.
In their presented DMNet, Wang et al. \cite{wang2021efficient} use RNN and self-attention mechanisms in their local memory module to optimally incorporate current temporal information as well as to better localize the currently viewed frame into the overall temporal structure.
Another possible application is presented by Jin et al. \cite{jin2019incorporating}, where they describe an attention-guided module placed in their network architecture that is used multiple times in a pyramid-like structure.
In the meta-training flow of their method, Zhao et al. \cite{zhao2021anchor} use attention-inspired techniques in the anchor matching mechanism, identifying noisy pixels for foreground and background.

\section{Instance Segmentation Methods}
\label{instance_segmentation_methods}

\begin{table*}[tbph]
	\centering
	\caption{Publications that have contributed in the area of single-image instance segmentation of surgical instruments. For each publication, the type of methodology (SV = supervised), whether or not attention mechanisms (Att.) were used in the methodology, and the datasets used are indicated. The works are grouped according to the used segmentation types visualized in Figure~\ref{fig:seg_types_ins_seg_single_frames}.}
	\begin{tabularx}{\textwidth}{Xp{1.8cm}P{0.4cm}P{0.3cm}P{0.3cm}P{0.3cm}P{0.3cm}P{0.4cm}P{0.7cm}P{0.7cm}P{0.7cm}P{0.7cm}P{0.5cm}P{0.5cm}}
		\toprule
		\multirow{3}{*}{Publication}
		& \multirow{3}{*}{\begin{tabular}{l}\\Super- \\ vision\end{tabular}}
		& \multirow{3}{*}{Att.}
		& \multicolumn{10}{c}{Used Datasets} \\
		\cmidrule(llrr){4-14}
		&
		& 
		& \multicolumn{4}{c}{EndoVis}
		& \multirow{2}[3]{*}{\shortstack{Lap. \\ I2I}}
		& \multirow{2}[3]{*}{\shortstack{Sinus \\ Surg.}}
		& \multirow{2}[3]{*}{\shortstack{UCL \\ dVRK}}
		& \multirow{2}[3]{*}{\shortstack{Kvasir \\ Inst.}}
		& \multirow{2}[3]{*}{\shortstack{Robo- \\ Tool}}
		& \multirow{2}[3]{*}{\shortstack{Pri- \\ vate}}
		& \multirow{2}[3]{*}{\shortstack{Ot- \\ her}} \\
		\cmidrule(llrr){4-7}
		&
		& 
		& 15
		& 17
		& 18
		& 19
		& & & & & & & \\
		\midrule
            Shimgekar et al. (2021) \cite{shimgekar2021voice} & SV & & & & & & & & & & & \ding{51} &  \\
            \cdashlinelr{1-14} 
            Baby et al. (2023) \cite{baby2023forks} & SV & \ding{51} & & \ding{51} & \ding{51} & & & & & & & &  \\
		Cer{\'o}n et al. (2021) \cite{ceron2021assessing} & SV & \ding{51} & & & & \ding{51} & & & & & & &  \\
		Cer{\'o}n et al. (2022) \cite{ceron2022real} & SV & \ding{51} & & \ding{51} & & \ding{51} & & & & & & &  \\
		Kitaguchi et al. (2022) \cite{kitaguchi2022development} & SV & & & & & & & & & & & \ding{51} &  \\
		Kitaguchi et al. (2022) \cite{kitaguchi2022limited} & SV & & & & & & & & & & & \ding{51} &  \\
            Kurmann et al. (2021) \cite{kurmann2021mask} & SV & & & \ding{51} & & & & & & & & &  \\
            \cdashlinelr{1-14}
		Kletz et al. (2019) \cite{kletz2019identifying} & SV & & & & & & & & & & & \ding{51} &  \\
            \cdashlinelr{1-14}
		Kong et al. (2021) \cite{kong2021accurate} & SV & & & \ding{51} & & & & & & & & \ding{51} &  \\
		Ro{\ss} et al. (2021) \cite{ross2021comparative} & SV & \ding{51} & & & & \ding{51} & & & & & & &  \\
		\cdashlinelr{1-14}
		Sun et al. (2022) \cite{sun2022parallel} & SV & \ding{51} & & \ding{51} & & & & & & & & &  \\
		\bottomrule
	\end{tabularx}
	\label{tab:seg:instance_seg:single_frame}
\end{table*}

In the following, we describe contributions regarding instance-based segmentation of surgical instruments in endoscopic video images.
The structure is analogous to the one presented in Figure~\ref{semantic_seg:structure} regarding semantic segmentation methods, with domain adaptation being the only deviation from this as no instance-based works could be identified for this category.
We first consider approaches focussing on the processing of individual frames, followed by methods incorporating temporal information.

\subsection{Single Frame Segmentation} 
\label{instance_segmentation:single_frame_seg}

Within the following, the developments regarding single-image segmentation using instance-based methods are subdivided concerning the topics segmentation type, learning strategies, inference speed, and use of attention mechanisms.

\subsubsection{Segmentation Type}
\label{instance_seg:single_frame:seg_type}

\begin{figure}[!b]
\includegraphics[width=0.49\textwidth]{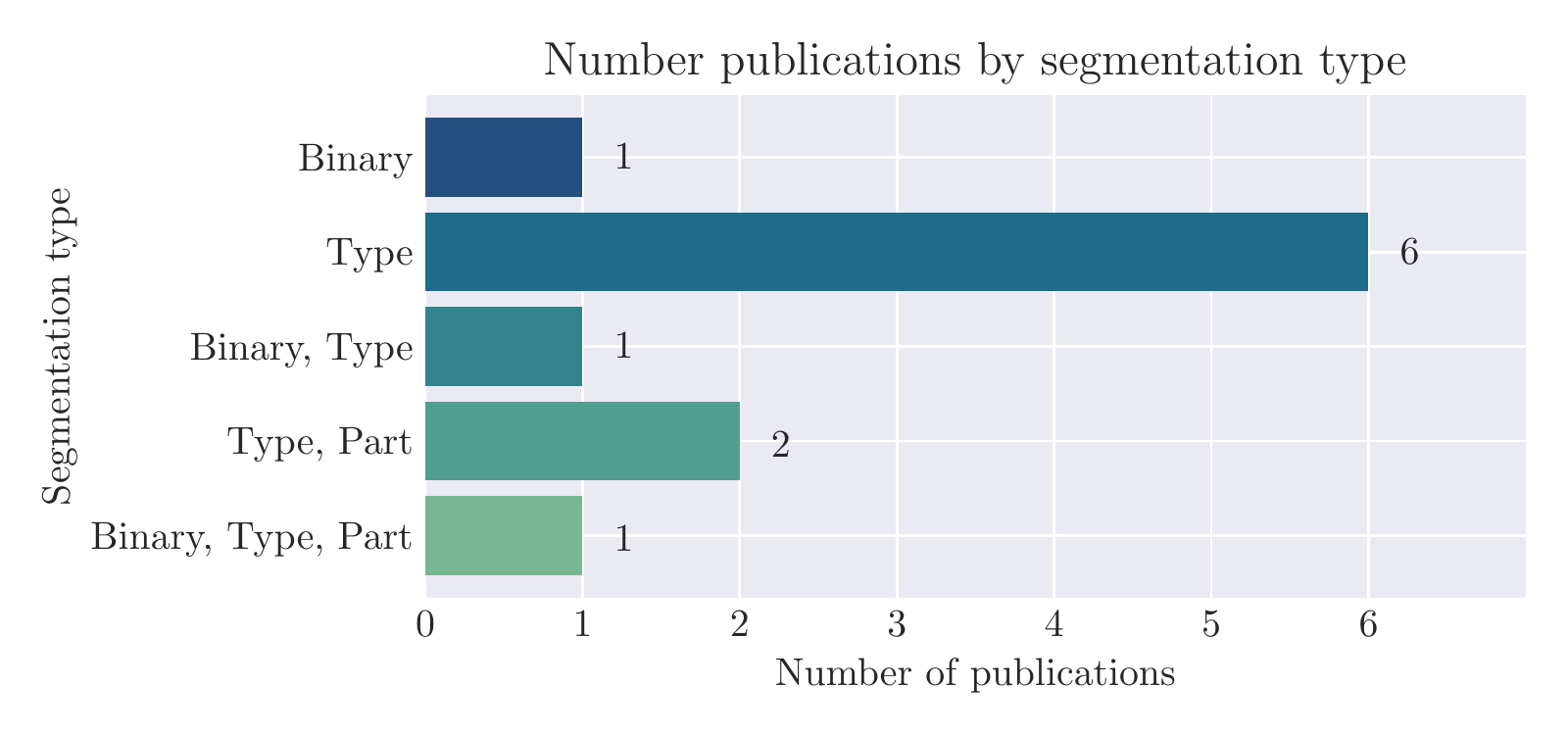}
\caption{Number of relevant publications per segmentation type for instance single image segmentation.}
\label{fig:seg_types_ins_seg_single_frames}
\end{figure}

Figure~\ref{fig:seg_types_ins_seg_single_frames} illustrates the number of publications per segmentation type and for each existing combination of them.
A representation of all papers based on this partitioning can be found in Table~\ref{tab:seg:instance_seg:single_frame}.

\subsubsection{Learning Strategies}
\label{instance_seg:single_frame:learning_strategy}

All work in this area follows the supervised learning approach.
However, the methodologies of these contributions can be subdivided into topic-related groups. 
Kurmann et al. \cite{kurmann2021mask} describe a newly developed architecture based on a shared encoder that forwards the detected features of each layer to the corresponding layers of two decoders.
The main idea is to first segment the instruments in the image and then classify the instances thus determined. 

In addition, recent research has focused on improving the quality of existing network architectures by incorporating novel developments, for example, by adding advanced modules to the backbone and the feature pyramid network (FPN) of an established architecture~\cite{ceron2021assessing, ceron2022real}.
Sun et al. \cite{sun2022parallel} also modify the backbone part of a network by adding novel modules, although the basic architecture has a different structure from the ones mentioned above.
An improvement of existing architecture is also presented by Kong et al. \cite{kong2021accurate}, which realizes this by employing an evolved region proposal network part and anchor optimization.
Baby et al. \cite{baby2023forks} extend existing instance segmentation architectures with an additional classification module, which they call a multi-scale mask-attended classifier.
Multi-scale mask attention highlights instrument features and suppresses features from background pixels, resulting in improved instrument classification.

In contrast, other research aims to evaluate the performance of established architectures using newly collected data from specific types of operations~\cite{kitaguchi2022development, kletz2019identifying}.
In this context, the ability to generalize to unknown data from different types of operations is also a key challenge~\cite{kitaguchi2022limited}.
An evaluation of various methods on several kinds of operations is also presented by Ro{\ss} et al. \cite{ross2021comparative}, who examine the concepts and quality of participants' submitted developments on the provided data of their organized challenge, especially concerning the criteria of robustness and generalization ability.

\subsubsection{Inference Speed}
\label{instance_seg:single_frame:inference_speed}

A quantitative evaluation of the speed of their approaches at test time is provided by three works, which are presented in Table~\ref{tab:ins_seg:single_frame:inference_speed} together with various characteristics.

\subsubsection{Attention Mechanisms}
\label{instance_seg:single_frame:attention_mechanisms}

\begin{table*}[tbhp]
	\centering
	\caption{Instance segmentation methods for single images that have specified processing speeds at inference time. Indicated are the publication, the speed in FPS, the image resolution, the GPU used as well as the number of graphics cards, and the CPU used and the number of CPUs. The publications are sorted in descending order according to the reported inference speed.}
	\begin{tabularx}{\textwidth}{p{4.5cm}rclclc}
	\toprule
        \multirow{2}{*}{Publication} & \multicolumn{1}{c}{Inf. Speed} & \multirow{2}{*}{Resolution} & \multirow{2}{*}{GPU} & \multirow{2}{*}{$\#$\,GPU} & \multirow{2}{*}{CPU} & \multirow{2}{*}{$\#$\,CPU} \\    
        & \multicolumn{1}{c}{(FPS)} & & & & & \\
        \midrule
        Cer{\'o}n et al. (2021) \cite{ceron2021assessing} & 49 & $540 \times 960$ & Tesla P100 & 1 & Xeon E5-2698 v4 2.2 GHz & 1 \\
        Cer{\'o}n et al. (2022) \cite{ceron2022real} & 24 - 69 & $540 \times 960$ & Tesla P100 & 1 & Xeon E5-2698 v4 2.2 GHz & 1 \\
        Kurmann et al. (2021) \cite{kurmann2021mask} & 15 & $512 \times 640$ & RTX 2080 Ti & 1 & - & - \\
        \bottomrule
	\end{tabularx}
	\label{tab:ins_seg:single_frame:inference_speed}
\end{table*}

The use of attention-based techniques is a central component in the development of instance-based approaches.
One variant is to modify established architectures by using criss-cross attention blocks to capture global contextual information \cite{ceron2021assessing} or by using convolutional block attention modules instead of criss-cross attention blocks~\cite{ceron2022real}.
In addition, attention-oriented techniques are integrated into existing architectures through swin-transformer blocks that act as feature fusion modules localized in the backbone.
The operations within these swin-transformer blocks essentially consist of multi-head window self-attention and multi-head shift window self-attention operations~\cite{sun2022parallel}.
Among the methods submitted by the participating teams in the EndoVis-2019 challenge, two rely on the use of attention-based techniques~\cite{ross2021comparative}.

\subsection{Incorporating Temporal Information}
\label{instance_segmentation:incorporating_temporal_information}

This section presents publications that have provided new findings concerning instance-based segmentation of surgical instruments in endoscopic images and videos by incorporating temporal context.
The structure of this section follows the one described in Chapter~\ref{semantic_segmentation:incorporating_temporal_information}.

\subsubsection{Segmentation Type}
\label{instance_seg:temporal_info:seg_type}

\begin{figure}[!b]
\includegraphics[width=0.49\textwidth]{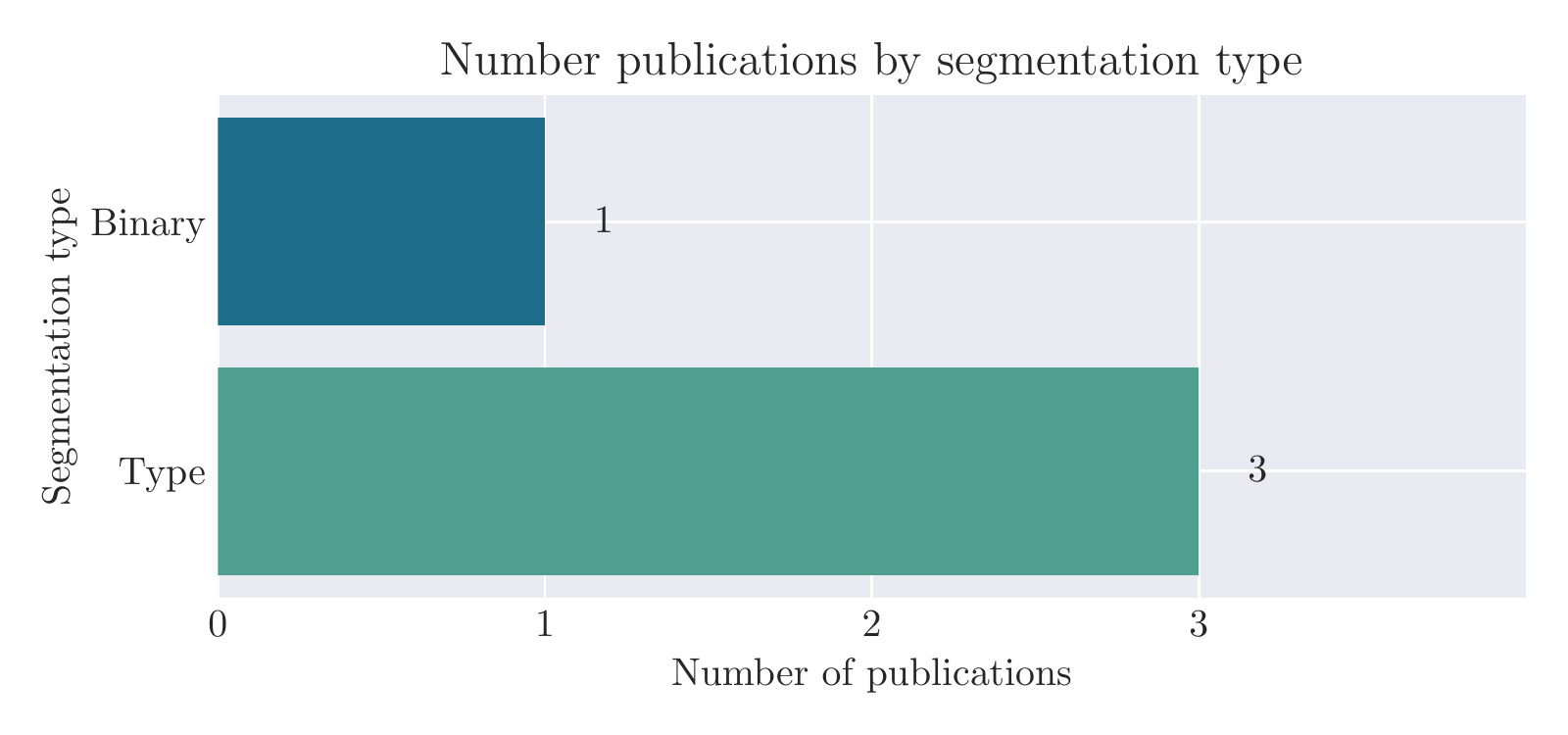}
\caption{Number of relevant publications per segmentation type for instance image segmentation incorporating temporal information.}
\label{fig:seg_types_ins_seg_temporal_info}
\end{figure}

A breakdown of papers by segmentation type is shown in Figure~\ref{fig:seg_types_ins_seg_temporal_info}, and a listing of publications by these categories is given in Table~\ref{tab:seg:instance_seg:temporal_info}.

\subsubsection{Learning Strategies}
\label{instance_seg:temporal_info:learning_strategy}

All the works employ supervised learning techniques.
One approach uses established network architectures that are adapted and extended with respect to the specific use case~\cite{gonzalez2020isinet, lee2020evaluation}.
Other methods rely on several different established architectures, processing the frames of a video sequence in parallel and merging the resulting features in the decoder part of the network~\cite{kanakatte2020surgical}.
Other work describes an approach using the principles of instance-based segmentation techniques to design a novel network architecture that is based on the parallel prediction of instrument types, their localization, and the determination of their unique identities~\cite{zhao2022trasetr}.

\subsubsection{Tracking Approaches}
\label{instance_seg:temporal_info:tracking_approaches}

The types of surgical tool tracking over time can be divided into several groups.
One approach is to use the frame-by-frame results of successive images by applying tracking techniques to improve the segmentation of subsequent images based on this information.
Lee et al. \cite{lee2020evaluation} present a method in which they propagate the predicted coordinates of the bounding boxes of surgical instruments from one frame to the next using the deep simple online and real-time tracker (deepSORT) algorithm.
Another component of the tracking framework is instrument re-identification, which is needed when instruments leave the field of view or are very close to each other~\cite{lee2020evaluation}.
Gonz\'{a}lez et al. \cite{gonzalez2020isinet} describe a method based on a instance segmentation architecture, which they extend by adding a temporal consistency module employing a two-step procedure and thus allows tracking the tools over time.

Other approaches aim at modifying or designing network architectures to process temporal information directly within the model.
Current research achieves this through spatio-temporal LSTM layers that model the relationships of video-level features resulting from a network that processes three-dimensional inputs by providing multiple input images simultaneously.
For the final segmentation prediction, the outputs of the LSTM layers are further processed by a downstream decoder \cite{kanakatte2020surgical}.
In addition to this approach, embedding novel modules into the network architecture is another option for tracking instruments.
Recently, Zhao et al. \cite{zhao2022trasetr} describe two modules explicitly designed to track instruments with huge temporal variations.
To achieve this, the authors incorporate the intermediate results of previous frames as well as those of the currently viewed video image and process this information both in an identity-matching module that follows a two-stage structure and in a contrastive query-based learning approach. 

\begin{table*}[tbph]
	\centering 
	\caption{Publications that have contributed in the area of instance segmentation of surgical instruments by incorporating temporal information. For each publication, the type of methodology (SV = supervised), whether or not attention mechanisms (Att.) were used in the methodology, and the datasets used are indicated. The works are grouped according to the used segmentation types visualized in Figure~\ref{fig:seg_types_ins_seg_temporal_info}.}
	\begin{tabularx}{\textwidth}{p{5.0cm}p{1.8cm}P{0.4cm}P{0.3cm}P{0.3cm}P{0.3cm}P{0.3cm}P{0.4cm}P{0.7cm}P{0.7cm}P{0.7cm}P{0.7cm}P{0.5cm}P{0.5cm}}
		\toprule
		\multirow{3}{*}{Publication}
		& \multirow{3}{*}{\begin{tabular}{l}\\Super- \\ vision\end{tabular}}
		& \multirow{3}{*}{Att.}
		& \multicolumn{10}{c}{Used Datasets} \\
		\cmidrule(llrr){4-14}
		&
		& 
		& \multicolumn{4}{c}{EndoVis}
		& \multirow{2}[3]{*}{\shortstack{Lap. \\ I2I}}
		& \multirow{2}[3]{*}{\shortstack{Sinus \\ Surg.}}
		& \multirow{2}[3]{*}{\shortstack{UCL \\ dVRK}}
		& \multirow{2}[3]{*}{\shortstack{Kvasir \\ Inst.}}
		& \multirow{2}[3]{*}{\shortstack{Robo- \\ Tool}}
		& \multirow{2}[3]{*}{\shortstack{Pri- \\ vate}}
		& \multirow{2}[3]{*}{\shortstack{Ot- \\ her}} \\
		\cmidrule(llrr){4-7}
		&
		& 
		& 15
		& 17
		& 18
		& 19
		& & & & & & & \\
		\midrule
            Kanakatte et al. (2020) \cite{kanakatte2020surgical} & SV & & & & & & & & & & & \ding{51} &  \\
            \cdashlinelr{1-14} 
		Gonz{\'a}lez et al. (2020) \cite{gonzalez2020isinet} & SV & & & \ding{51} & \ding{51} & & & & & & & &  \\
		Lee et al. (2020) \cite{lee2020evaluation} & SV & & & \ding{51} & & & & & & & & \ding{51} &  \\
		Zhao et al. (2022) \cite{zhao2022trasetr} & SV & \ding{51} & & \ding{51} & \ding{51} & & & & & & & & \ding{51} \\
		\bottomrule
	\end{tabularx}
	\label{tab:seg:instance_seg:temporal_info}
\end{table*}

\subsubsection{Inference Speed}
\label{instance_seg:temporal_info:inference_speed}

The processing speed at inference time is stated with 23 FPS by the work of Zhao et al. \cite{zhao2022trasetr}.
The used datasets have a resolution of $540 \times 960$ and $1024 \times 1280$, however, the utilized image size is not explicitly specified. 
A Titan Xp is employed as hardware, but there is no information about the CPU used.

\subsubsection{Attention Mechanisms}
\label{instance_seg:temporal_info:attention_mechanisms}

The method proposed by Zhao et al. \cite{zhao2022trasetr} is based on a transformer module that further processes the low-level features of the input image obtained from an encoder network. 
The module consists of a transformer encoder-decoder network that converts the input image features and query embeddings by applying self- and cross-attention mechanisms to instance embeddings that encode global information about the instruments within the scene.
These are then further processed in the contrastive query learning process as well as in the instance fusion module that generates the final classes, bounding box, and mask predictions from these embeddings.

\section{Further Processing of Segmentations}
\label{further_processing}

In addition to the segmentation of surgical instruments, approaches for further processing the results are described.
Zhang et al. \cite{zhang2023visual} propose an application-oriented method with a specific focus on the endoscope-holding robotic assistant stably holding the endoscope and modifying the position of the endoscopic camera.
The authors describe a surgical instrument tracking control method based on visual tracking features in combination with hand-eye coordination, including the feedback of multiple sensors.
Accurate binary segmentation and localization of surgical instrument tips are performed to use this information for the autonomous control of the endoscopic field of view.
A related approach is published by Gruijthuijsen et al. \cite{gruijthuijsen2022robotic}, whose work also describes the interaction between a human surgeon and an autonomous robotic endoscope holder.
The authors investigate the applicability of semantically rich instructions of the operator to communicate with the robot.
To this end, they explain a technique for determining the positions of instrument tips in which the binary segmentation of surgical tools is of central importance. 
\pagebreak This information is then processed for visual servo control of the robotic endoscope holder.
Li et al. \cite{li2023autonomous} also present an autonomous three-dimensional instrument tracking framework for a robotic laparoscope based on a binocular camera. 
Based on instrument tip segmentation and stereo matching by a semi-global matching algorithm the depth in the current field of view is determined.

Another approach that combines deep-learning-based segmentation of surgical instruments with visual servo control is described by Cheng et al. \cite{cheng2021deep}.
The tips of the instruments are also binarily segmented using an encoder-decoder-based architecture.
Subsequently, a robotic system consisting of a robotic arm and a magnetically anchored and actuated endoscope is autonomously moved based on these results.

Zinchenko and Song \cite{zinchenko2021autonomous} present another method based on instrument tip segmentation.
For the segmentation, rectangular bounding boxes around the instrument tips are first identified by an object detection network, within which the exact structures of the instrument tips are then refined by a network with a different architecture.
A difference between the two previously described works arises from the fact that the visualization in the proposed approach is achieved through a simulation in a three-dimensional virtual reality environment.
The positioning of the surgical robot is based on the current locations of the instruments and the direction of the surgeon's gaze to determine an optimal new camera image center.

\section{Discussion}
\label{discussion}

This chapter discusses the key findings that arise from the literature reviewed.
For this purpose, the numerical results of the review obtained using the method described in Chapter \ref{review_methodology} are evaluated first, and an assessment regarding the possible further development of the research field is presented. 
This is followed by a consideration of currently publicly available datasets described in more detail in Chapter \ref{datasets}.
For this purpose, the strengths and weaknesses of the existing datasets are highlighted, and potentials for future developments in this area are presented.
Subsequently, the identified work is discussed concerning semantic and instance-based segmentation methods.
Again, we describe existing strengths and weaknesses and identify promising opportunities for future developments.

\subsection{Review Methodology}
\label{disucssion:review_results}

Our search terms focus on the publication titles, leading to a large number of highly specific contributions to the topic of this review.
However, there might be more general approaches, which show their results exemplarily also on the public datasets for minimally invasive surgery instrument segmentation and which are not covered by this review, e.g., \cite{mendel2023error, ni2022space, wang2021noisy}.

The number of included papers analyzed in Chapter \ref{method:analysis_of_the_results} and presented graphically in Figure \ref{fig:review_method:publications} show a clear positive trend regarding the considered time period.
On the one hand, this result shows the high relevance of the research area, both in terms of current research topics and practical application areas, and suggests that the research community will continue to grow steadily.
On the other hand, the observations of published works lead to the conclusion that the considered research area still has enormous potential for future development in the coming years.

\subsection{Public Datasets}
\label{disucssion:datasets}

The availability of high-quality and publicly available datasets represents a central component for the development and validation of novel methods.
Furthermore, such datasets provide the basis for reproducible results, which is also essential for the traceability of scientific findings and for establishing the plausibility of results.
The segmentation of surgical instruments poses a particular challenge in this context since this application requires polygonal lines to be drawn around the relevant regions within an image to achieve pixel-precise segmentation ground truths masks, requiring a great deal of manual effort, which involves both time and subsequent financial drawbacks.
Based on these reasons, it can be assumed that the number of freely accessible datasets is rather limited despite the increasingly active research in this area.
Within the 123 relevant publications identified in Chapter \ref{method:analysis_of_the_results}, only nine different freely available datasets providing endoscopic images or videos are employed, as shown in Figure \ref{fig:datasets:usage}.
It can further be seen here that the number of times these datasets are used varies greatly.
The high number of publications that have used these datasets and the fact that the respective datasets of the four MICCAI challenges are ranked among the six most used datasets suggests the high popularity of these data.
Furthermore, a correlation can be observed between the timing of dataset publications within the MICCAI challenges and the successfully submitted number of publications depicted in Figure \ref{fig:review_method:publications}, where the timing of the first publicly accessible dataset provided within the scope of the MICCAI in 2015 conference can be considered as a starting point for an increasingly active research community.
These results further demonstrate the enormous impact that the challenges organized and conducted as part of the annual MICCAI conferences have on the development of new methods.

However, Figure \ref{fig:datasets:usage} also shows that a total of 40 different publications use self-collected and annotated data to develop and validate their methods, ranking the group of private datasets second among the most commonly used datasets.
This development is associated with several disadvantages since results generated in this way are not reproducible, verification of the results is not possible, and further research based on the used data is excluded for other research groups.
Furthermore, the results obtained based on data collected in this way cannot be compared with the results of other approaches which use different data, which makes it much more difficult to classify a presented method about the current state of research.
To improve these drawbacks, we would like to encourage future researchers to use publicly accessible and comparable datasets to develop and validate their approaches or to disclose the collected and self-labeled data, making them accessible to other researchers.
Having said this, we are aware of legitimate issues regarding data security of personal medical data and financial considerations in third-party funded projects involving the medical device industry.

The datasets shown in Figure \ref{fig:datasets:usage} and described in Chapter \ref{datasets} also provide information about current shortcomings and possible future developments.
Regarding the data based on manually created annotations, it can be seen that in the creation of more recent datasets, there is a stronger focus on incorporating challenges from real-world operations to increase the robustness of developed algorithms.
Furthermore, the generalization capability of developed methods to other types of operations or operations performed in other institutions is becoming increasingly important.
This development results in the EndoVis-2019 dataset, which currently corresponds to the largest freely accessible annotated dataset used in the publications considered.
In the coming years, this trend will surely increase, and it will be exciting to see how the structure and design, as well as the focus on the various real-world challenges, will evolve in the coming years concerning the provision of new datasets.

In addition to traditional approaches in terms of collecting and labeling data, currently published datasets are using other approaches that represent an interesting trend.
As described in \ref{datasets}, current work explores methods to minimize manual annotation effort by generating synthetic training data, as presented in the Laparoscopic I2I Translation and RoboTool datasets.
Also, the procedure used to generate the UCL dVRK dataset based on movements of the surgical instruments leads to a significantly lower manual label effort and provides a variety of promising opportunities for future developments.
Again, there is great potential for designing new datasets in the coming years, supporting and encouraging research and development of new methods in this area.

\subsection{Semantic Segmentation Methods}
\label{disucssion:semantic_segmentation}

In the area of semantic segmentation, it can be observed that the majority of publications concern single-frame segmentation of surgical instruments in endoscopic images, ignoring the temporal context.
One reason may be that frame-by-frame segmentation is more in line with the conventional development process, and most available methods are designed for this use case.
Furthermore, methods that consider temporal information in processing have to deal with unlabeled images since usually not every video sequence frame is annotated.
Novel approaches have to deal with these challenges, resulting in increased complexity of the development process.
However, the consideration of successive results in processing offers great potential for new methods.

Regarding the type of segmentation, most methods are based on the binary distinction between instrument and background, regardless of whether approaches for single-frame segmentation or for tracking surgical instruments over time are considered.
This is reasonable at first glance, as it allows distinguishing the surgical tools from the background, but upon closer examination introduces some shortcomings.
For applications that further process these segmentations and are made for real-world scenarios, distinguishing the different types of instruments is often of great relevance.
Usually, different types of instruments have varying fields of application and appear only in certain phases of an operation.
In addition, the behavior of downstream algorithms that make decisions based on this information is often directly affected by the type or part of the instrument. 
Segmentation at the level of instrument parts, however, allows us to determine the detailed position of an instrument in an image by its shaft and tip and to use this information to track tools over time.

With a few exceptions published in the early years of the considered time period, the vast majority of approaches use deep-learning-based techniques for their methods.
Within this group of approaches, most methods apply a supervised learning strategy both for frame-by-frame-based algorithms and concerning tracking methods.
Here, the ground truth target data is created by manual labeling since domain-specific knowledge about the existing data is often required.
This finding is not particularly surprising since supervised learning is also the most commonly used method in many other application domains, as it represents an intuitive and relatively well-studied method for learning a model compared to other learning strategies.
However, the reliance on annotated datasets presents a variety of challenges.
First, the creation of manual labels is time-consuming and potentially costly, as this process is typically performed by experts.
Second, a reasonably large amount of annotated data is usually required to obtain accurate results.

Dealing with these challenges is playing an increasingly important role in current research. 
Parallels can be identified between currently published methodological approaches and the structure and nature of recently published datasets since a trend towards fewer manually created annotations can be observed.
The semi-supervised learning strategy shows great potential for developing new methods since the data to be processed often consists of video recordings, of which only a small fraction is manually annotated and a much larger fraction of unlabeled video images is available.
In addition, promising results have been achieved regarding weakly supervised learning in recent years, which can significantly reduce the effort of manual annotation.
In both single-image segmentation methods and, in particular, approaches that incorporate the temporal component, some elements of completely unsupervised learning strategies can also be identified in recent work, opening up new possibilities for further developments to reduce the annotation effort.
In addition, a central challenge of methods based on annotated data is the transfer to other domains, showing, for example, different types of operations or coming from other institutions.
Current research concerning domain adaptation is yielding very promising results, often based on the incorporation of synthetic data, which also reduces the annotation effort.
We think this research areas have great potential for further development and will show an increasing impact in the coming years.

Real-time capable processing at inference time is essential for methods to be applicable in real-world scenarios and medical devices.
However, awareness of this issue is low in most of the approaches. 
Only a small fraction of the frame-by-frame segmentation methods focus on this issue in particular, indicating that generating high-quality results is more important than the fast processing of inputs.
In contrast, fast inference time plays a much more important role for methods that consider temporal information.
Since segmentation methods can only be used in real-world surgical procedures if they are fast and show high-quality results, it will become more and more important to consider both aspects in the future.
Incorporating the temporal component can be a viable approach, as has already been shown in some recent work.

Our review shows that attention-based approaches are equally important in publications on single-image segmentation and in papers on segmentation involving temporal information.
Embedding attention mechanisms into network architecture allows new possibilities that are difficult to realize with solely convolutional-based architectures. In some considered works in this review, this leads to an improvement of the results of the previous state-of-the-art.
The publication of articles using attention-oriented approaches has followed a clear positive trend in recent years, and it can be expected that this trend will continue in the future, making the utilization of attention layers within the network structure an active research topic.

\subsection{Instance Segmentation Methods}
\label{disucssion:instance_segmentation}

In addition to semantic segmentation methods, the instance-based segmentation techniques described in Chapter \ref{instance_segmentation_methods} represent another important group of approaches providing several additional benefits.
Instance segmentation allows each surgical tool in the image to be assigned a unique identifier that remains unchanged over time, which is not easily achieved with semantic training labels.
This feature is particularly useful in surgeries where instruments of the same type occur multiple times in an image to distinguish between these instruments and to track the individual tools independently over time.
In our review, we determined that the number of instance-based papers is nevertheless significantly lower than the number of publications relying on semantic segmentation methods.
Since the above advantages have a significant impact on the practical applicability of the methods and due to the promising results of existing methods, the field of instance-based approaches represents a research area with high potential in the future.

Differences between instance-based and semantic-based methods can be seen concerning the segmentation types, since in contrast to the semantic variants only one approach for instance segmentation is based on binary segmentation only, and most of the works are specialized to distinguish the different types of instruments.
This finding is hardly surprising since the conceptual design of established instance-based network architectures is well suited for this task and highlights one of the strengths of these methods.

Regarding the learning strategies used, the considered instance segmentation methods differ significantly from the semantic ones discussed before.
In single-image processing and in methods involving the temporal component, exclusively supervised learning approaches are used.
Concerning the previously mentioned disadvantages of this learning strategy, the advanced and more efficient processing of annotated data represents a conceivable future research focus.
Another interesting finding is that no synthetically generated or computer-simulated training data is employed in the publications associated with instance segmentation.
Analogous to semantic segmentation, this opens up new possibilities for instance segmentation by including data of this type, thus providing an incentive for further research in this area.

The situation regarding the consideration of processing speed in the development of novel methods is similar to that of semantic segmentation.
It would be desirable if awareness of this aspect would also increase for instance-based approaches besides the primary goal to improve the segmentation quality.

The popularity of attention-based techniques, previously discussed for semantic segmentation methods, continues here.
Some instance-based publications take advantage of diverse attention modules to better incorporate image features, both in frame-by-frame approaches and in sequence processing based on temporal information.
It is expected that also in the field of instance segmentation, the use of attention-based techniques will play a central role in the following years, both as components within conventional convolutional network architectures as well as through the design of completely attention-based networks that do not use convolutional layers at all.

\subsection{Further Processing of Segmentations}
\label{disucssion:further_processing}

In Chapter \ref{further_processing}, we describe papers that further process the segmentation results for controlling a robotic system. These already show the potential existing in high-quality and fast to compute segmentations for practical use in applications based on them.
We do not claim to be exhaustive concerning articles dealing with this topic, as this requires a more detailed search that may not be covered by the search terms used in this review.
However, we want to draw attention to this issue by showing that there are promising ways to further process segmentation results. 

\subsection{Summary of Challenges and Potential Future Work}
\label{disucssion:challenges_future_work}

Our described findings confirm that methods based on the segmentation of surgical tools have a high value for RAMIS, as evidenced by the rapidly growing number of publications in this field.
However, our results also show that many developments rely on self-collected and annotated data, which makes reproducibility of results and further research based on these data impossible.
Moreover, few publicly accessible datasets exist that contain real-world challenges.
Providing synthetic or simulated datasets represents a promising approach for the future to reduce manual annotation efforts.

Most reviewed articles ignore the temporal context, although its inclusion provides the potential for higher-quality results and faster processing speed.
Furthermore, most semantic approaches perform only binary segmentation, which often represents an insufficient level of accuracy for RAMIS.
Most publications are based on supervised deep-learning techniques, which means heavy reliance on manually labeled data.  
Although real-time processing is an essential requirement for using a method in RAMIS, the focus is often on producing high-quality results, and fast processing of inputs is of secondary importance.
A comparison of segmentation types shows that current research often relies on semantic methods.
Nevertheless, instance-based approaches have advantages, such as unique identification and tracking of surgical instruments over time, even if multiple tools of the same type are present in an image.
 
\section{Conclusion}
\label{conclusion}

In this paper, we provided an overview of the current state of the art concerning semantic and instance-based segmentation of surgical instruments in endoscopic images and videos.
To do this, we identified datasets commonly used for method development and validation and quantified their use in the literature.
We further conducted a systematic search, divided the results concerning semantic and instance-based segmentation methods, and divided the papers within each segmentation type into single-image segmentation and approaches that use temporal information.
In addition, we presented approaches for further processing the semantic segmentation results.
We provide a discussion regarding the findings of the reviewed literature, identifying existing shortcomings and highlighting the potential for future developments.

Assisting surgeons in MIS through robotic assistance represents an active research area of increasing importance.
In this context, purely vision-based methods for localizing surgical instruments in endoscopic images and videos provide the foundation for many procedures with instrument segmentation enabling accurate prediction and, due to technical and methodological advances, can be performed in real time.

\section{Acknowledgements}
\label{acknowledgements}

Funding: This work was supported by the Bavarian Research Foundation [BFS, grant number AZ-1506-21], and the Bavarian Academic Forum [BayWISS]. Open Access funded by Ostbayerische Technische Hochschule Regensburg.

\bibliographystyle{unsrtnat}

\bibliography{Bibliography}

\begin{thebibliography}{152}
\providecommand{\natexlab}[1]{#1}
\providecommand{\url}[1]{\texttt{#1}}
\expandafter\ifx\csname urlstyle\endcsname\relax
  \providecommand{\doi}[1]{doi: #1}\else
  \providecommand{\doi}{doi: \begingroup \urlstyle{rm}\Url}\fi

\bibitem[Darzi and Mackay(2002)]{darzi2002recent}
Ara Darzi and Sean Mackay.
\newblock Recent advances in minimal access surgery.
\newblock \emph{BMJ}, 324\penalty0 (7328):\penalty0 31--34, 2002.
\newblock \url{https://doi.org/10.1136/bmj.324.7328.31}.

\bibitem[Hammad et~al.(2019)Hammad, Wirries, Ardeshiri, Nikiforov, and
  Geiger]{hammad2019open}
Ahmed Hammad, Andr{\'e} Wirries, Ardavan Ardeshiri, Olexandr Nikiforov, and
  Florian Geiger.
\newblock Open versus minimally invasive tlif: literature review and
  meta-analysis.
\newblock \emph{Journal of Orthopaedic Surgery and Research}, 14\penalty0 (1),
  2019.
\newblock \url{https://doi.org/10.1186/s13018-019-1266-y}.

\bibitem[de~Rooij et~al.(2018)de~Rooij, van Hilst, Boerma, van Dam, van Eijck,
  Gerhards, Klaase, Kazemier, Luyer, Busch, and Besselink]{de2019minimally}
T.~de~Rooij, J.~van Hilst, D.~Boerma, R.~van Dam, C.~van Eijck, M.~Gerhards,
  J.~Klaase, G.~Kazemier, M.~Luyer, O.~Busch, and M.~Besselink.
\newblock Minimally invasive versus open distal pancreatectomy (leopard): a
  multicenter patient-blinded randomized controlled trial.
\newblock \emph{Annals of Surgery}, 20\penalty0 (1):\penalty0 S293--S294, 2018.
\newblock \url{https://doi.org/10.1016/j.hpb.2018.06.280}.

\bibitem[van~der Sluis et~al.(2019)van~der Sluis, van~der Horst, May,
  Schippers, Brosens, Joore, Kroese, Mohammad, Mook, Vleggaar, Rinkes, Ruurda,
  and van Hillegersberg]{sluis2019robot}
Pieter~C. van~der Sluis, Sylvia. van~der Horst, Anne~M. May, Carlo Schippers,
  Lodewijk A.~A. Brosens, Hans C.~A. Joore, Christiaan~C. Kroese, Nadia~Haj
  Mohammad, Stella Mook, Frank~P. Vleggaar, Inne H. M.~Borel Rinkes, Jelle~P.
  Ruurda, and Richard van Hillegersberg.
\newblock Robot-assisted minimally invasive thoracolaparoscopic esophagectomy
  versus open transthoracic esophagectomy for resectable esophageal cancer: A
  randomized controlled trial.
\newblock \emph{Annals of Surgery}, 269\penalty0 (4):\penalty0 621--630, 2019.
\newblock \url{https://doi.org/10.1097/sla.0000000000003031}.

\bibitem[Fuchs(2002)]{fuchs2002minimally}
K.~H. Fuchs.
\newblock Minimally invasive surgery.
\newblock \emph{Endoscopy}, 34\penalty0 (2):\penalty0 154--159, 2002.
\newblock \url{https://doi.org/10.1055%2Fs-2002-19857}.

\bibitem[Haidegger et~al.(2022)Haidegger, Speidel, Stoyanov, and
  Satava]{haidegger2022robot}
Tamas Haidegger, Stefanie Speidel, Danail Stoyanov, and Richard~M Satava.
\newblock Robot-assisted minimally invasive surgery {\textendash} surgical
  robotics in the data age.
\newblock \emph{Proceedings of the IEEE}, 110\penalty0 (7):\penalty0 835--846,
  2022.
\newblock \url{https://doi.org/10.1109/jproc.2022.3180350}.

\bibitem[Maier{-}Hein et~al.(2022)Maier{-}Hein, Eisenmann, Sarikaya,
  M{\"{a}}rz, Collins, Malpani, Fallert, Feussner, Giannarou, Mascagni,
  Nakawala, Park, Pugh, Stoyanov, Vedula, Cleary, Fichtinger, Forestier,
  Gibaud, Grantcharov, Hashizume, Heckmann{-}N{\"{o}}tzel, Kenngott, Kikinis,
  M{\"{u}}ndermann, Navab, Onogur, Ro{\ss}, Sznitman, Taylor, Tizabi, Wagner,
  Hager, Neumuth, Padoy, Collins, Gockel, Goedeke, Hashimoto, Joyeux, Lam,
  Leff, Madani, Marcus, Meireles, Seitel, Teber, {\"{U}}ckert,
  M{\"{u}}ller{-}Stich, Jannin, and Speidel]{maier2022surgical}
Lena Maier{-}Hein, Matthias Eisenmann, Duygu Sarikaya, Keno M{\"{a}}rz, Toby
  Collins, Anand Malpani, Johannes Fallert, Hubertus Feussner, Stamatia
  Giannarou, Pietro Mascagni, Hirenkumar Nakawala, Adrian Park, Carla~M. Pugh,
  Danail Stoyanov, S.~Swaroop Vedula, Kevin Cleary, Gabor Fichtinger, Germain
  Forestier, Bernard Gibaud, Teodor~P. Grantcharov, Makoto Hashizume, Doreen
  Heckmann{-}N{\"{o}}tzel, Hannes~G{\"{o}}tz Kenngott, Ron Kikinis, Lars
  M{\"{u}}ndermann, Nassir Navab, Sinan Onogur, Tobias Ro{\ss}, Raphael
  Sznitman, Russell~H. Taylor, Minu~Dietlinde Tizabi, Martin Wagner, Gregory~D.
  Hager, Thomas Neumuth, Nicolas Padoy, Justin Collins, Ines Gockel, Jan
  Goedeke, Daniel~A. Hashimoto, Luc Joyeux, Kyle Lam, Daniel~Richard Leff, Amin
  Madani, Hani~J. Marcus, Ozanan~R. Meireles, Alexander Seitel, Dogu Teber,
  Frank {\"{U}}ckert, Beat~P. M{\"{u}}ller{-}Stich, Pierre Jannin, and Stefanie
  Speidel.
\newblock Surgical data science {\textendash} from concepts toward clinical
  translation.
\newblock \emph{Medical Image Analysis}, 76:\penalty0 102306, 2022.
\newblock \url{https://doi.org/10.1016/j.media.2021.102306}.

\bibitem[Bouarfa et~al.(2011)Bouarfa, Akman, Schneider, Jonker, and
  Dankelman]{bouarfa2012vivo}
Loubna Bouarfa, Oytun Akman, Armin Schneider, Pieter~P Jonker, and Jenny
  Dankelman.
\newblock In-vivo real-time tracking of surgical instruments in endoscopic
  video.
\newblock \emph{Minimally Invasive Therapy \& Allied Technologies}, 21\penalty0
  (3):\penalty0 129--134, 2011.
\newblock \url{https://doi.org/10.3109/13645706.2011.580764}.

\bibitem[Mamone et~al.(2017)Mamone, Viglialoro, Cutolo, Cavallo, Guadagni, and
  Ferrari]{Mamone2017RobustLI}
Virginia Mamone, Rosanna~Maria Viglialoro, Fabrizio Cutolo, Filippo Cavallo,
  Simone Guadagni, and Vincenzo Ferrari.
\newblock Robust laparoscopic instruments tracking using colored strips.
\newblock In \emph{2017 4th International Conference on Augmented Reality,
  Virtual Reality, and Computer Graphics (AVR)}, volume 10325 of \emph{Lecture
  Notes in Computer Science}, pages 129--143. Springer International
  Publishing, 2017.
\newblock \url{https://doi.org/10.1007/978-3-319-60928-7_11}.

\bibitem[Sorriento et~al.(2019)Sorriento, Porfido, Mazzoleni, Calvosa, Tenucci,
  Ciuti, and Dario]{sorriento2019optical}
Angela Sorriento, Maria~Bianca Porfido, Stefano Mazzoleni, Giuseppe Calvosa,
  Miria Tenucci, Gastone Ciuti, and Paolo Dario.
\newblock Optical and electromagnetic tracking systems for biomedical
  applications: A critical review on potentialities and limitations.
\newblock \emph{IEEE Reviews in Biomedical Engineering}, 13:\penalty0 212--232,
  2019.
\newblock \url{https://doi.org/10.1109%2Frbme.2019.2939091}.

\bibitem[Wang et~al.(2022{\natexlab{a}})Wang, Sun, Liu, and Gu]{wang2022visual}
Yan Wang, Qiyuan Sun, Zhenzhong Liu, and Lin Gu.
\newblock Visual detection and tracking algorithms for minimally invasive
  surgical instruments: {A} comprehensive review of the state-of-the-art.
\newblock \emph{Robotics and Autonomous Systems}, 149:\penalty0 103945,
  2022{\natexlab{a}}.
\newblock \url{https://doi.org/10.1016/j.robot.2021.103945}.

\bibitem[Anteby et~al.(2021)Anteby, Horesh, Soffer, Zager, Barash, Amiel,
  Rosin, Gutman, and Klang]{anteby2021deep}
Roi Anteby, Nir Horesh, Shelly Soffer, Yaniv Zager, Yiftach Barash, Imri Amiel,
  Danny Rosin, Mordechai Gutman, and Eyal Klang.
\newblock Deep learning visual analysis in laparoscopic surgery: a systematic
  review and diagnostic test accuracy meta-analysis.
\newblock \emph{Surgical Endoscopy}, 35\penalty0 (4):\penalty0 1521--1533,
  2021.
\newblock \url{https://doi.org/10.1007/s00464-020-08168-1}.

\bibitem[Rivas{-}Blanco et~al.(2021)Rivas{-}Blanco, P{\'{e}}rez{-}del{-}Pulgar,
  Garc{\'{\i}}a{-}Morales, and Mu{\\textasciitilde {n}}oz]{rivas2021review}
Irene Rivas{-}Blanco, Carlos~Jes{\'{u}}s P{\'{e}}rez{-}del{-}Pulgar, Isabel
  Garc{\'{\i}}a{-}Morales, and Victor~F. Mu{\\textasciitilde {n}}oz.
\newblock A review on deep learning in minimally invasive surgery.
\newblock \emph{{IEEE} Access}, 9:\penalty0 48658--48678, 2021.
\newblock \url{https://doi.org/10.1109/ACCESS.2021.3068852}.

\bibitem[Yang et~al.(2020)Yang, Zhao, and Hu]{yang2020image}
Congmin Yang, Zijian Zhao, and Sanyuan Hu.
\newblock Image-based laparoscopic tool detection and tracking using
  convolutional neural networks: a review of the literature.
\newblock \emph{Computer Assisted Surgery}, 25\penalty0 (1):\penalty0 15--28,
  2020.
\newblock \url{https://doi.org/10.1080%2F24699322.2020.1801842}.

\bibitem[Qiu et~al.(2019)Qiu, Li, and Ren]{qiu2019real}
Liang Qiu, Changsheng Li, and Hongliang Ren.
\newblock Real-time surgical instrument tracking in robot-assisted surgery
  using multi-domain convolutional neural network.
\newblock \emph{Healthcare Technology Letters}, 6\penalty0 (6):\penalty0
  159--164, 2019.
\newblock \url{https://doi.org/10.1049%2Fhtl.2019.0068}.

\bibitem[Zhao et~al.(2019)Zhao, Chen, Voros, and Cheng]{zhao2019real}
Zijian Zhao, Zhaorui Chen, Sandrine Voros, and Xiaolin Cheng.
\newblock Real-time tracking of surgical instruments based on spatio-temporal
  context and deep learning.
\newblock \emph{Computer Assisted Surgery}, 24\penalty0 (sup1):\penalty0
  20--29, 2019.
\newblock \url{https://doi.org/10.1080%2F24699322.2018.1560097}.

\bibitem[Huang et~al.(2022{\natexlab{a}})Huang, Nguyen, Wang, Wang, Mayer,
  Tuch, Vyas, Giannarou, and Elson]{huang2022simultaneous}
Baoru Huang, Anh Nguyen, Siyao Wang, Ziyang Wang, Erik Mayer, David Tuch, Kunal
  Vyas, Stamatia Giannarou, and Daniel~S Elson.
\newblock Simultaneous depth estimation and surgical tool segmentation in
  laparoscopic images.
\newblock \emph{IEEE Transactions on Medical Robotics and Bionics}, 4\penalty0
  (2):\penalty0 335--338, 2022{\natexlab{a}}.
\newblock \url{https://doi.org/10.1109%2Ftmrb.2022.3170215}.

\bibitem[Islam et~al.(2019{\natexlab{a}})Islam, Li, and Ren]{islam2019learning}
Mobarakol Islam, Yueyuan Li, and Hongliang Ren.
\newblock Learning where to look while tracking instruments in robot-assisted
  surgery.
\newblock In \emph{International Conference on Medical Image Computing and
  Computer-Assisted Intervention {\textendash} MICCAI 2019}, pages 412--420.
  Springer International Publishing, 2019{\natexlab{a}}.
\newblock \url{https://doi.org/10.1007%2F978-3-030-32254-0_46}.

\bibitem[Islam et~al.(2019{\natexlab{b}})Islam, Atputharuban, Ramesh, and
  Ren]{islam2019real}
Mobarakol Islam, Daniel~Anojan Atputharuban, Ravikiran Ramesh, and Hongliang
  Ren.
\newblock Real-time instrument segmentation in robotic surgery using auxiliary
  supervised deep adversarial learning.
\newblock \emph{IEEE Robotics and Automation Letters}, 4\penalty0 (2):\penalty0
  2188--2195, 2019{\natexlab{b}}.
\newblock \url{https://doi.org/10.1109%2Flra.2019.2900854}.

\bibitem[Jha et~al.(2021{\natexlab{a}})Jha, Ali, Tomar, Riegler, Johansen,
  Johansen, and Halvorsen]{jha2021exploring}
Debesh Jha, Sharib Ali, Nikhil~Kumar Tomar, Michael~A. Riegler, Dag Johansen,
  H{\aa}vard~D. Johansen, and P{\aa}l Halvorsen.
\newblock Exploring deep learning methods for real-time surgical instrument
  segmentation in laparoscopy.
\newblock In \emph{2021 {IEEE} {EMBS} International Conference on Biomedical
  and Health Informatics (BHI)}, pages 1--4. {IEEE}, 2021{\natexlab{a}}.
\newblock \url{https://doi.org/10.1109/BHI50953.2021.9508610}.

\bibitem[Pakhomov and Navab(2020)]{pakhomov2020searching}
Daniil Pakhomov and Nassir Navab.
\newblock Searching for efficient architecture for instrument segmentation in
  robotic surgery.
\newblock In \emph{Medical Image Computing and Computer Assisted Intervention
  {\textendash} MICCAI 2020}, volume 12263 of \emph{Lecture Notes in Computer
  Science}, pages 648--656. Springer International Publishing, 2020.
\newblock \url{https://doi.org/10.1007/978-3-030-59716-0_62}.

\bibitem[Rodrigues et~al.(2022)Rodrigues, Mayo, and
  Patros]{surgical_tool_datasets}
Mark Rodrigues, Michael Mayo, and Panos Patros.
\newblock Surgical tool datasets for machine learning research: {A} survey.
\newblock \emph{International Journal of Computer Vision}, 130\penalty0
  (9):\penalty0 2222--2248, 2022.
\newblock \url{https://doi.org/10.1007/s11263-022-01640-6}.

\bibitem[Bouget et~al.(2017)Bouget, Allan, Stoyanov, and
  Jannin]{bouget2017vision}
David Bouget, Max Allan, Danail Stoyanov, and Pierre Jannin.
\newblock Vision-based and marker-less surgical tool detection and tracking: a
  review of the literature.
\newblock \emph{Medical Image Analysis}, 35:\penalty0 633--654, 2017.
\newblock \url{https://doi.org/10.1016/j.media.2016.09.003}.

\bibitem[Nema and Vachhani(2022)]{nema2022surgical}
Shubhangi Nema and Leena Vachhani.
\newblock Surgical instrument detection and tracking technologies: Automating
  dataset labeling for surgical skill assessment.
\newblock \emph{Frontiers in Robotics and {AI}}, 9, 2022.
\newblock \url{https://doi.org/10.3389/frobt.2022.1030846}.

\bibitem[Bodenstedt et~al.(2018)Bodenstedt, Allan, Agustinos, Du,
  Garc{\'{\i}}a{-}Peraza{-}Herrera, Kenngott, Kurmann, M{\"{u}}ller{-}Stich,
  Ourselin, Pakhomov, Sznitman, Teichmann, Thoma, Vercauteren, Voros, Wagner,
  Wochner, Maier{-}Hein, Stoyanov, and Speidel]{endovis_2015}
Sebastian Bodenstedt, Max Allan, Anthony Agustinos, Xiaofei Du, Luis~C.
  Garc{\'{\i}}a{-}Peraza{-}Herrera, Hannes Kenngott, Thomas Kurmann, Beat~P.
  M{\"{u}}ller{-}Stich, S{\'{e}}bastien Ourselin, Daniil Pakhomov, Raphael
  Sznitman, Marvin Teichmann, Martin Thoma, Tom Vercauteren, Sandrine Voros,
  Martin Wagner, Pamela Wochner, Lena Maier{-}Hein, Danail Stoyanov, and
  Stefanie Speidel.
\newblock Comparative evaluation of instrument segmentation and tracking
  methods in minimally invasive surgery.
\newblock \emph{CoRR}, abs/1805.02475, 2018.
\newblock \url{http://arxiv.org/abs/1805.02475}.

\bibitem[Allan et~al.(2019)Allan, Shvets, Kurmann, Zhang, Duggal, Su, Rieke,
  Laina, Kalavakonda, Bodenstedt, Garc{\'{\i}}a{-}Peraza, Li, Iglovikov, Luo,
  Yang, Stoyanov, Maier{-}Hein, Speidel, and Azizian]{endovis_2017}
Max Allan, Alexey Shvets, Thomas Kurmann, Zichen Zhang, Rahul Duggal,
  Yun{-}Hsuan Su, Nicola Rieke, Iro Laina, Niveditha Kalavakonda, Sebastian
  Bodenstedt, Luis~C. Garc{\'{\i}}a{-}Peraza, Wenqi Li, Vladimir Iglovikov,
  Huoling Luo, Jian Yang, Danail Stoyanov, Lena Maier{-}Hein, Stefanie Speidel,
  and Mahdi Azizian.
\newblock 2017 robotic instrument segmentation challenge.
\newblock \emph{CoRR}, abs/1902.06426, 2019.
\newblock \url{http://arxiv.org/abs/1902.06426}.

\bibitem[Allan et~al.(2020)Allan, Kondo, Bodenstedt, Leger, Kadkhodamohammadi,
  Luengo, Fuentes{-}Hurtado, Flouty, Mohammed, Pedersen, Kori, Varghese,
  Krishnamurthi, Rauber, Mendel, Palm, Bano, Saibro, Shih, Chiang, Zhuang,
  Yang, Iglovikov, Dobrenkii, Reddiboina, Reddy, Liu, Gao, Unberath, Azizian,
  Stoyanov, Maier{-}Hein, and Speidel]{endovis_2018}
Max Allan, Satoshi Kondo, Sebastian Bodenstedt, Stefan Leger, Rahim
  Kadkhodamohammadi, Imanol Luengo, F{\'{e}}lix Fuentes{-}Hurtado, Evangello
  Flouty, Ahmed~Kedir Mohammed, Marius Pedersen, Avinash Kori, Alex Varghese,
  Ganapathy Krishnamurthi, David Rauber, Robert Mendel, Christoph Palm, Sophia
  Bano, Guinther Saibro, Chi{-}Sheng Shih, Hsun{-}An Chiang, Juntang Zhuang,
  Junlin Yang, Vladimir Iglovikov, Anton Dobrenkii, Madhu Reddiboina, Anubhav
  Reddy, Xingtong Liu, Cong Gao, Mathias Unberath, Mahdi Azizian, Danail
  Stoyanov, Lena Maier{-}Hein, and Stefanie Speidel.
\newblock 2018 robotic scene segmentation challenge.
\newblock \emph{CoRR}, abs/2001.11190, 2020.
\newblock \url{https://arxiv.org/abs/2001.11190}.

\bibitem[Ro{\ss} et~al.(2021)Ro{\ss}, Reinke, Full, Wagner, Kenngott, Apitz,
  Hempe, M{\^{\i}}ndroc{-}Filimon, Scholz, Tran, Bruno, Arbel{\'{a}}ez, Bian,
  Bodenstedt, Bolmgren, S{\'{a}}nchez, Chen, Gonz{\'{a}}lez, Guo, Halvorsen,
  Heng, Hosgor, Hou, Isensee, Jha, Jiang, Jin, Kirta{\c{c}}, Kletz, Leger, Li,
  Maier{-}Hein, Ni, Riegler, Schoeffmann, Shi, Speidel, Stenzel, Twick, Wang,
  Wang, Wang, Wang, Zhang, Zhou, Zhu, Wiesenfarth, Kopp{-}Schneider,
  M{\"{u}}ller{-}Stich, and Maier{-}Hein]{ross2021comparative}
Tobias Ro{\ss}, Annika Reinke, Peter~M. Full, Martin Wagner, Hannes Kenngott,
  Martin Apitz, Hellena Hempe, Diana M{\^{\i}}ndroc{-}Filimon, Patrick Scholz,
  Thuy~Nuong Tran, Pierangela Bruno, Pablo Arbel{\'{a}}ez, Gui{-}Bin Bian,
  Sebastian Bodenstedt, Jon~Lindstr{\"{o}}m Bolmgren, Laura~Bravo
  S{\'{a}}nchez, Hua{-}Bin Chen, Cristina Gonz{\'{a}}lez, Dong Guo, P{\aa}l
  Halvorsen, Pheng{-}Ann Heng, Enes Hosgor, Zeng{-}Guang Hou, Fabian Isensee,
  Debesh Jha, Tingting Jiang, Yueming Jin, Kadir Kirta{\c{c}}, Sabrina Kletz,
  Stefan Leger, Zhixuan Li, Klaus~H. Maier{-}Hein, Zhen{-}Liang Ni, Michael~A.
  Riegler, Klaus Schoeffmann, Ruohua Shi, Stefanie Speidel, Michael Stenzel,
  Isabell Twick, Guotai Wang, Jiacheng Wang, Liansheng Wang, Lu~Wang, Yujie
  Zhang, Yan{-}Jie Zhou, Lei Zhu, Manuel Wiesenfarth, Annette Kopp{-}Schneider,
  Beat~P. M{\"{u}}ller{-}Stich, and Lena Maier{-}Hein.
\newblock Comparative validation of multi-instance instrument segmentation in
  endoscopy: Results of the {ROBUST-MIS} 2019 challenge.
\newblock \emph{Medical Image Analysis}, 70:\penalty0 101920, 2021.
\newblock \url{https://doi.org/10.1016/j.media.2020.101920}.

\bibitem[Pfeiffer et~al.(2019)Pfeiffer, Funke, Robu, Bodenstedt, Strenger,
  Engelhardt, Ro{\ss}, Clarkson, Gurusamy, Davidson, Maier{-}Hein, Riediger,
  Welsch, Weitz, and Speidel]{laparoscopic_I2I_translation}
Micha Pfeiffer, Isabel Funke, Maria~R. Robu, Sebastian Bodenstedt, Leon
  Strenger, Sandy Engelhardt, Tobias Ro{\ss}, Matthew~J. Clarkson, Kurinchi
  Gurusamy, Brian~R. Davidson, Lena Maier{-}Hein, Carina Riediger, Thilo
  Welsch, J{\"{u}}rgen Weitz, and Stefanie Speidel.
\newblock Generating large labeled data sets for laparoscopic image processing
  tasks using unpaired image-to-image translation.
\newblock In \emph{Medical Image Computing and Computer Assisted Intervention
  {\textendash} {MICCAI} 2019}, volume 11768 of \emph{Lecture Notes in Computer
  Science}, pages 119--127. Springer International Publishing, 2019.
\newblock \url{https://doi.org/10.1007/978-3-030-32254-0_14}.

\bibitem[Qin et~al.(2020)Qin, Lin, Li, Bly, Moe, and Hannaford]{qin2020towards}
Fangbo Qin, Shan Lin, Yangming Li, Randall~A Bly, Kris~S Moe, and Blake
  Hannaford.
\newblock Towards better surgical instrument segmentation in endoscopic vision:
  Multi-angle feature aggregationand contour supervision.
\newblock \emph{IEEE Robotics and Automation Letters}, 5\penalty0 (4):\penalty0
  6639--6646, 2020.
\newblock \url{https://doi.org/10.1109/LRA.2020.3009073}.

\bibitem[Lin et~al.(2020)Lin, Qin, Li, Bly, Moe, and
  Hannaford]{lc_gan_image_to_image_translation}
Shan Lin, Fangbo Qin, Yangming Li, Randall~A. Bly, Kris~S. Moe, and Blake
  Hannaford.
\newblock {LC-GAN:} image-to-image translation based on generative adversarial
  network for endoscopic images.
\newblock In \emph{2020 {IEEE}/{RSJ} International Conference on Intelligent
  Robots and Systems ({IROS})}, pages 2914--2920. {IEEE}, 2020.
\newblock \url{https://doi.org/10.1109/IROS45743.2020.9341556}.

\bibitem[Colleoni et~al.(2020)Colleoni, Edwards, and Stoyanov]{dvrk}
Emanuele Colleoni, Philip~J. Edwards, and Danail Stoyanov.
\newblock Synthetic and real inputs for tool segmentation in robotic surgery.
\newblock In \emph{Medical Image Computing and Computer Assisted Intervention
  {\textendash} {MICCAI} 2020}, volume 12263 of \emph{Lecture Notes in Computer
  Science}, pages 700--710. Springer International Publishing, 2020.
\newblock \url{https://doi.org/10.1007/978-3-030-59716-0_67}.

\bibitem[Jha et~al.(2021{\natexlab{b}})Jha, Ali, Emanuelsen, Hicks, Thambawita,
  Garcia{-}Ceja, Riegler, de~Lange, Schmidt, Johansen, Johansen, and
  Halvorsen]{kvasir}
Debesh Jha, Sharib Ali, Krister Emanuelsen, Steven~Alexander Hicks, Vajira
  Thambawita, Enrique Garcia{-}Ceja, Michael~A. Riegler, Thomas de~Lange,
  Peter~Thelin Schmidt, H{\aa}vard~D. Johansen, Dag Johansen, and P{\aa}l
  Halvorsen.
\newblock Kvasir-instrument: Diagnostic and therapeutic tool segmentation
  dataset in gastrointestinal endoscopy.
\newblock In \emph{2021 27th MultiMedia Modeling International Conference
  (MMM)}, volume 12573 of \emph{Lecture Notes in Computer Science}, pages
  218--229. Springer International Publishing, 2021{\natexlab{b}}.
\newblock \url{https://doi.org/10.1007/978-3-030-67835-7_19}.

\bibitem[Garc{\'{\i}}a{-}Peraza{-}Herrera
  et~al.(2021{\natexlab{a}})Garc{\'{\i}}a{-}Peraza{-}Herrera, Fidon, D'Ettorre,
  Stoyanov, Vercauteren, and Ourselin]{robo_tool}
Luis~C. Garc{\'{\i}}a{-}Peraza{-}Herrera, Lucas Fidon, Claudia D'Ettorre,
  Danail Stoyanov, Tom Vercauteren, and S{\'{e}}bastien Ourselin.
\newblock Image compositing for segmentation of surgical tools without manual
  annotations.
\newblock \emph{{IEEE} Transactions on Medical Imaging}, 40\penalty0
  (5):\penalty0 1450--1460, 2021{\natexlab{a}}.
\newblock \url{https://doi.org/10.1109/TMI.2021.3057884}.

\bibitem[Maier{-}Hein et~al.(2014)Maier{-}Hein, Mersmann, Kondermann,
  Bodenstedt, Sanchez, Stock, Kenngott, Eisenmann, and Speidel]{maier2014can}
Lena Maier{-}Hein, Sven Mersmann, Daniel Kondermann, Sebastian Bodenstedt,
  Alexandro Sanchez, Christian Stock, Hannes~G{\"{o}}tz Kenngott, Matthias
  Eisenmann, and Stefanie Speidel.
\newblock Can masses of non-experts train highly accurate image classifiers?
  {\textendash} {A} crowdsourcing approach to instrument segmentation in
  laparoscopic images.
\newblock In \emph{Medical Image Computing and Computer-Assisted Intervention
  {\textendash} {MICCAI} 2014}, volume 8674 of \emph{Lecture Notes in Computer
  Science}, pages 438--445. Springer International Publishing, 2014.
\newblock \url{https://doi.org/10.1007/978-3-319-10470-6_55}.

\bibitem[Bouget et~al.(2015)Bouget, Benenson, Omran, Riffaud, Schiele, and
  Jannin]{bouget2015detecting}
David Bouget, Rodrigo Benenson, Mohamed Omran, Laurent Riffaud, Bernt Schiele,
  and Pierre Jannin.
\newblock Detecting surgical tools by modelling local appearance and global
  shape.
\newblock \emph{{IEEE} Transactions on Medical Imaging}, 34\penalty0
  (12):\penalty0 2603--2617, 2015.
\newblock \url{https://doi.org/10.1109/TMI.2015.2450831}.

\bibitem[Hong et~al.(2020)Hong, Kao, Kuo, Wang, Chang, and
  Shih]{Hong2020CholecSeg8kAS}
W.{-}Y. Hong, Chia{-}Lung Kao, Yi{-}Hung Kuo, J.{-}R. Wang, W.{-}L. Chang, and
  Chi{-}Sheng Shih.
\newblock Cholecseg8k: {A} semantic segmentation dataset for laparoscopic
  cholecystectomy based on cholec80.
\newblock \emph{CoRR}, abs/2012.12453, 2020.
\newblock \url{https://arxiv.org/abs/2012.12453}.

\bibitem[HeiSurf(2021)]{heisurf}
HeiSurf.
\newblock Surgical workflow analysis and full scene segmentation.
\newblock 2021.
\newblock \url{https://www.synapse.org/\#!Synapse:syn25101790/wiki/608802}.

\bibitem[Hasan et~al.(2021{\natexlab{a}})Hasan, Calvet, Rabbani, and
  Bartoli]{hasan2021detection}
Md.~Kamrul Hasan, Lilian Calvet, Navid Rabbani, and Adrien Bartoli.
\newblock Detection, segmentation, and 3d pose estimation of surgical tools
  using convolutional neural networks and algebraic geometry.
\newblock \emph{Medical Image Analysis}, 70:\penalty0 101994,
  2021{\natexlab{a}}.
\newblock \url{https://doi.org/10.1016/j.media.2021.101994}.

\bibitem[Grammatikopoulou et~al.(2021)Grammatikopoulou, Flouty,
  Kadkhodamohammadi, Quellec, Chow, Nehme, Luengo, and Stoyanov]{cadis}
Maria Grammatikopoulou, Evangello Flouty, Abdolrahim Kadkhodamohammadi,
  Gwenol{\'{e}} Quellec, Andre Chow, Jean Nehme, Imanol Luengo, and Danail
  Stoyanov.
\newblock Cadis: Cataract dataset for surgical rgb-image segmentation.
\newblock \emph{Medical Image Analysis}, 71:\penalty0 102053, 2021.
\newblock \url{https://doi.org/10.1016/j.media.2021.102053}.

\bibitem[Wang et~al.(2022{\natexlab{b}})Wang, Lu, Long, Zhong, Cheung, Dou, and
  Liu]{wang2022autolaparo}
Ziyi Wang, Bo~Lu, Yonghao Long, Fangxun Zhong, Tak~Hong Cheung, Qi~Dou, and
  Yunhui Liu.
\newblock Autolaparo: {A} new dataset of integrated multi-tasks for
  image-guided surgical automation in laparoscopic hysterectomy.
\newblock In \emph{Medical Image Computing and Computer Assisted Intervention
  {\textendash} MICCAI 2022}, volume 13437 of \emph{Lecture Notes in Computer
  Science}, pages 486--496. Springer International Publishing,
  2022{\natexlab{b}}.
\newblock \url{https://doi.org/10.1007/978-3-031-16449-1_46}.

\bibitem[Psychogyios et~al.(2023)Psychogyios, Colleoni, van Amsterdam, Li,
  Huang, Li, Jia, Zou, Wang, Liu, Boels, Huo, Sparks, Dasgupta, Granados,
  Ourselin, Xu, Wang, Wu, Bai, Ren, Yamada, Harai, Ishikawa, Hayashi, Simoens,
  DeBacker, Cisternino, Furnari, Mottrie, Ferraguti, Kondo, Kasai, Hirasawa,
  Kim, Lee, Lee, Kong, Fu, Li, An, Krell, Bodenstedt, Ayobi, Perez, Rodriguez,
  Puentes, Arbelaez, Mohareri, and Stoyanov]{Psychogyios2023SARRARP50}
Dimitrios Psychogyios, Emanuele Colleoni, Beatrice van Amsterdam, Chih-Yang Li,
  Shu-Yu Huang, Yuchong Li, Fucang Jia, Baosheng Zou, Guotai Wang, Yang Liu,
  Maxence Boels, Jiayu Huo, Rachel Sparks, Prokar Dasgupta, Alejandro Granados,
  S{\'e}bastien Ourselin, Mengya Xu, An-Chi Wang, Yanan Wu, Long Bai, Hongliang
  Ren, Atsushi Yamada, Yuriko Harai, Yuto Ishikawa, Kazuyuki Hayashi, Jente
  Simoens, Pieter DeBacker, Francesco Cisternino, Gabriele Furnari, Alex
  Mottrie, Federica Ferraguti, Satoshi Kondo, Satoshi Kasai, Kousuke Hirasawa,
  Soohee Kim, Seung~Hyun Lee, Kyu~Eun Lee, Hyoun-Joong Kong, Kui Fu, Chao Li,
  Shan An, Stefanie Krell, Sebastian Bodenstedt, Nicol{\'a}s Ayobi, Alejandra
  Perez, Santiago Rodriguez, Juanita Puentes, Pablo Arbelaez, Omid Mohareri,
  and Danail Stoyanov.
\newblock Sar-rarp50: Segmentation of surgical instrumentation and action
  recognition on robot-assisted radical prostatectomy challenge.
\newblock \emph{ArXiv}, abs/2401.00496, 2023.
\newblock \url{https://doi.org/10.48550/arXiv.2401.00496}.

\bibitem[Twinanda et~al.(2017)Twinanda, Shehata, Mutter, Marescaux,
  de~Mathelin, and Padoy]{twinanda2016endonet}
Andru~Putra Twinanda, Sherif Shehata, Didier Mutter, Jacques Marescaux, Michel
  de~Mathelin, and Nicolas Padoy.
\newblock Endonet: {A} deep architecture for recognition tasks on laparoscopic
  videos.
\newblock \emph{{IEEE} Transactions on Medical Imaging}, 36\penalty0
  (1):\penalty0 86--97, 2017.
\newblock \url{https://doi.org/10.1109/TMI.2016.2593957}.

\bibitem[Banik et~al.(2021)Banik, Roy, and Bhattacharjee]{banik2021net}
Debapriya Banik, Kaushiki Roy, and Debotosh Bhattacharjee.
\newblock Em-net: An efficient m-net for segmentation of surgical instruments
  in colonoscopy frames.
\newblock \emph{Nordic Machine Intelligence}, 1\penalty0 (1):\penalty0 14--16,
  2021.
\newblock \url{https://doi.org/10.5617%2Fnmi.9122}.

\bibitem[Chou(2021)]{chou2021automatic}
YuCheng Chou.
\newblock Automatic polyp and instrument segmentation in medai-2021.
\newblock \emph{Nordic Machine Intelligence}, 1\penalty0 (1):\penalty0 17--19,
  2021.
\newblock \url{https://doi.org/10.5617%2Fnmi.9125}.

\bibitem[da~Costa~Rocha et~al.(2019)da~Costa~Rocha, Padoy, and
  Rosa]{da2019self}
Cristian da~Costa~Rocha, Nicolas Padoy, and Benoit Rosa.
\newblock Self-supervised surgical tool segmentation using kinematic
  information.
\newblock In \emph{2019 International Conference on Robotics and Automation
  (ICRA)}, pages 8720--8726. IEEE, 2019.
\newblock \url{https://doi.org/10.1109%2Ficra.2019.8794334}.

\bibitem[Devi et~al.(2022)Devi, Roy, Thongam, and Chingangbam]{devi2022msdfn}
Wangkheirakpam~Reema Devi, Sudipta Roy, Khelchandra Thongam, and Chiranjiv
  Chingangbam.
\newblock Multi-scale dilated fusion network ({MSDFN}) for automatic instrument
  segmentation.
\newblock \emph{Journal of Computer Science and Technology Studies}, 4\penalty0
  (1):\penalty0 66--72, 2022.
\newblock \url{https://doi.org/10.32996/jcsts.2022.4.1.7}.

\bibitem[Galdran(2021)]{galdran2021polyp}
Adrian Galdran.
\newblock Polyp and surgical instrument segmentation with double
  encoder-decoder networks.
\newblock \emph{Nordic Machine Intelligence}, 1\penalty0 (1):\penalty0 5--7,
  2021.
\newblock \url{https://doi.org/10.5617/nmi.9107}.

\bibitem[Garcia-Peraza-Herrera et~al.(2017)Garcia-Peraza-Herrera, Li, Fidon,
  Gruijthuijsen, Devreker, Attilakos, Deprest, Vander~Poorten, Stoyanov,
  Vercauteren, et~al.]{garcia2017toolnet}
Luis~C Garcia-Peraza-Herrera, Wenqi Li, Lucas Fidon, Caspar Gruijthuijsen,
  Alain Devreker, George Attilakos, Jan Deprest, Emmanuel Vander~Poorten,
  Danail Stoyanov, Tom Vercauteren, et~al.
\newblock Toolnet: Holistically-nested real-time segmentation of robotic
  surgical tools.
\newblock In \emph{2017 IEEE/RSJ International Conference on Intelligent Robots
  and Systems (IROS)}, pages 5717--5722. IEEE, 2017.
\newblock \url{https://doi.org/10.1109%2Firos.2017.8206462}.

\bibitem[Garc{\'{\i}}a{-}Peraza{-}Herrera
  et~al.(2021{\natexlab{b}})Garc{\'{\i}}a{-}Peraza{-}Herrera, Fidon, D'Ettorre,
  Stoyanov, Vercauteren, and Ourselin]{garcia2021image}
Luis~C. Garc{\'{\i}}a{-}Peraza{-}Herrera, Lucas Fidon, Claudia D'Ettorre,
  Danail Stoyanov, Tom Vercauteren, and S{\'{e}}bastien Ourselin.
\newblock Image compositing for segmentation of surgical tools without manual
  annotations.
\newblock \emph{{IEEE} Transactions on Medical Imaging}, 40\penalty0
  (5):\penalty0 1450--1460, 2021{\natexlab{b}}.
\newblock \url{https://doi.org/10.1109/TMI.2021.3057884}.

\bibitem[Hasan et~al.(2021{\natexlab{b}})Hasan, Simon, and
  Linte]{hasan2021segmentation}
SM~Kamrul Hasan, Richard~A Simon, and Cristian~A Linte.
\newblock Segmentation and removal of surgical instruments for background
  scenevisualization from endoscopic / laparoscopic video.
\newblock In \emph{Medical Imaging 2021: Image-Guided Procedures, Robotic
  Interventions, and Modeling}, volume 11598, pages 55--61. SPIE,
  2021{\natexlab{b}}.
\newblock \url{https://doi.org/10.1117%2F12.2580668}.

\bibitem[P{\'{e}}rez et~al.(2020)P{\'{e}}rez, Marinho, Harada, and
  Mitsuishi]{heredia2020effects}
Sa{\'{u}}l Alexis~Heredia P{\'{e}}rez, Murilo~Marques Marinho, Kanako Harada,
  and Mamoru Mitsuishi.
\newblock The effects of different levels of realism on the training of cnns
  with only synthetic images for the semantic segmentation of robotic
  instruments in a head phantom.
\newblock \emph{International Journal of Computer Assisted Radiology and
  Surgery}, 15\penalty0 (8):\penalty0 1257--1265, 2020.
\newblock \url{https://doi.org/10.1007/s11548-020-02185-0}.

\bibitem[Huang et~al.(2022{\natexlab{b}})Huang, Chitrakar, Jiang, Yung, and
  Su]{huang2022surgical}
Kevin Huang, Digesh Chitrakar, Wenfan Jiang, Isabella Yung, and Yun-Hsuan Su.
\newblock Surgical tool segmentation with pose-informed morphological polar
  transform of endoscopic images.
\newblock \emph{Journal of Medical Robotics Research}, 7\penalty0
  (02n03):\penalty0 2241003, 2022{\natexlab{b}}.
\newblock \url{https://doi.org/10.1142%2Fs2424905x22410033}.

\bibitem[Kalavakonda et~al.(2019)Kalavakonda, Hannaford, Qazi, and
  Sekhar]{kalavakonda2019autonomous}
Niveditha Kalavakonda, Blake Hannaford, Zeeshan Qazi, and Laligam Sekhar.
\newblock Autonomous neurosurgical instrument segmentation using end-to-end
  learning.
\newblock In \emph{2019 IEEE/CVF Conference on Computer Vision and Pattern
  Recognition Workshops (CVPRW)}, pages 514--516. IEEE, 2019.
\newblock \url{https://doi.org/10.1109%2Fcvprw.2019.00076}.

\bibitem[Kalia et~al.(2021)Kalia, Aleef, Navab, Black, and
  Salcudean]{kalia2021co}
Megha Kalia, Tajwar~Abrar Aleef, Nassir Navab, Peter~C. Black, and Septimiu~E.
  Salcudean.
\newblock Co-generation and segmentation for generalized surgical instrument
  segmentation on unlabelled data.
\newblock In \emph{Medical Image Computing and Computer Assisted Intervention
  {\textendash} {MICCAI} 2021}, volume 12904 of \emph{Lecture Notes in Computer
  Science}, pages 403--412. Springer International Publishing, 2021.
\newblock \url{https://doi.org/10.1007/978-3-030-87202-1_39}.

\bibitem[Keprate and Pandey(2021)]{keprate2021kvasir}
Arvind Keprate and Sumit Pandey.
\newblock Kvasir-instruments and polyp segmentation using unet.
\newblock \emph{Nordic Machine Intelligence}, 1\penalty0 (1):\penalty0 26--28,
  2021.
\newblock \url{https://doi.org/10.5617/nmi.9130}.

\bibitem[Lee et~al.(2019{\natexlab{a}})Lee, Plishker, Liu, Bhattacharyya, and
  Shekhar]{lee2019weakly}
Eung-Joo Lee, William Plishker, Xinyang Liu, Shuvra~S Bhattacharyya, and Raj
  Shekhar.
\newblock Weakly supervised segmentation for real-time surgical tool tracking.
\newblock \emph{Healthcare Technology Letters}, 6\penalty0 (6):\penalty0
  231--236, 2019{\natexlab{a}}.
\newblock \url{https://doi.org/10.1049%2Fhtl.2019.0083}.

\bibitem[Lee et~al.(2019{\natexlab{b}})Lee, Plishker, Liu, Kane, Bhattacharyya,
  and Shekhar]{lee2019segmentation}
Eung-Joo Lee, William Plishker, Xinyang Liu, Timothy Kane, Shuvra~S
  Bhattacharyya, and Raj Shekhar.
\newblock Segmentation of surgical instruments in laparoscopic videos: training
  dataset generation and deep-learning-based framework.
\newblock In \emph{Medical Imaging 2019: Image-Guided Procedures, Robotic
  Interventions, and Modeling}, volume 10951, pages 461--469. SPIE,
  2019{\natexlab{b}}.
\newblock \url{https://doi.org/10.1117%2F12.2512994}.

\bibitem[Leifman et~al.(2022)Leifman, Aides, Golany, Freedman, and
  Rivlin]{leifman2022pixel}
George Leifman, Amit Aides, Tomer Golany, Daniel Freedman, and Ehud Rivlin.
\newblock Pixel-accurate segmentation of surgical tools based on bounding box
  annotations.
\newblock In \emph{2022 26th International Conference on Pattern Recognition
  (ICPR)}, pages 5096--5103. {IEEE}, 2022.
\newblock \url{https://doi.org/10.1109/ICPR56361.2022.9956530}.

\bibitem[Lou et~al.(2023)Lou, Tawfik, Yao, Liu, and Noble]{lou2023min}
Ange Lou, Kareem Tawfik, Xing Yao, Ziteng Liu, and Jack Noble.
\newblock Min-max similarity: A contrastive semi-supervised deep learning
  network for surgical tools segmentation.
\newblock \emph{IEEE Transactions on Medical Imaging}, 2023.
\newblock \url{https://doi.org/10.1109%2Ftmi.2023.3266137}.

\bibitem[Ni et~al.(2019)Ni, Bian, Xie, Hou, Zhou, and Zhou]{ni2019rasnet}
Zhen-Liang Ni, Gui-Bin Bian, Xiao-Liang Xie, Zeng-Guang Hou, Xiao-Hu Zhou, and
  Yan-Jie Zhou.
\newblock Rasnet: Segmentation for tracking surgical instruments in surgical
  videos using refined attention segmentation network.
\newblock In \emph{2019 41st Annual International Conference of the IEEE
  Engineering in Medicine and Biology Society (EMBC)}, pages 5735--5738. IEEE,
  2019.
\newblock \url{https://doi.org/10.1109%2Fembc.2019.8856495}.

\bibitem[Pakhomov et~al.(2020)Pakhomov, Shen, and Navab]{pakhomov2020towards}
Daniil Pakhomov, Wei Shen, and Nassir Navab.
\newblock Towards unsupervised learning for instrument segmentation in robotic
  surgery with cycle-consistent adversarial networks.
\newblock In \emph{2020 IEEE/RSJ International Conference on Intelligent Robots
  and Systems (IROS)}, pages 8499--8504. IEEE, 2020.
\newblock \url{https://doi.org/10.1109%2Firos45743.2020.9340816}.

\bibitem[Papp et~al.(2022)Papp, Elek, and Haidegger]{papp2022surgical}
Dóra Papp, Renáta~Nagyné Elek, and Tamás Haidegger.
\newblock Surgical tool segmentation on the jigsaws dataset for autonomous
  image-based skill assessment.
\newblock In \emph{2022 IEEE 10th Jubilee International Conference on
  Computational Cybernetics and Cyber-Medical Systems (ICCC)}, pages
  000049--000056. IEEE, 2022.
\newblock \url{https://doi.org/10.1109%2Ficcc202255925.2022.9922713}.

\bibitem[Psychogyios et~al.(2022)Psychogyios, Mazomenos, Vasconcelos, and
  Stoyanov]{psychogyios2022msdesis}
Dimitrios Psychogyios, Evangelos Mazomenos, Francisco Vasconcelos, and Danail
  Stoyanov.
\newblock Msdesis: Multi-task stereo disparity estimation and surgical
  instrument segmentation.
\newblock \emph{IEEE Transactions on Medical Imaging}, 41\penalty0
  (11):\penalty0 3218--3230, 2022.
\newblock \url{https://doi.org/10.1109%2Ftmi.2022.3181229}.

\bibitem[Qin et~al.(2019)Qin, Li, Su, Xu, and Hannaford]{qin2019surgical}
Fangbo Qin, Yangming Li, Yun-Hsuan Su, De~Xu, and Blake Hannaford.
\newblock Surgical instrument segmentation for endoscopic vision with data
  fusion of cnn prediction and kinematic pose.
\newblock In \emph{2019 International Conference on Robotics and Automation
  (ICRA)}, pages 9821--9827. IEEE, 2019.
\newblock \url{https://doi.org/10.1109%2Ficra.2019.8794122}.

\bibitem[Rajak and Mirza(2021)]{rajak2021segmentation}
Rishav~Kumar Rajak and Ashar Mirza.
\newblock Segmentation of polyp instruments using unet based deep learning
  model.
\newblock \emph{Nordic Machine Intelligence}, 1\penalty0 (1):\penalty0 44--46,
  2021.
\newblock \url{https://doi.org/10.5617%2Fnmi.9145}.

\bibitem[Sahu et~al.(2020)Sahu, Str{\"{o}}msd{\"{o}}rfer, Mukhopadhyay, and
  Zachow]{sahu2020endo}
Manish Sahu, Ronja Str{\"{o}}msd{\"{o}}rfer, Anirban Mukhopadhyay, and Stefan
  Zachow.
\newblock Endo-sim2real: Consistency learning-based domain adaptation for
  instrument segmentation.
\newblock In \emph{Medical Image Computing and Computer Assisted Intervention
  {\textendash} {MICCAI} 2020}, volume 12263 of \emph{Lecture Notes in Computer
  Science}, pages 784--794. Springer International Publishing, 2020.
\newblock \url{https://doi.org/10.1007/978-3-030-59716-0_75}.

\bibitem[Sahu et~al.(2021)Sahu, Mukhopadhyay, and Zachow]{sahu2021simulation}
Manish Sahu, Anirban Mukhopadhyay, and Stefan Zachow.
\newblock Simulation-to-real domain adaptation with teacher-student learning
  for endoscopic instrument segmentation.
\newblock \emph{International Journal of Computer Assisted Radiology and
  Surgery}, 16\penalty0 (5):\penalty0 849--859, 2021.
\newblock \url{https://doi.org/10.1007/s11548-021-02383-4}.

\bibitem[Su et~al.(2018{\natexlab{a}})Su, Huang, Huang, and
  Hannaford]{su2018comparison}
Yun-Hsuan Su, Issac Huang, Kevin Huang, and Blake Hannaford.
\newblock Comparison of 3d surgical tool segmentation procedures with robot
  kinematics prior.
\newblock In \emph{2018 IEEE/RSJ International Conference on Intelligent Robots
  and Systems (IROS)}, pages 4411--4418. IEEE, 2018{\natexlab{a}}.
\newblock \url{https://doi.org/10.1109%2Firos.2018.8594428}.

\bibitem[Su et~al.(2018{\natexlab{b}})Su, Huang, and Hannaford]{su2018real}
Yun-Hsuan Su, Kevin Huang, and Blake Hannaford.
\newblock Real-time vision-based surgical tool segmentation with robot
  kinematics prior.
\newblock In \emph{2018 International Symposium on Medical Robotics (ISMR)}.
  IEEE, 2018{\natexlab{b}}.
\newblock \url{https://doi.org/10.1109%2Fismr.2018.8333305}.

\bibitem[Suzuki et~al.(2019)Suzuki, Doman, and Mekada]{suzuki2019depth}
Takuya Suzuki, Keisuke Doman, and Yoshito Mekada.
\newblock Depth estimation for instrument segmentation from a single
  laparoscopic video toward laparoscopic surgery.
\newblock In \emph{Proceedings of the 2019 International Conference on
  Intelligent Medicine and Image Processing (IMIP)}, pages 21--24. ACM, 2019.
\newblock \url{https://doi.org/10.1145%2F3332340.3332347}.

\bibitem[Wang et~al.(2021{\natexlab{a}})Wang, Qiu, Hu, Chen, Ye, and
  Liu]{wang2021surgical}
Yiming Wang, Zhongxi Qiu, Yan Hu, Hao Chen, Fangfu Ye, and Jiang Liu.
\newblock Surgical instrument segmentation based on multi-scale and multi-level
  feature network.
\newblock In \emph{2021 43rd Annual International Conference of the {IEEE}
  Engineering in Medicine \& Biology Society (EMBC)}, pages 2672--2675. {IEEE},
  2021{\natexlab{a}}.
\newblock \url{https://doi.org/10.1109/EMBC46164.2021.9629891}.

\bibitem[Wang et~al.(2022{\natexlab{c}})Wang, Islam, Xu, and
  Ren]{wang2022rethinking}
An~Wang, Mobarakol Islam, Mengya Xu, and Hongliang Ren.
\newblock Rethinking surgical instrument segmentation: {A} background image can
  be all you need.
\newblock In \emph{Medical Image Computing and Computer Assisted Intervention
  {\textendash} {MICCAI} 2022}, volume 13437 of \emph{Lecture Notes in Computer
  Science}, pages 355--364. Springer International Publishing,
  2022{\natexlab{c}}.
\newblock \url{https://doi.org/10.1007/978-3-031-16449-1_34}.

\bibitem[Yang et~al.(2022{\natexlab{a}})Yang, Gu, Bian, and
  Liu]{yang2022attention}
Lei Yang, Yuge Gu, Gui{-}Bin Bian, and Yanhong Liu.
\newblock An attention-guided network for surgical instrument segmentation from
  endoscopic images.
\newblock \emph{Computers in Biology and Medicine}, 151\penalty0
  (Part):\penalty0 106216, 2022{\natexlab{a}}.
\newblock \url{https://doi.org/10.1016/j.compbiomed.2022.106216}.

\bibitem[Yang et~al.(2022{\natexlab{b}})Yang, Gu, Bian, and Liu]{yang2022tmf}
Lei Yang, Yuge Gu, Guibin Bian, and Yanhong Liu.
\newblock Tmf-net: A transformer-based multiscale fusion network for surgical
  instrument segmentation from endoscopic images.
\newblock \emph{IEEE Transactions on Instrumentation and Measurement},
  72:\penalty0 1--15, 2022{\natexlab{b}}.
\newblock \url{https://doi.org/10.1109%2Ftim.2022.3225922}.

\bibitem[Yang et~al.(2022{\natexlab{c}})Yang, Simon, and Linte]{yang2022weakly}
Zixin Yang, Richard Simon, and Cristian Linte.
\newblock A weakly supervised learning approach for surgical instrument
  segmentation from laparoscopic video sequences.
\newblock In \emph{Medical Imaging 2022: Image-Guided Procedures, Robotic
  Interventions, and Modeling}, volume 12034, pages 412--417. SPIE,
  2022{\natexlab{c}}.
\newblock \url{https://doi.org/10.1117%2F12.2610778}.

\bibitem[Yang et~al.(2023{\natexlab{a}})Yang, Gu, Bian, and Liu]{yang2023maf}
Lei Yang, Yuge Gu, Guibin Bian, and Yanhong Liu.
\newblock Maf-net: A multi-scale attention fusion network for automatic
  surgical instrument segmentation.
\newblock \emph{Biomedical Signal Processing and Control}, 85:\penalty0 104912,
  2023{\natexlab{a}}.
\newblock \url{https://doi.org/10.1016%2Fj.bspc.2023.104912}.

\bibitem[Yang et~al.(2023{\natexlab{b}})Yang, Wang, Gu, Bian, Liu, and
  Yu]{yang2023tma}
Lei Yang, Hongyong Wang, Yuge Gu, Guibin Bian, Yanhong Liu, and Hongnian Yu.
\newblock Tma-net: A transformer-based multi-scale attention network for
  surgical instrument segmentation.
\newblock \emph{IEEE Transactions on Medical Robotics and Bionics}, 5\penalty0
  (2):\penalty0 323--334, 2023{\natexlab{b}}.
\newblock \url{https://doi.org/10.1109%2Ftmrb.2023.3269856}.

\bibitem[Yeung(2021)]{yeung2021attention}
Michael Yeung.
\newblock Attention u-net ensemble for interpretable polyp and instrument
  segmentation.
\newblock \emph{Nordic Machine Intelligence}, 1\penalty0 (1):\penalty0 47--49,
  2021.
\newblock \url{https://doi.org/10.5617%2Fnmi.9157}.

\bibitem[Yu et~al.(2020)Yu, Wang, Yu, Yan, and Xia]{yu2020holistically}
Lingtao Yu, Pengcheng Wang, Xiaoyan Yu, Yusheng Yan, and Yongqiang Xia.
\newblock A holistically-nested u-net: Surgical instrument segmentation basedon
  convolutional neural network.
\newblock \emph{Journal of Digital Imaging}, 33\penalty0 (2):\penalty0
  341--347, 2020.
\newblock \url{https://doi.org/10.1007%2Fs10278-019-00277-1}.

\bibitem[Zhang et~al.(2021)Zhang, Rosa, and Nageotte]{zhang2021surgical}
Zhongkai Zhang, Beno{\^{\i}}t Rosa, and Florent Nageotte.
\newblock Surgical tool segmentation using generative adversarial networks with
  unpaired training data.
\newblock \emph{{IEEE} Robotics and Automation Letters}, 6\penalty0
  (4):\penalty0 6266--6273, 2021.
\newblock \url{https://doi.org/10.1109/LRA.2021.3092302}.

\bibitem[Dong et~al.(2021)Dong, Wang, Luo, Lai, Wang, and
  Wang]{dong2021semantic}
Yu~Dong, Hongbo Wang, Jingjing Luo, Zhiping Lai, Fuhao Wang, and Jiawei Wang.
\newblock Semantic segmentation of surgical instruments based on enhanced
  multi-scale receptive field.
\newblock In \emph{Journal of Physics: Conference Series}, volume 2003, page
  012006. IOP Publishing, 2021.
\newblock \url{https://doi.org/10.1088/1742-6596/2003/1/012006}.

\bibitem[Guo et~al.(2022)Guo, Ye, Zhang, and He]{guo2022conditional}
Yue Guo, Changjian Ye, Yongchang Zhang, and Wenhao He.
\newblock Conditional relativistic gan for fast part segmentation of surgical
  instruments.
\newblock In \emph{CAIBDA 2022; 2nd International Conference on Artificial
  Intelligence, Big Data and Algorithms}, pages 1--6. VDE, 2022.

\bibitem[He et~al.(2020)He, Song, Guo, Bian, Sun, Zhou, and
  Wang]{he2020multiscale}
Wenhao He, Haitao Song, Yue Guo, Guibin Bian, Yuejie Sun, Xiaowei Zhou, and
  Xiaonan Wang.
\newblock Multiscale matters for part segmentation of instruments in robotic
  surgery.
\newblock \emph{{IET} Image Processing}, 14\penalty0 (13):\penalty0 3215--3222,
  2020.
\newblock \url{https://doi.org/10.1049/iet-ipr.2020.0320}.

\bibitem[Kamrul~Hasan and Linte(2019)]{kamrul2019unetplus}
S.~M. Kamrul~Hasan and Cristian~A. Linte.
\newblock U-netplus: A modified encoder-decoder u-net architecture for semantic
  and instance segmentation of surgical instruments from laparoscopic images.
\newblock In \emph{2019 41st Annual International Conference of the IEEE
  Engineering in Medicine and Biology Society (EMBC)}, pages 7205--7211. IEEE,
  2019.
\newblock \url{https://doi.org/10.1109/embc.2019.8856791}.

\bibitem[Laina et~al.(2017)Laina, Rieke, Rupprecht, Vizca{\'\i}no, Eslami,
  Tombari, and Navab]{laina2017concurrent}
Iro Laina, Nicola Rieke, Christian Rupprecht, Josu{\'e}~Page Vizca{\'\i}no,
  Abouzar Eslami, Federico Tombari, and Nassir Navab.
\newblock Concurrent segmentation and localization for tracking of surgical
  instruments.
\newblock In \emph{Medical Image Computing and Computer-Assisted Intervention
  {\textendash} MICCAI 2017}, pages 664--672. Springer International
  Publishing, 2017.
\newblock \url{https://doi.org/10.1007/978-3-319-66185-8_75}.

\bibitem[Shen et~al.(2023)Shen, Wang, Liu, Wang, Ding, Zhang, and
  Meijering]{shen2023branch}
Wenting Shen, Yaonan Wang, Min Liu, Jiazheng Wang, Renjie Ding, Zhe Zhang, and
  Erik Meijering.
\newblock Branch aggregation attention network for robotic surgical instrument
  segmentation.
\newblock \emph{IEEE Transactions on Medical Imaging}, 2023.
\newblock \url{https://doi.org/10.1109%2Ftmi.2023.3288127}.

\bibitem[Vishal and Kumar(2018)]{vishal2018robotic}
V~Vishal and C~Udaya Kumar.
\newblock Robotic surgical instrument segmentation using dual global attention
  upsample.
\newblock In \emph{2018 32nd Conference on Neural Information Processing
  Systems (NIPS)}, 2018.

\bibitem[Wang et~al.(2021{\natexlab{b}})Wang, Wang, Zhong, Bai, Huang, Zhao,
  and Xia]{wang2021pai}
Xiaoyan Wang, Luyao Wang, Xingyu Zhong, Cong Bai, Xiaojie Huang, Ruiyi Zhao,
  and Ming Xia.
\newblock Pai-net: {A} modified u-net of reducing semantic gap for surgical
  instrument segmentation.
\newblock \emph{{IET} Image Processing}, 15\penalty0 (12):\penalty0 2959--2969,
  2021{\natexlab{b}}.
\newblock \url{https://doi.org/10.1049/ipr2.12283}.

\bibitem[Wang et~al.(2023)Wang, Hu, Shen, Zhang, Li, Qiu, Ye, and
  Liu]{wang2023cgba}
Yiming Wang, Yan Hu, Junyong Shen, Xiaoqing Zhang, Heng Li, Zhongxi Qiu, Fangfu
  Ye, and Jiang Liu.
\newblock Cgba-net: context-guided bidirectional attention network for surgical
  instrument segmentation.
\newblock \emph{International Journal of Computer Assisted Radiology and
  Surgery}, 2023.
\newblock \url{https://doi.org/10.1007%2Fs11548-023-02906-1}.

\bibitem[Zhou et~al.(2021)Zhou, Guo, He, and Song]{zhou2021hierarchical}
Xiaowei Zhou, Yue Guo, Wenhao He, and Haitao Song.
\newblock Hierarchical attentional feature fusion for surgical instrument
  segmentation.
\newblock In \emph{2021 43rd Annual International Conference of the {IEEE}
  Engineering in Medicine \& Biology Society (EMBC)}, pages 3061--3065. {IEEE},
  2021.
\newblock \url{https://doi.org/10.1109/EMBC46164.2021.9630553}.

\bibitem[Islam et~al.(2020)Islam, VS, and Ren]{islam2020ap}
Mobarakol Islam, Vibashan VS, and Hongliang Ren.
\newblock {AP-MTL:} attention pruned multi-task learning model for real-time
  instrument detection and segmentation in robot-assisted surgery.
\newblock In \emph{2020 {IEEE} International Conference on Robotics and
  Automation ({ICRA})}, pages 8433--8439. {IEEE}, 2020.
\newblock \url{https://doi.org/10.1109/ICRA40945.2020.9196905}.

\bibitem[Han et~al.(2021)Han, Yue, Duan, Bai, Du, Zhou, and Wang]{han2021ceid}
Wanwan Han, Guanghui Yue, Lvyin Duan, Xue Bai, Jingfeng Du, Tianwei Zhou, and
  Tianfu Wang.
\newblock Ceid: Benchmark dataset for designing segmentation algorithms of
  instruments used in colorectal endoscopy.
\newblock In \emph{International Conference on Image and Graphics}, pages
  618--629. Springer, 2021.
\newblock \url{https://doi.org/10.1007%2F978-3-030-87358-5_50}.

\bibitem[Sanchez{-}Matilla et~al.(2021)Sanchez{-}Matilla, Robu, Luengo, and
  Stoyanov]{sanchez2021scalable}
Ricardo Sanchez{-}Matilla, Maria Robu, Imanol Luengo, and Danail Stoyanov.
\newblock Scalable joint detection and segmentation of surgical instruments
  with weak supervision.
\newblock In \emph{Medical Image Computing and Computer Assisted Intervention
  {\textendash} {MICCAI} 2021}, volume 12902 of \emph{Lecture Notes in Computer
  Science}, pages 501--511. Springer International Publishing, 2021.
\newblock \url{https://doi.org/10.1007/978-3-030-87196-3_47}.

\bibitem[Liu et~al.(2021)Liu, Guo, and Yuan]{liu2021prototypical}
Jie Liu, Xiaoqing Guo, and Yixuan Yuan.
\newblock Prototypical interaction graph for unsupervised domain adaptation in
  surgical instrument segmentation.
\newblock In \emph{Medical Image Computing and Computer Assisted Intervention
  {\textendash} {MICCAI} 2021}, volume 12903 of \emph{Lecture Notes in Computer
  Science}, pages 272--281. Springer International Publishing, 2021.
\newblock \url{https://doi.org/10.1007/978-3-030-87199-4_26}.

\bibitem[Ni et~al.(2020{\natexlab{a}})Ni, Bian, Hou, Zhou, Xie, and
  Li]{ni2020attention}
Zhen{-}Liang Ni, Gui{-}Bin Bian, Zeng{-}Guang Hou, Xiao{-}Hu Zhou, Xiao{-}Liang
  Xie, and Zhen Li.
\newblock Attention-guided lightweight network for real-time segmentation of
  robotic surgical instruments.
\newblock In \emph{2020 {IEEE} International Conference on Robotics and
  Automation {ICRA}}, pages 9939--9945. {IEEE}, 2020{\natexlab{a}}.
\newblock \url{https://doi.org/10.1109/ICRA40945.2020.9197425}.

\bibitem[Ni et~al.(2020{\natexlab{b}})Ni, Bian, Wang, Zhou, Hou, Chen, and
  Xie]{ni2020pyramid}
Zhen{-}Liang Ni, Gui{-}Bin Bian, Guan'an Wang, Xiao{-}Hu Zhou, Zeng{-}Guang
  Hou, Hua{-}Bin Chen, and Xiao{-}Liang Xie.
\newblock Pyramid attention aggregation network for semantic segmentation of
  surgical instruments.
\newblock \emph{Proceedings of the {AAAI} Conference on Artificial
  Intelligence}, pages 11782--11790, 2020{\natexlab{b}}.
\newblock \url{https://ojs.aaai.org/index.php/AAAI/article/view/6850}.

\bibitem[Ni et~al.(2020{\natexlab{c}})Ni, Bian, Wang, Zhou, Hou, Xie, Li, and
  Wang]{ni2021barnet}
Zhen{-}Liang Ni, Gui{-}Bin Bian, Guan'an Wang, Xiao{-}Hu Zhou, Zeng{-}Guang
  Hou, Xiao{-}Liang Xie, Zhen Li, and Yu{-}Han Wang.
\newblock Barnet: Bilinear attention network with adaptive receptive fields for
  surgical instrument segmentation.
\newblock In \emph{Proceedings of the Twenty-Ninth International Joint
  Conference on Artificial Intelligence (IJCAI)}, pages 832--838. International
  Joint Conferences on Artificial Intelligence Organization,
  2020{\natexlab{c}}.
\newblock \url{https://doi.org/10.24963/ijcai.2020/116}.

\bibitem[Ni et~al.(2022{\natexlab{a}})Ni, Zhou, Wang, Yue, Li, Bian, and
  Hou]{ni2022surginet}
Zhen{-}Liang Ni, Xiao{-}Hu Zhou, Guan'an Wang, Wen{-}Qian Yue, Zhen Li,
  Gui{-}Bin Bian, and Zeng{-}Guang Hou.
\newblock Surginet: Pyramid attention aggregation and class-wise
  self-distillation for surgical instrument segmentation.
\newblock \emph{Medical Image Analysis}, 76:\penalty0 102310,
  2022{\natexlab{a}}.
\newblock \url{https://doi.org/10.1016/j.media.2021.102310}.

\bibitem[Xue and Gu(2021)]{xue2021surgical}
Mengchen Xue and Lixu Gu.
\newblock Surgical instrument segmentation method based on improved mobilenetv2
  network.
\newblock In \emph{2021 6th International Symposium on Computer and Information
  Processing Technology (ISCIPT)}, pages 744--747. IEEE, 2021.
\newblock \url{https://doi.org/10.1109%2Fiscipt53667.2021.00157}.

\bibitem[Andersen et~al.(2021)Andersen, Schwaner, and
  Savarimuthu]{andersen2021real}
Jakob K.~H. Andersen, Kim~Lindberg Schwaner, and Thiusius~Rajeeth Savarimuthu.
\newblock Real-time segmentation of surgical tools and needle using a
  mobile-u-net.
\newblock In \emph{2021 20th International Conference on Advanced Robotics
  (ICAR)}, pages 148--154. {IEEE}, 2021.
\newblock \url{https://doi.org/10.1109/ICAR53236.2021.9659326}.

\bibitem[Nema and Vachhani(2023)]{nema2023unpaired}
Shubhangi Nema and Leena Vachhani.
\newblock Unpaired deep adversarial learning for multi-class segmentation of
  instruments in robot-assisted surgical videos.
\newblock \emph{The International Journal of Medical Robotics and Computer
  Assisted Surgery}, 19\penalty0 (4):\penalty0 e2514, 2023.
\newblock \url{https://doi.org/10.1002%2Frcs.2514}.

\bibitem[Ozawa et~al.(2021)Ozawa, Hayashi, Oda, Oda, Kitasaka, Takeshita, Ito,
  and Mori]{ozawa2021synthetic}
Takuya Ozawa, Yuichiro Hayashi, Hirohisa Oda, Masahiro Oda, Takayuki Kitasaka,
  Nobuyoshi Takeshita, Masaaki Ito, and Kensaku Mori.
\newblock Synthetic laparoscopic video generation for machine learning-based
  surgical instrument segmentation from real laparoscopic video and virtual
  surgical instruments.
\newblock \emph{Computer Methods in Biomechanics and Biomedical Engineering:
  Imaging \& Visualization}, 9\penalty0 (3):\penalty0 225--232, 2021.
\newblock \url{https://doi.org/10.1080/21681163.2020.1835560}.

\bibitem[Sun et~al.(2021)Sun, Pan, and Fu]{sun2021lightweight}
Yanwen Sun, Bo~Pan, and Yili Fu.
\newblock Lightweight deep neural network for real-time instrument semantic
  segmentation in robot assisted minimally invasive surgery.
\newblock \emph{{IEEE} Robotics and Automation Letters}, 6\penalty0
  (2):\penalty0 3870--3877, 2021.
\newblock \url{https://doi.org/10.1109/LRA.2021.3066956}.

\bibitem[Colleoni et~al.(2022)Colleoni, Psychogyios, van Amsterdam,
  Vasconcelos, and Stoyanov]{colleoni2022ssis}
Emanuele Colleoni, Dimitris Psychogyios, Beatrice van Amsterdam, Francisco
  Vasconcelos, and Danail Stoyanov.
\newblock Ssis-seg: Simulation-supervised image synthesis for surgical
  instrument segmentation.
\newblock \emph{{IEEE} Transactions on Medical Imaging}, 41\penalty0
  (11):\penalty0 3074--3086, 2022.
\newblock \url{https://doi.org/10.1109/TMI.2022.3178549}.

\bibitem[Mahmood et~al.(2022)Mahmood, Cho, and Park]{mahmood2022dsrd}
Tahir Mahmood, Se~Woon Cho, and Kang~Ryoung Park.
\newblock Dsrd-net: Dual-stream residual dense network for semantic
  segmentation of instruments in robot-assisted surgery.
\newblock \emph{Expert Systems with Applications}, 202:\penalty0 117420, 2022.
\newblock \url{https://doi.org/10.1016/j.eswa.2022.117420}.

\bibitem[Pakhomov et~al.(2019)Pakhomov, Premachandran, Allan, Azizian, and
  Navab]{pakhomov2019deep}
Daniil Pakhomov, Vittal Premachandran, Max Allan, Mahdi Azizian, and Nassir
  Navab.
\newblock Deep residual learning for instrument segmentation in robotic
  surgery.
\newblock In \emph{2019 10th International Machine Learning in Medical Imaging
  (MLMI) Workshop}, volume 11861 of \emph{Lecture Notes in Computer Science},
  pages 566--573. Springer International Publishing, 2019.
\newblock \url{https://doi.org/10.1007/978-3-030-32692-0_65}.

\bibitem[Liu et~al.(2022)Liu, Guo, and Yuan]{liu2021graph}
Jie Liu, Xiaoqing Guo, and Yixuan Yuan.
\newblock Graph-based surgical instrument adaptive segmentation via
  domain-common knowledge.
\newblock \emph{{IEEE} Transactions on Medical Imaging}, 41\penalty0
  (3):\penalty0 715--726, 2022.
\newblock \url{https://doi.org/10.1109/TMI.2021.3121138}.

\bibitem[Shvets et~al.(2018)Shvets, Rakhlin, Kalinin, and
  Iglovikov]{shvets2018automatic}
Alexey~A Shvets, Alexander Rakhlin, Alexandr~A Kalinin, and Vladimir~I
  Iglovikov.
\newblock Automatic instrument segmentation in robot-assisted surgery using
  deep learning.
\newblock In \emph{2018 17th IEEE International Conference on Machine Mearning
  and Applications (ICMLA)}, pages 624--628. Cold Spring Harbor Laboratory,
  2018.
\newblock \url{https://doi.org/10.1101%2F275867}.

\bibitem[Birodkar et~al.(2021)Birodkar, Lu, Li, Rathod, and
  Huang]{birodkar2021surprising}
Vighnesh Birodkar, Zhichao Lu, Siyang Li, Vivek Rathod, and Jonathan Huang.
\newblock The surprising impact of mask-head architecture on novel class
  segmentation.
\newblock In \emph{Proceedings of the IEEE/CVF International Conference on
  Computer Vision}, pages 7015--7025. IEEE, 2021.
\newblock \url{https://doi.org/10.1109%2Ficcv48922.2021.00693}.

\bibitem[Zhu et~al.(2017)Zhu, Park, Isola, and Efros]{zhu2017unpaired}
Jun{-}Yan Zhu, Taesung Park, Phillip Isola, and Alexei~A. Efros.
\newblock Unpaired image-to-image translation using cycle-consistent
  adversarial networks.
\newblock In \emph{2017 {IEEE} International Conference on Computer Vision
  ({ICCV})}, pages 2242--2251. IEEE, 2017.
\newblock \url{https://doi.org/10.1109/ICCV.2017.244}.

\bibitem[Colleoni and Stoyanov(2021)]{colleoni2021robotic}
Emanuele Colleoni and Danail Stoyanov.
\newblock Robotic instrument segmentation with image-to-image translation.
\newblock \emph{{IEEE} Robotics and Automation Letters}, 6\penalty0
  (2):\penalty0 935--942, 2021.
\newblock \url{https://doi.org/10.1109/LRA.2021.3056354}.

\bibitem[Agustinos and Voros(2015)]{agustinos20152d}
Anthony Agustinos and Sandrine Voros.
\newblock 2d/3d real-time tracking of surgical instruments based on endoscopic
  image processing.
\newblock In \emph{Computer-Assisted and Robotic Endoscopy}, pages 90--100.
  Springer International Publishing, 2015.
\newblock \url{https://doi.org/10.1007%2F978-3-319-29965-5_9}.

\bibitem[Amini~Khoiy et~al.(2016)Amini~Khoiy, Mirbagheri, and
  Farahmand]{amini2016automatic}
Keyvan Amini~Khoiy, Alireza Mirbagheri, and Farzam Farahmand.
\newblock Automatic tracking of laparoscopic instruments for autonomous control
  of a cameraman robot.
\newblock \emph{Minimally Invasive Therapy \& Allied Technologies}, 25\penalty0
  (3):\penalty0 121--128, 2016.
\newblock \url{https://doi.org/10.3109%2F13645706.2016.1141101}.

\bibitem[Attia et~al.(2017)Attia, Hossny, Nahavandi, and
  Asadi]{attia2017surgical}
Mohammed~Hassan Attia, Mohammed Hossny, Saeid Nahavandi, and Hamed Asadi.
\newblock Surgical tool segmentation using a hybrid deep {CNN-RNN} auto
  encoder-decoder.
\newblock In \emph{2017 {IEEE} International Conference on Systems, Man, and
  Cybernetics ({SMC})}, pages 3373--3378. {IEEE}, 2017.
\newblock \url{https://doi.org/10.1109/SMC.2017.8123151}.

\bibitem[Du et~al.(2016)Du, Allan, Dore, Ourselin, Hawkes, Kelly, and
  Stoyanov]{du2016combined}
Xiaofei Du, Maximilian Allan, Alessio Dore, Sebastien Ourselin, David Hawkes,
  John~D Kelly, and Danail Stoyanov.
\newblock Combined 2d and 3d tracking of surgical instrumentsfor minimally
  invasive and robotic-assisted surgery.
\newblock \emph{International Journal of Computer Assisted Radiology and
  Surgery}, 11\penalty0 (6):\penalty0 1109--1119, 2016.
\newblock \url{https://doi.org/10.1007%2Fs11548-016-1393-4}.

\bibitem[Garc{\'{\i}}a{-}Peraza{-}Herrera
  et~al.(2016)Garc{\'{\i}}a{-}Peraza{-}Herrera, Li, Gruijthuijsen, Devreker,
  Attilakos, Deprest, Poorten, Stoyanov, Vercauteren, and
  Ourselin]{garcia2016real}
Luis~C. Garc{\'{\i}}a{-}Peraza{-}Herrera, Wenqi Li, Caspar Gruijthuijsen, Alain
  Devreker, George Attilakos, Jan Deprest, Emmanuel B.~Vander Poorten, Danail
  Stoyanov, Tom Vercauteren, and S{\'{e}}bastien Ourselin.
\newblock Real-time segmentation of non-rigid surgical tools based on deep
  learning and tracking.
\newblock In \emph{Computer-Assisted and Robotic Endoscopy}, volume 10170,
  pages 84--95. Springer International Publishing, 2016.
\newblock \url{https://doi.org/10.1007/978-3-319-54057-3_8}.

\bibitem[Lin et~al.(2019)Lin, Qin, Bly, Moe, and Hannaford]{lin2019automatic}
Shan Lin, Fangbo Qin, Randall~A. Bly, Kris~S. Moe, and Blake Hannaford.
\newblock Automatic sinus surgery skill assessment based on instrument
  segmentation and tracking in endoscopic video.
\newblock In \emph{2019 First International Multiscale Multimodal Medical
  Imaging (MMMI) Workshop}, volume 11977 of \emph{Lecture Notes in Computer
  Science}, pages 93--100. Springer International Publishing, 2019.
\newblock \url{https://doi.org/10.1007/978-3-030-37969-8_12}.

\bibitem[Lin et~al.(2021)Lin, Qin, Peng, Bly, Moe, and Hannaford]{lin2021multi}
Shan Lin, Fangbo Qin, Haonan Peng, Randall~A. Bly, Kris~S. Moe, and Blake
  Hannaford.
\newblock Multi-frame feature aggregation for real-time instrument segmentation
  in endoscopic video.
\newblock \emph{{IEEE} Robotics and Automation Letters}, 6\penalty0
  (4):\penalty0 6773--6780, 2021.
\newblock \url{https://doi.org/10.1109/lra.2021.3096156}.

\bibitem[Liu et~al.(2020)Liu, Wei, Jiang, Wang, Miao, Shan, and
  Li]{liu2020unsupervised}
Daochang Liu, Yuhui Wei, Tingting Jiang, Yizhou Wang, Rulin Miao, Fei Shan, and
  Ziyu Li.
\newblock Unsupervised surgical instrument segmentation via anchor generation
  and semantic diffusion.
\newblock In \emph{Medical Image Computing and Computer Assisted Intervention
  {\textendash} {MICCAI} 2020}, pages 657--667. Springer International
  Publishing, 2020.
\newblock \url{https://doi.org/10.1007%2F978-3-030-59716-0_63}.

\bibitem[Sestini et~al.(2023)Sestini, Rosa, De~Momi, Ferrigno, and
  Padoy]{sestini2023fun}
Luca Sestini, Benoit Rosa, Elena De~Momi, Giancarlo Ferrigno, and Nicolas
  Padoy.
\newblock Fun-sis: A fully unsupervised approach for surgical instrument
  segmentation.
\newblock \emph{Medical Image Analysis}, 85:\penalty0 102751, 2023.
\newblock \url{https://doi.org/10.1016%2Fj.media.2023.102751}.

\bibitem[Yang et~al.(2022{\natexlab{d}})Yang, Gu, Bian, and Liu]{yang2022drr}
Lei Yang, Yuge Gu, Guibin Bian, and Yanhong Liu.
\newblock Drr-net: A dense-connected residual recurrent convolutional network
  for surgical instrument segmentation from endoscopic images.
\newblock \emph{IEEE Transactions on Medical Robotics and Bionics}, 4\penalty0
  (3):\penalty0 696--707, 2022{\natexlab{d}}.
\newblock \url{https://doi.org/10.1109%2Ftmrb.2022.3193420}.

\bibitem[Zhao et~al.(2021{\natexlab{a}})Zhao, Jin, Chen, Lu, Ng, Liu, Dou, and
  Heng]{zhao2021anchor}
Zixu Zhao, Yueming Jin, Junming Chen, Bo~Lu, Chi{-}Fai Ng, Yun{-}Hui Liu,
  Qi~Dou, and Pheng{-}Ann Heng.
\newblock Anchor-guided online meta adaptation for fast one-shot instrument
  segmentation from robotic surgical videos.
\newblock \emph{Medical Image Analysis}, 74:\penalty0 102240,
  2021{\natexlab{a}}.
\newblock \url{https://doi.org/10.1016/j.media.2021.102240}.

\bibitem[Zhao et~al.(2021{\natexlab{b}})Zhao, Jin, Lu, Ng, Dou, Liu, and
  Heng]{zhao2021one}
Zixu Zhao, Yueming Jin, Bo~Lu, Chi{-}Fai Ng, Qi~Dou, Yun{-}Hui Liu, and
  Pheng{-}Ann Heng.
\newblock One to many: Adaptive instrument segmentation via meta learning and
  dynamic online adaptation in robotic surgical video.
\newblock In \emph{2021 {IEEE} International Conference on Robotics and
  Automation {ICRA}}, pages 13553--13559. {IEEE}, 2021{\natexlab{b}}.
\newblock \url{https://doi.org/10.1109/ICRA48506.2021.9561690}.

\bibitem[Li et~al.(2021)Li, Li, and Si]{li2021preserving}
Yaoqian Li, Caizi Li, and Weixin Si.
\newblock Preserving the temporal consistency of video sequences for surgical
  instruments segmentation.
\newblock In \emph{2021 3rd International Conference on Intelligent Medicine
  and Image Processing (IMIP)}, pages 78--82. ACM, 2021.
\newblock \url{https://doi.org/10.1145%2F3468945.3468958}.

\bibitem[Zhang and Gao(2020)]{zhang2020object}
Jiayi Zhang and Xin Gao.
\newblock Object extraction via deep learning-based marker-free tracking
  framework of surgical instruments for laparoscope-holder robots.
\newblock \emph{International Journal of Computer Assisted Radiology and
  Surgery}, 15\penalty0 (8):\penalty0 1335--1345, 2020.
\newblock \url{https://doi.org/10.1007/s11548-020-02214-y}.

\bibitem[Islam et~al.(2021)Islam, VS, Lim, and Ren]{islam2021st}
Mobarakol Islam, Vibashan VS, Chwee~Ming Lim, and Hongliang Ren.
\newblock {ST-MTL:} spatio-temporal multitask learning model to predict
  scanpath while tracking instruments in robotic surgery.
\newblock \emph{Medical Image Analysis}, 67:\penalty0 101837, 2021.
\newblock \url{https://doi.org/10.1016/j.media.2020.101837}.

\bibitem[Wang et~al.(2021{\natexlab{c}})Wang, Jin, Wang, Cai, Heng, and
  Qin]{wang2021efficient}
Jiacheng Wang, Yueming Jin, Liansheng Wang, Shuntian Cai, Pheng-Ann Heng, and
  Jing Qin.
\newblock Efficient global-local memory for real-time instrument segmentation
  ofrobotic surgical video.
\newblock In \emph{International Conference on Medical Image Computing and
  Computer-Assisted Intervention {\textendash} MICCAI 2021}, pages 341--351.
  Springer International Publishing, 2021{\natexlab{c}}.
\newblock \url{https://doi.org/10.1007%2F978-3-030-87202-1_33}.

\bibitem[Jin et~al.(2019)Jin, Cheng, Dou, and Heng]{jin2019incorporating}
Yueming Jin, Keyun Cheng, Qi~Dou, and Pheng{-}Ann Heng.
\newblock Incorporating temporal prior from motion flow for instrument
  segmentation in minimally invasive surgery video.
\newblock In \emph{Medical Image Computing and Computer Assisted Intervention
  {\textendash} {MICCAI} 2019}, volume 11768 of \emph{Lecture Notes in Computer
  Science}, pages 440--448. Springer International Publishing, 2019.
\newblock \url{https://doi.org/10.1007/978-3-030-32254-0_49}.

\bibitem[Zhao et~al.(2020)Zhao, Jin, Gao, Dou, and Heng]{zhao2020learning}
Zixu Zhao, Yueming Jin, Xiaojie Gao, Qi~Dou, and Pheng-Ann Heng.
\newblock Learning motion flows for semi-supervisedinstrument segmentation
  fromrobotic surgical video.
\newblock In \emph{Medical Image Computing and Computer Assisted Intervention
  {\textendash} {MICCAI} 2020}, pages 679--689. Springer International
  Publishing, 2020.
\newblock \url{https://doi.org/10.1007%2F978-3-030-59716-0_65}.

\bibitem[Shimgekar et~al.(2021)Shimgekar, Pathi, and
  Santhi]{shimgekar2021voice}
Soorva~Ram Shimgekar, Preetham~Reddy Pathi, and V~Santhi.
\newblock Voice based segmentation of laparoscopic surgical tools and its image
  enhancement.
\newblock In \emph{2021 Innovations in Power and Advanced Computing
  Technologies (i-{PACT})}. IEEE, 2021.
\newblock \url{https://doi.org/10.1109%2Fi-pact52855.2021.9696600}.

\bibitem[Baby et~al.(2023)Baby, Thapar, Chasmai, Banerjee, Dargan, Suri,
  Banerjee, and Arora]{baby2023forks}
Britty Baby, Daksh Thapar, Mustafa Chasmai, Tamajit Banerjee, Kunal Dargan,
  Ashish Suri, Subhashis Banerjee, and Chetan Arora.
\newblock From forks to forceps: A new framework for instance segmentation of
  surgical instruments.
\newblock In \emph{Proceedings of the IEEE/CVF Winter Conference on
  Applications of Computer Vision}, pages 6191--6201. IEEE, 2023.
\newblock \url{https://doi.org/10.1109%2Fwacv56688.2023.00613}.

\bibitem[Cer{\'o}n et~al.(2021)Cer{\'o}n, Chang, Ruiz, and
  Ali]{ceron2021assessing}
Juan Carlos~{\'A}ngeles Cer{\'o}n, Leonardo Chang, Gilberto~Ochoa Ruiz, and
  Sharib Ali.
\newblock Assessing yolact++ for real time and robust instance segmentation of
  medical instruments in endoscopic procedures.
\newblock In \emph{2021 43rd Annual International Conference of the IEEE
  Engineering in Medicine \& Biology Society (EMBC)}, pages 1824--1827. IEEE,
  2021.
\newblock \url{https://doi.org/10.1109%2Fembc46164.2021.9629914}.

\bibitem[Cer{\'o}n et~al.(2022)Cer{\'o}n, Ruiz, Chang, and Ali]{ceron2022real}
Juan Carlos~{\'A}ngeles Cer{\'o}n, Gilberto~Ochoa Ruiz, Leonardo Chang, and
  Sharib Ali.
\newblock Real-time instance segmentation of surgical instruments using
  attention and multi-scale feature fusion.
\newblock \emph{Medical Image Analysis}, 81:\penalty0 102569, 2022.
\newblock \url{https://doi.org/10.1016%2Fj.media.2022.102569}.

\bibitem[Kitaguchi et~al.(2022{\natexlab{a}})Kitaguchi, Lee, Hayashi, Nakajima,
  Kojima, Hasegawa, Takeshita, Mori, and Ito]{kitaguchi2022development}
Daichi Kitaguchi, Younae Lee, Kazuyuki Hayashi, Kei Nakajima, Shigehiro Kojima,
  Hiro Hasegawa, Nobuyoshi Takeshita, Kensaku Mori, and Masaaki Ito.
\newblock Development and validation of a model for laparoscopic colorectal
  surgical instrument recognition using convolutional neural network--based
  instance segmentation and videos of laparoscopic procedures.
\newblock \emph{JAMA network open}, 5\penalty0 (8):\penalty0
  e2226265--e2226265, 2022{\natexlab{a}}.
\newblock \url{https://doi.org/10.1001%2Fjamanetworkopen.2022.26265}.

\bibitem[Kitaguchi et~al.(2022{\natexlab{b}})Kitaguchi, Fujino, Takeshita,
  Hasegawa, Mori, and Ito]{kitaguchi2022limited}
Daichi Kitaguchi, Toru Fujino, Nobuyoshi Takeshita, Hiro Hasegawa, Kensaku
  Mori, and Masaaki Ito.
\newblock Limited generalizability of single deep neural network for surgical
  instrument segmentation in different surgical environments.
\newblock \emph{Scientific Reports}, 12\penalty0 (1):\penalty0 12575,
  2022{\natexlab{b}}.
\newblock \url{https://doi.org/10.1038%2Fs41598-022-16923-8}.

\bibitem[Kurmann et~al.(2021)Kurmann, M{\'a}rquez-Neila, Allan, Wolf, and
  Sznitman]{kurmann2021mask}
Thomas Kurmann, Pablo M{\'a}rquez-Neila, Max Allan, Sebastian Wolf, and Raphael
  Sznitman.
\newblock Mask then classify: multi-instance segmentation for surgical
  instruments.
\newblock \emph{International journal of computer assisted radiology and
  surgery}, 16\penalty0 (7):\penalty0 1227--1236, 2021.
\newblock \url{https://doi.org/10.1007%2Fs11548-021-02404-2}.

\bibitem[Kletz et~al.(2019)Kletz, Schoeffmann, Benois-Pineau, and
  Husslein]{kletz2019identifying}
Sabrina Kletz, Klaus Schoeffmann, Jenny Benois-Pineau, and Heinrich Husslein.
\newblock Identifying surgical instruments in laparoscopy using deep learning
  instance segmentation.
\newblock In \emph{2019 International Conference on Content-Based Multimedia
  Indexing ({CBMI})}, pages 1--6. IEEE, 2019.
\newblock \url{https://doi.org/10.1109%2Fcbmi.2019.8877379}.

\bibitem[Kong et~al.(2021)Kong, Jin, Dou, Wang, Wang, Lu, Dong, Liu, and
  Sun]{kong2021accurate}
Xiaowen Kong, Yueming Jin, Qi~Dou, Ziyi Wang, Zerui Wang, Bo~Lu, Erbao Dong,
  Yun-Hui Liu, and Dong Sun.
\newblock Accurate instance segmentation of surgical instruments in robotic
  surgery: model refinement and cross-dataset evaluation.
\newblock \emph{International Journal of Computer Assisted Radiology and
  Surgery}, 16\penalty0 (9):\penalty0 1607--1614, 2021.
\newblock \url{https://doi.org/10.1007%2Fs11548-021-02438-6}.

\bibitem[Sun et~al.(2022)Sun, Zou, Wang, Su, and Guan]{sun2022parallel}
Xinan Sun, Yuelin Zou, Shuxin Wang, He~Su, and Bo~Guan.
\newblock A parallel network utilizing local features and global
  representations for segmentation of surgical instruments.
\newblock \emph{International Journal of Computer Assisted Radiology and
  Surgery}, pages 1--11, 2022.
\newblock \url{https://doi.org/10.1007%2Fs11548-022-02687-z}.

\bibitem[Gonz{\'a}lez et~al.(2020)Gonz{\'a}lez, Bravo-S{\'a}nchez, and
  Arbelaez]{gonzalez2020isinet}
Cristina Gonz{\'a}lez, Laura Bravo-S{\'a}nchez, and Pablo Arbelaez.
\newblock Isinet: An instance-based approachfor surgical instrument
  segmentation.
\newblock In \emph{Medical Image Computing and Computer Assisted Intervention
  {\textendash} {MICCAI} 2020}, pages 595--605. Springer International
  Publishing, 2020.
\newblock \url{https://doi.org/10.1007%2F978-3-030-59716-0_57}.

\bibitem[Lee et~al.(2020)Lee, Yu, Kwon, Kong, Lee, and Kim]{lee2020evaluation}
Dongheon Lee, Hyeong~Won Yu, Hyungju Kwon, Hyoun-Joong Kong, Kyu~Eun Lee, and
  Hee~Chan Kim.
\newblock Evaluation of surgical skills during robotic surgery by deep
  learning-based multiple surgical instrument tracking in training and actual
  operations.
\newblock \emph{Journal of Clinical Medicine}, 9\penalty0 (6):\penalty0 1964,
  2020.
\newblock \url{https://doi.org/10.3390%2Fjcm9061964}.

\bibitem[Kanakatte et~al.(2020)Kanakatte, Ramaswamy, Gubbi, Ghose, and
  Purushothaman]{kanakatte2020surgical}
Aparna Kanakatte, Akshaya Ramaswamy, Jayavardhana Gubbi, Avik Ghose, and
  Balamuralidhar Purushothaman.
\newblock Surgical tool segmentation and localization using spatio-temporal
  deep network.
\newblock In \emph{2020 42nd Annual International Conferences of the IEEE
  Engineering in Medicine and Biology Society (EMBC)}, pages 1658--1661. IEEE,
  2020.
\newblock \url{https://doi.org/10.1109%2Fembc44109.2020.9176676}.

\bibitem[Zhao et~al.(2022)Zhao, Jin, and Heng]{zhao2022trasetr}
Zixu Zhao, Yueming Jin, and Pheng-Ann Heng.
\newblock Trasetr: track-to-segment transformer with contrastive query for
  instance-level instrument segmentation in robotic surgery.
\newblock In \emph{2022 International Conference on Robotics and Automation
  (ICRA)}, pages 11186--11193. IEEE, 2022.
\newblock \url{https://doi.org/10.1109%2Ficra46639.2022.9811873}.

\bibitem[Zhang et~al.(2023)Zhang, Zhu, Peng, Han, and Liu]{zhang2023visual}
Chi Zhang, Wangru Zhu, Jianqing Peng, Yu~Han, and Wanquan Liu.
\newblock Visual servo control of endoscope-holding robot based on
  multi-objective optimization: System modeling and instrument tracking.
\newblock \emph{{SSRN} Electronic Journal}, 211:\penalty0 112658, 2023.
\newblock \url{https://doi.org/10.2139%2Fssrn.4174849}.

\bibitem[Gruijthuijsen et~al.(2022)Gruijthuijsen,
  Garc{\'{\i}}a{-}Peraza{-}Herrera, Borghesan, Reynaerts, Deprest, Ourselin,
  Vercauteren, and Poorten]{gruijthuijsen2022robotic}
Caspar Gruijthuijsen, Luis~C. Garc{\'{\i}}a{-}Peraza{-}Herrera, Gianni
  Borghesan, Dominiek Reynaerts, Jan Deprest, S{\'{e}}bastien Ourselin, Tom
  Vercauteren, and Emmanuel B.~Vander Poorten.
\newblock Robotic endoscope control via autonomous instrument tracking.
\newblock \emph{Frontiers in Robotics an {AI}}, 9:\penalty0 832208, 2022.
\newblock \url{https://doi.org/10.3389/frobt.2022.832208}.

\bibitem[Li et~al.(2023)Li, Huang, Zhang, Xie, Xian, Luo, Chiu, and
  Li]{li2023autonomous}
Jian Li, Yisen Huang, Xue Zhang, Ke~Xie, Yitian Xian, Xiao Luo, Philip Wai~Yan
  Chiu, and Zheng Li.
\newblock An autonomous surgical instrument tracking framework with a binocular
  camera for a robotic flexible laparoscope.
\newblock \emph{IEEE Robotics and Automation Letters}, 8\penalty0 (7):\penalty0
  4291--4298, 2023.
\newblock \url{https://doi.org/10.1109%2Flra.2023.3281934}.

\bibitem[Cheng et~al.(2021)Cheng, Li, Ng, Huang, Li, Ng, Chiu, and
  Li]{cheng2021deep}
Truman Cheng, Weibing Li, Wing~Yin Ng, Yisen Huang, Jixiu Li, Calvin Sze~Hang
  Ng, Philip Wai~Yan Chiu, and Zheng Li.
\newblock Deep learning assisted robotic magnetic anchored and guided endoscope
  for real-time instrument tracking.
\newblock \emph{{IEEE} Robotics and Automation Letters}, 6\penalty0
  (2):\penalty0 3979--3986, 2021.
\newblock \url{https://doi.org/10.1109/LRA.2021.3066834}.

\bibitem[Zinchenko and Song(2021)]{zinchenko2021autonomous}
Kateryna Zinchenko and Kai{-}Tai Song.
\newblock Autonomous endoscope robot positioning using instrument segmentation
  with virtual reality visualization.
\newblock \emph{{IEEE} Access}, 9:\penalty0 72614--72623, 2021.
\newblock \url{https://doi.org/10.1109/ACCESS.2021.3079427}.

\bibitem[Mendel et~al.(2023)Mendel, Rauber, de~Souza~Jr, Papa, and
  Palm]{mendel2023error}
Robert Mendel, David Rauber, Luis~A de~Souza~Jr, Jo{\\textasciitilde a}o~P
  Papa, and Christoph Palm.
\newblock Error-correcting mean-teacher: Corrections instead of
  consistency-targets applied to semi-supervised medical image segmentation.
\newblock \emph{Computers in Biology and Medicine}, 154:\penalty0 106585, 2023.
\newblock \url{https://doi.org/10.1016%2Fj.compbiomed.2023.106585}.

\bibitem[Ni et~al.(2022{\natexlab{b}})Ni, Bian, Li, Zhou, Li, and
  Hou]{ni2022space}
Zhen-Liang Ni, Gui-Bin Bian, Zhen Li, Xiao-Hu Zhou, Rui-Qi Li, and Zeng-Guang
  Hou.
\newblock Space squeeze reasoning and low-rank bilinear feature fusion for
  surgical image segmentation.
\newblock \emph{IEEE Journal of Biomedical and Health Informatics}, 26\penalty0
  (7):\penalty0 3209--3217, 2022{\natexlab{b}}.
\newblock \url{https://doi.org/10.1109%2Fjbhi.2022.3154925}.

\bibitem[Wang et~al.(2021{\natexlab{d}})Wang, Li, Nakashima, Kawasaki,
  Nagahara, and Yagi]{wang2021noisy}
Bowen Wang, Liangzhi Li, Yuta Nakashima, Ryo Kawasaki, Hajime Nagahara, and
  Yasushi Yagi.
\newblock Noisy-{LSTM}: Improving temporal awareness for video semantic
  segmentation.
\newblock \emph{IEEE Access}, 9:\penalty0 46810--46820, 2021{\natexlab{d}}.
\newblock \url{https://doi.org/10.1109%2Faccess.2021.3067928}.

\end{thebibliography}

\end{document}